\documentclass{article}

\usepackage[preprint]{arxiv}

\usepackage[utf8]{inputenc} 
\usepackage[T1]{fontenc}    
\usepackage{hyperref}       
\usepackage{url}    
\usepackage{booktabs}       
\usepackage{amsfonts}       
\usepackage{nicefrac}       
\usepackage{microtype}      
\usepackage{amssymb}
\usepackage{amsmath}
\usepackage{amsthm}
\usepackage{mathtools}
\usepackage{multirow}
\usepackage{pifont}
\usepackage{graphicx}
\usepackage{subcaption}
\usepackage{float}
\usepackage{makecell}
\usepackage{xcolor}
\usepackage{wrapfig}
\newcommand{\cmark}{\ding{51}}
\newcommand{\xmark}{\ding{55}}
\newcommand{\model}{CM3Leon}
\newcommand{\mm}{multi-modal}
\usepackage[font=small]{caption}

\definecolor{MyDarkGreen}{rgb}{0.015,0.45,0.015}
\definecolor{MyPurple}{RGB}{111,0,255}
\definecolor{saffron}{RGB}{227, 170, 0}
\usepackage{xcolor}

\newcommand{\exper}[1]{\textsc{#1}}
\newcommand{\pexpert}{p_\exper{exp}}
\newcommand{\pamateur}{p_\exper{ama}}
\newcommand{\kmax}[2]{\underset{#2}{\mathrm{kmax}}\left(#1\right)}

\title{Scaling Autoregressive Multi-Modal Models: Pretraining and Instruction Tuning}

\author{%
  Lili Yu\thanks{First Author} \And Bowen Shi\footnotemark[1] \And Ramakanth Pasunuru\footnotemark[1] \And Benjamin Muller \And Olga Golovneva \And Tianlu Wang \And Arun Babu \And Binh Tang\And Brian Karrer\And Shelly Sheynin \And Candace Ross \And Adam Polyak \And Russell Howes 
  \And Vasu Sharma \And Puxin Xu \And Hovhannes Tamoyan$^1$ \And Oron Ashual \And Uriel Singer \And Shang-Wen Li \And Susan Zhang \And Richard James \And Gargi Ghosh \And Yaniv Taigman \And Maryam Fazel-Zarandi \And Asli Celikyilmaz \And Luke Zettlemoyer \And Armen Aghajanyan\footnotemark[1] \\
  \hspace{-17em} FAIR, YerevaNN$^1$ \\
  \hspace{-17em} \texttt{armenag@meta.com} \\
}

\begin{document}
\maketitle
\vspace{-3em}
\begin{figure}[H]
    \centering
    \begin{tabular}{cc}
        \begin{subfigure}{0.25\textwidth}
    \includegraphics[width=\textwidth]{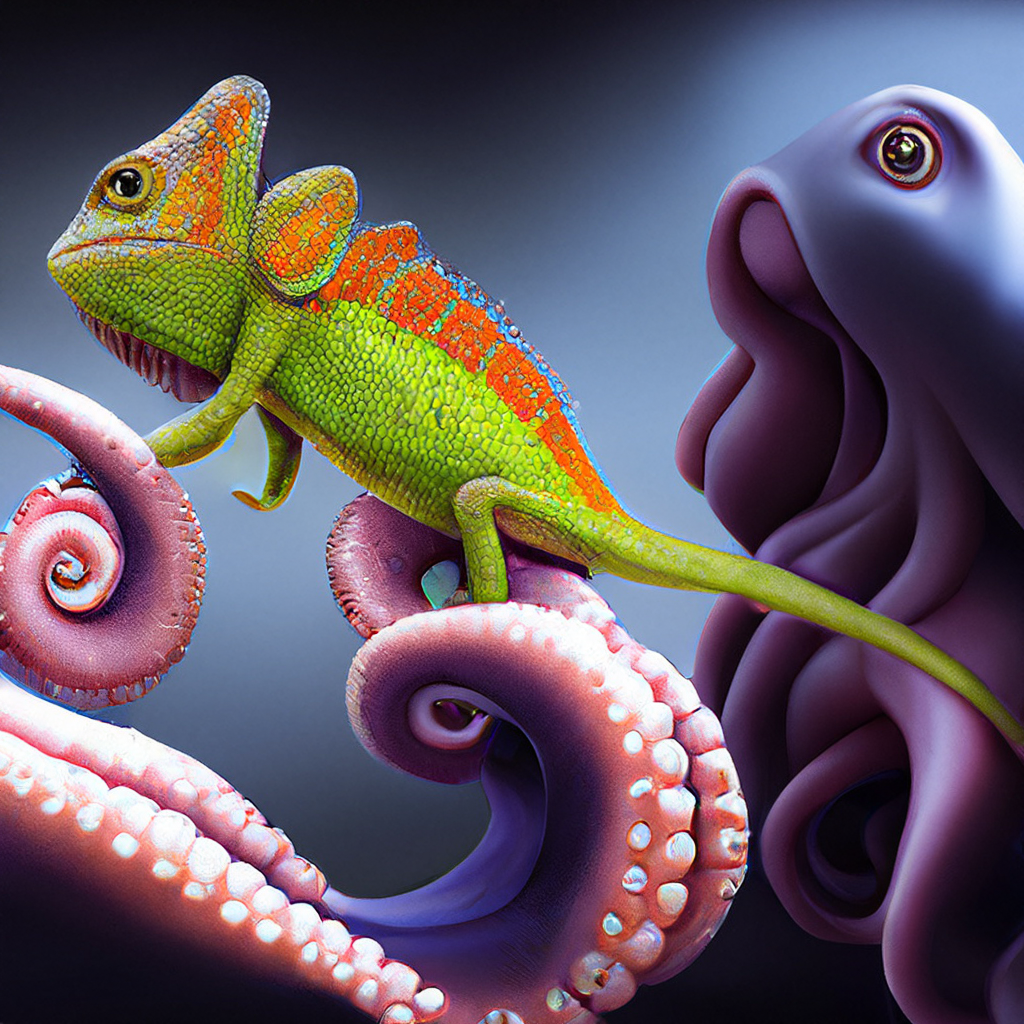}
        \end{subfigure}
        \begin{subfigure}{0.25\textwidth}
    \includegraphics[width=\textwidth]{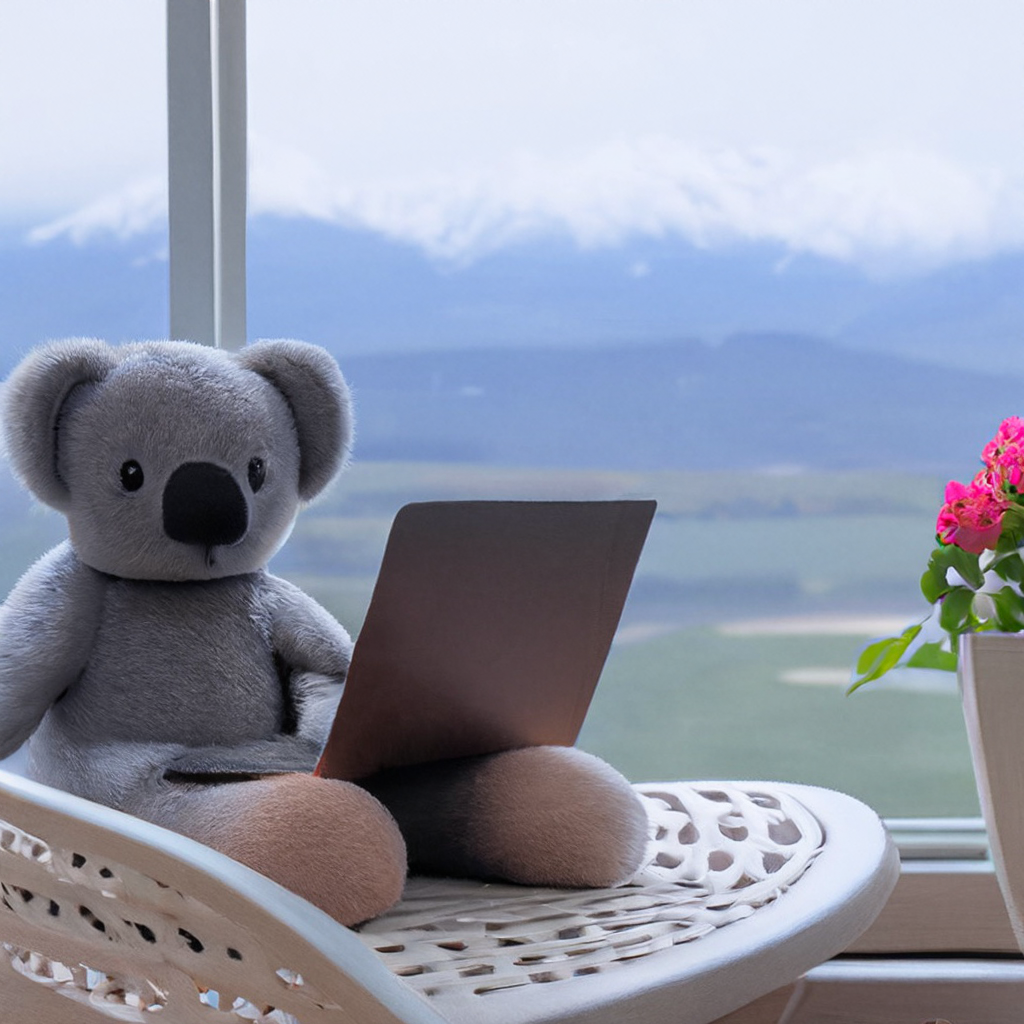}
        \end{subfigure}
        \begin{subfigure}{0.25\textwidth}
    \includegraphics[width=\textwidth]{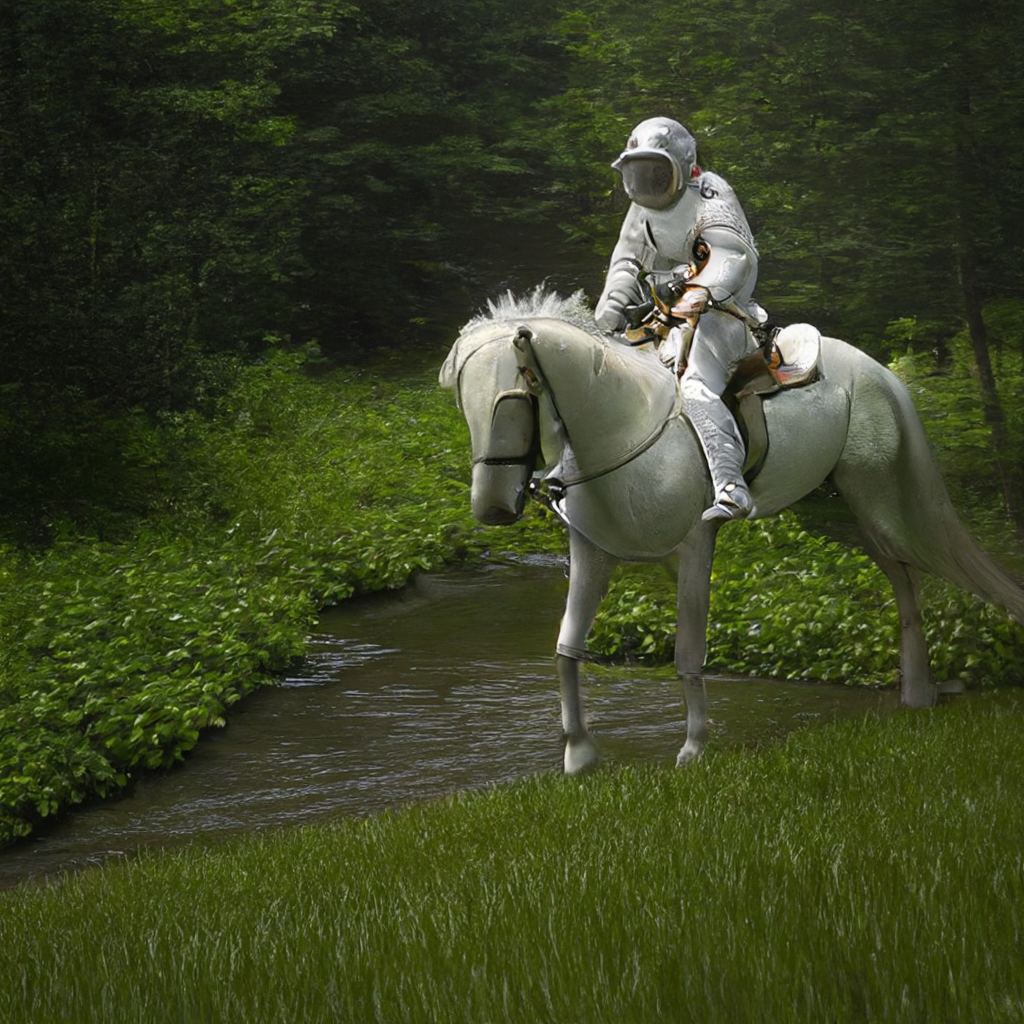}
        \end{subfigure}
        \begin{subfigure}{0.25\textwidth}
    \includegraphics[width=\textwidth]{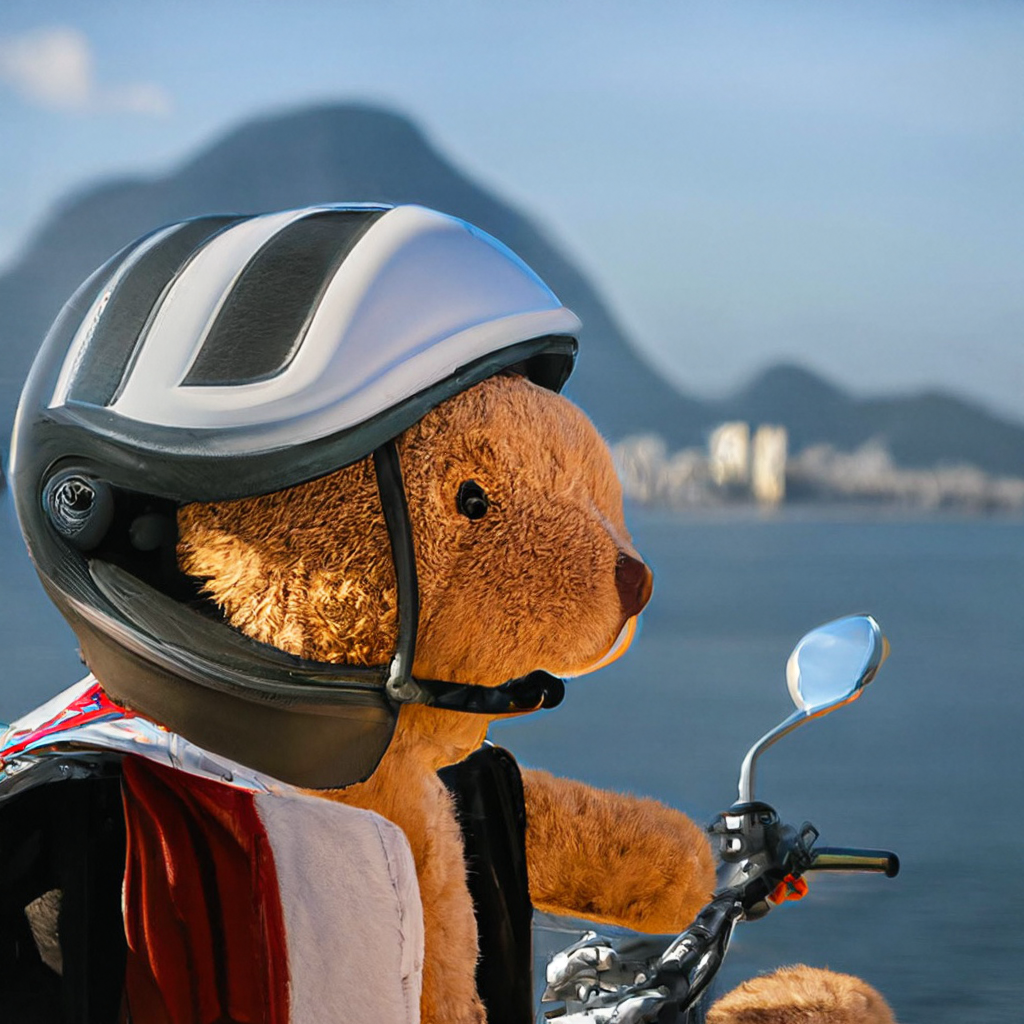}
        \end{subfigure}\\
        \begin{subfigure}{0.25\textwidth}
    \includegraphics[width=\textwidth]{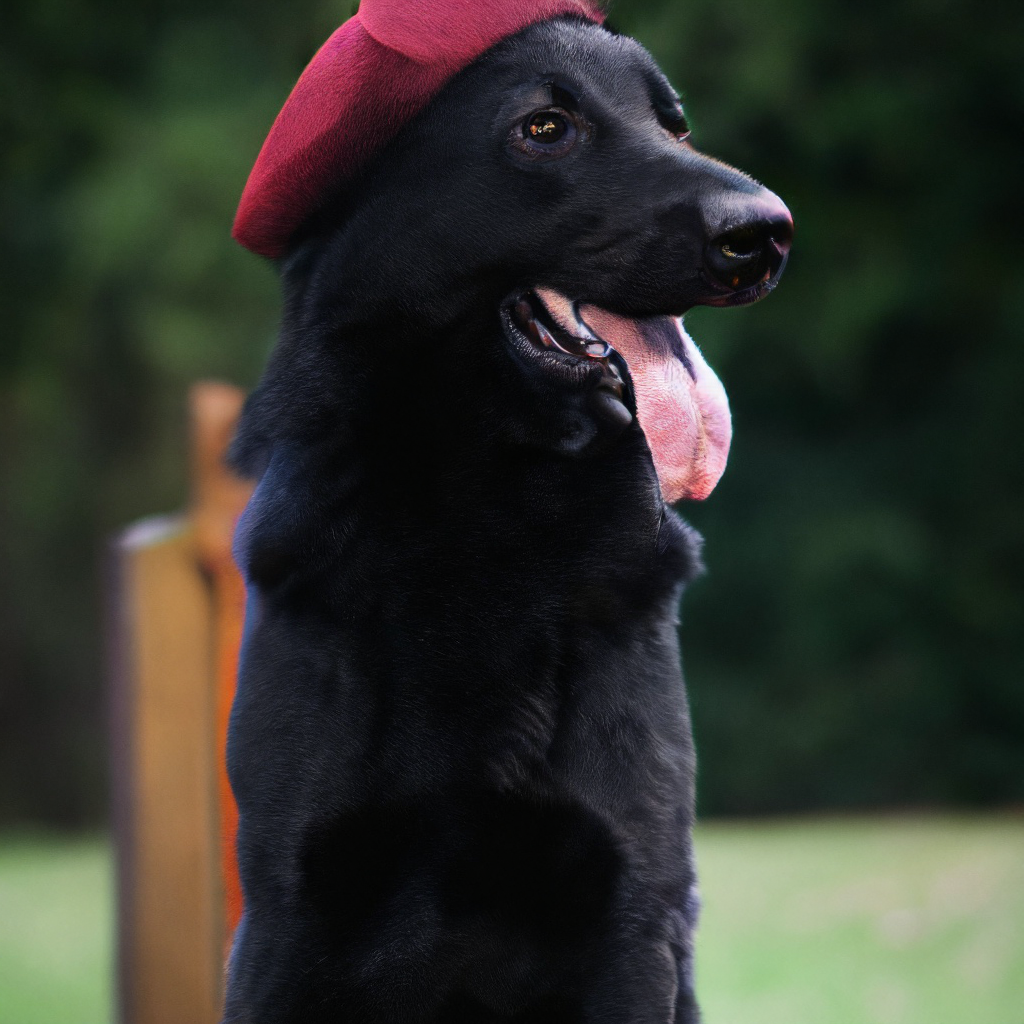}
        \end{subfigure}
        \begin{subfigure}{0.25\textwidth}
    \includegraphics[width=\textwidth]{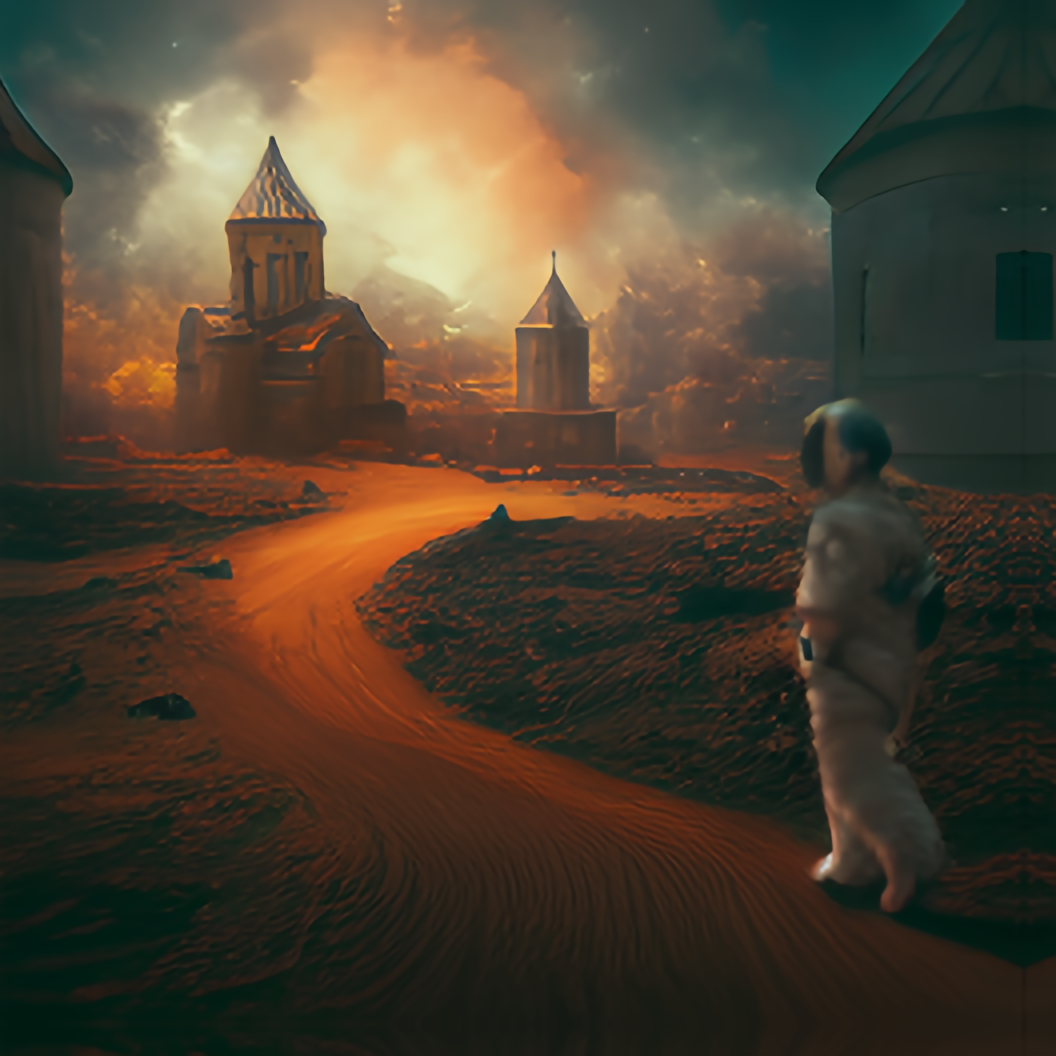}
        \end{subfigure}
        \begin{subfigure}{0.25\textwidth}
    \includegraphics[width=\textwidth]{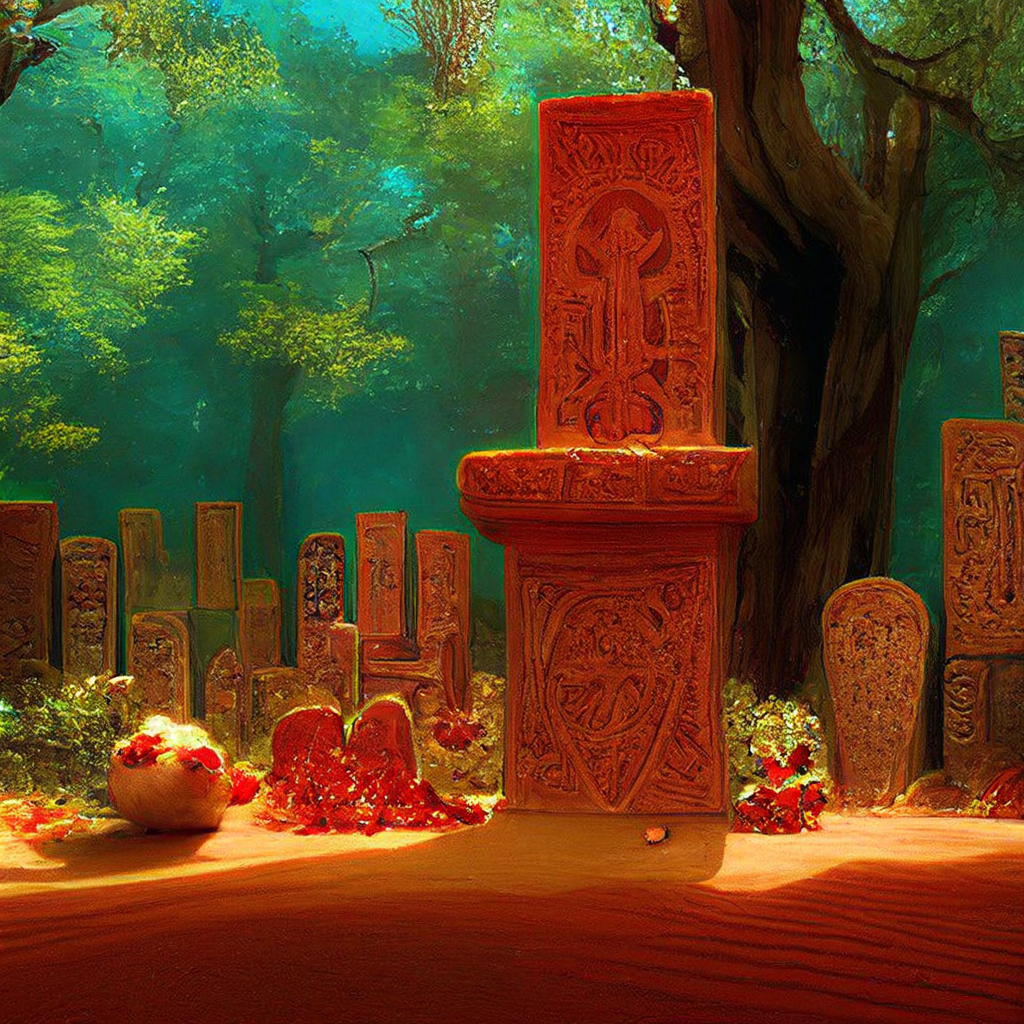}
        \end{subfigure}
        \begin{subfigure}{0.25\textwidth}
    \includegraphics[width=\textwidth]{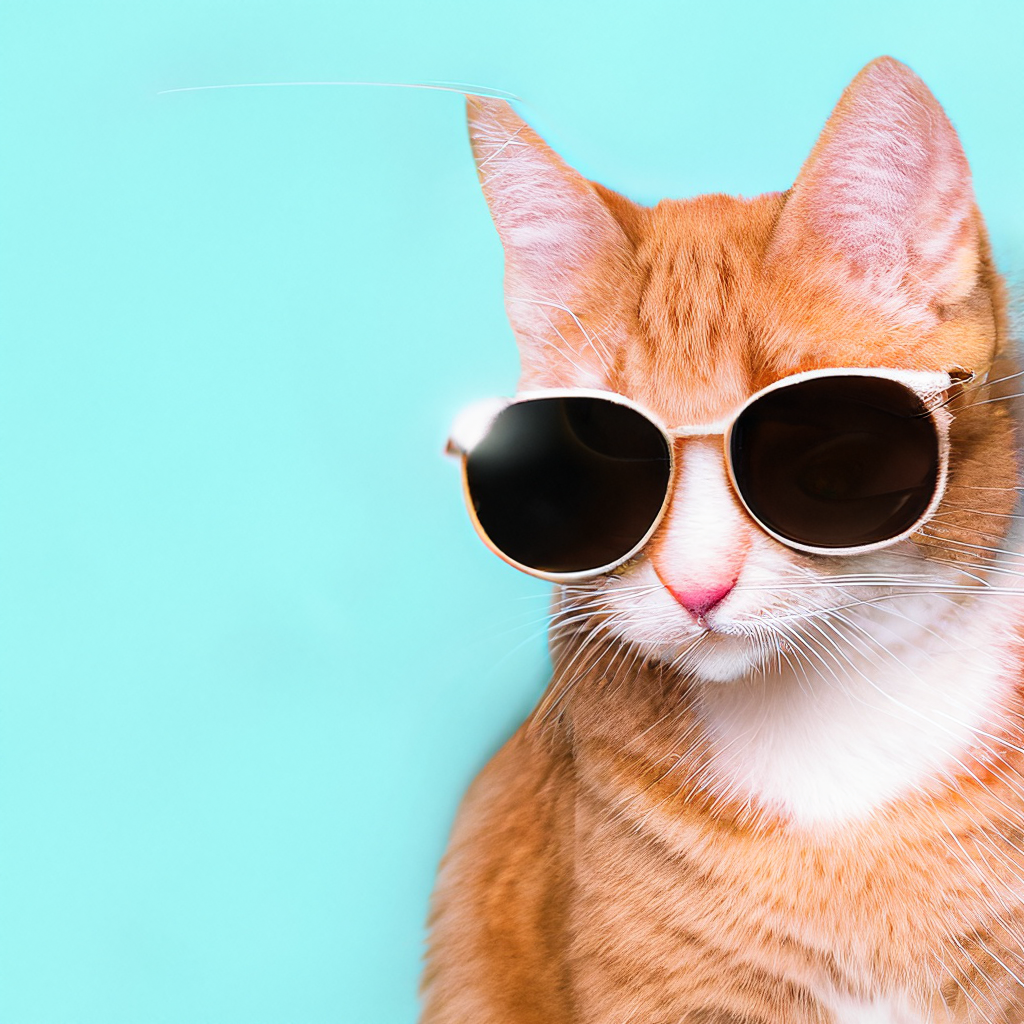}
        \end{subfigure}\\
        \begin{subfigure}{0.25\textwidth}
    \includegraphics[width=\textwidth]{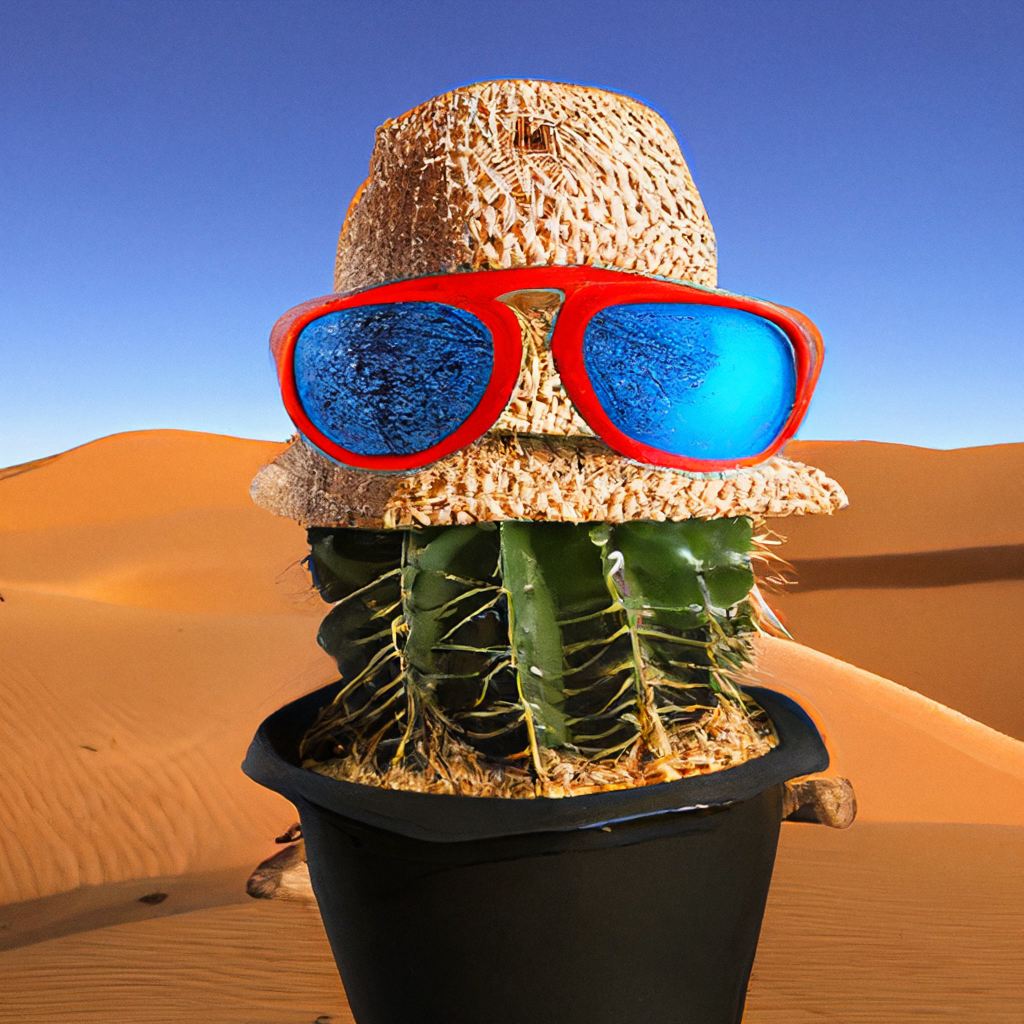}
        \end{subfigure}
        \begin{subfigure}{0.25\textwidth}
    \includegraphics[width=\textwidth]{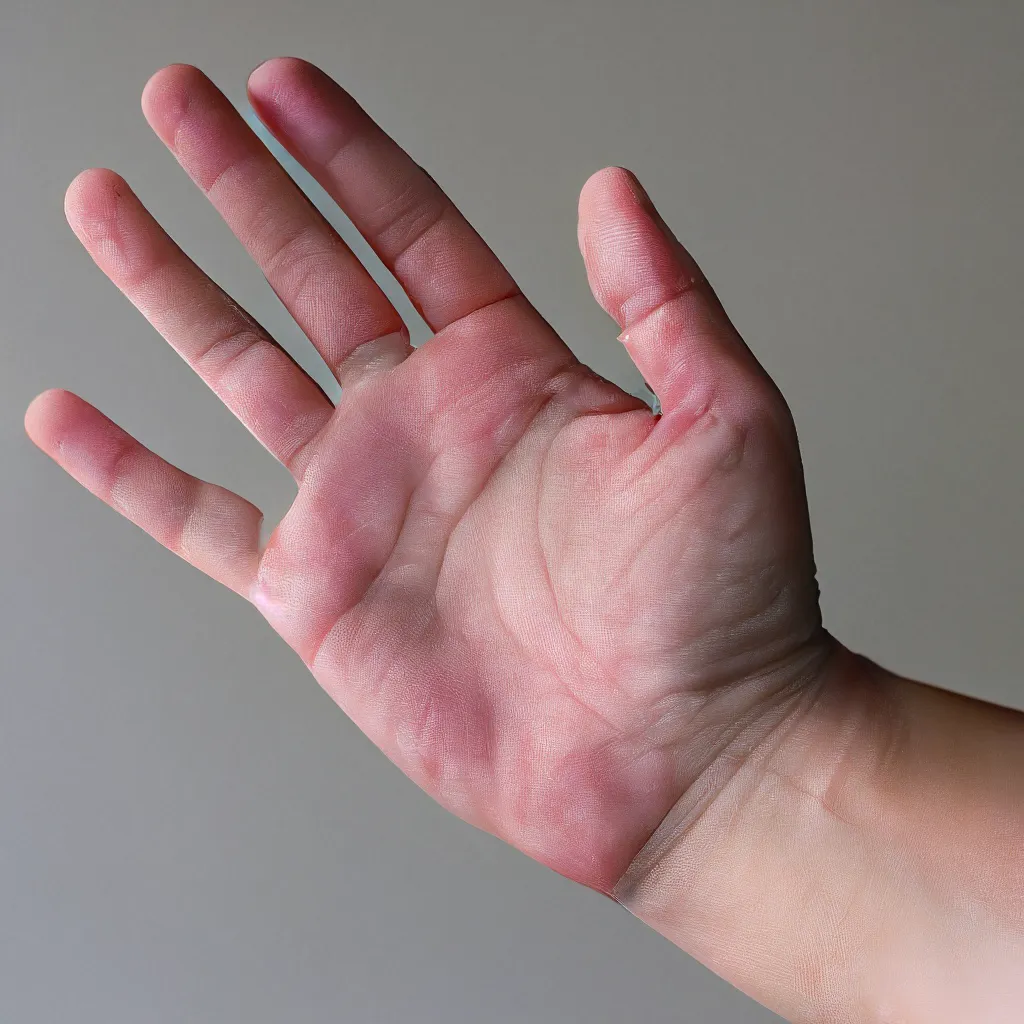}
        \end{subfigure}
        \begin{subfigure}{0.25\textwidth}
    \includegraphics[width=\textwidth]{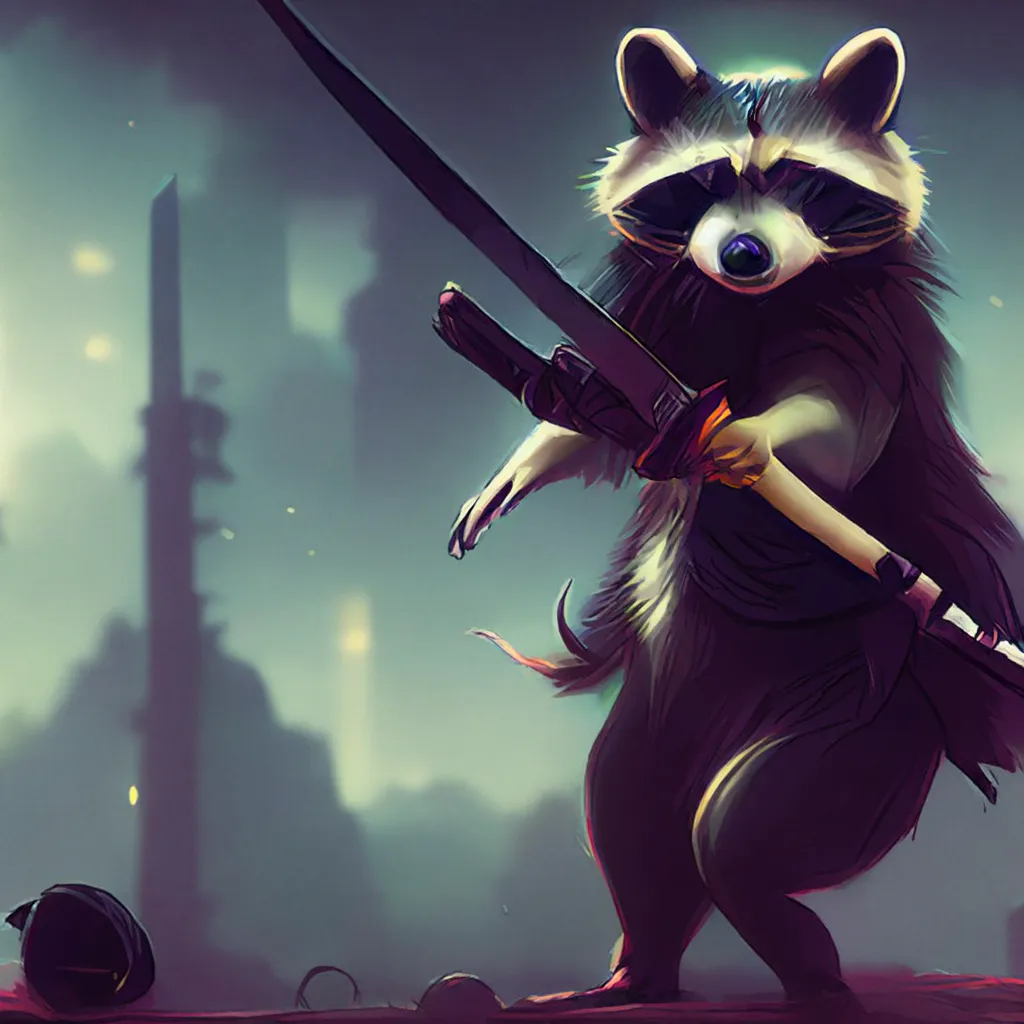}
        \end{subfigure}
        \begin{subfigure}{0.25\textwidth}
    \includegraphics[width=\textwidth]{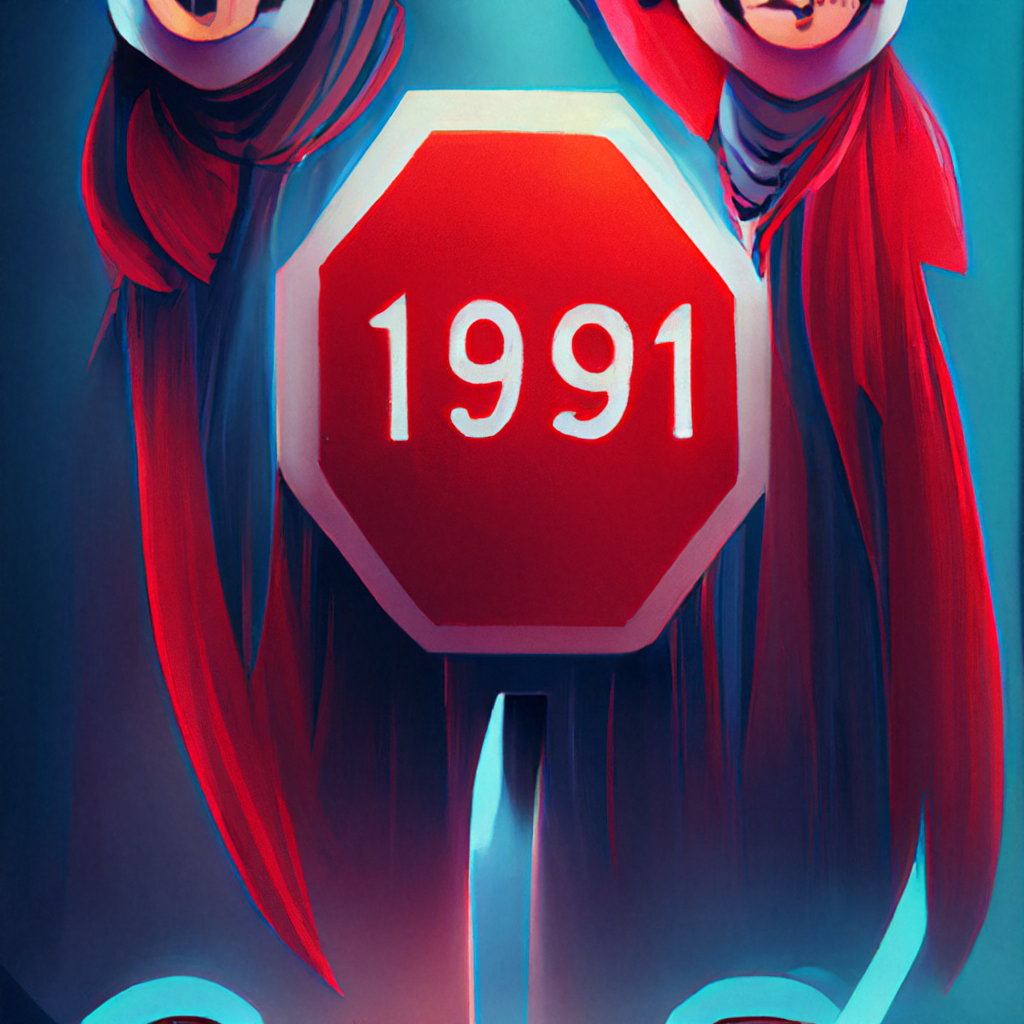}
        \end{subfigure}
    \end{tabular}
    \caption{Showcase of \model{} zero-shot generations (no-retrieval augmentation). Refer to \S~\ref{sec:showcase_prompts} for a complete list of prompts. \model{} can generate complex compositional objects, tail entities (Khachkar--Armenian crosses carved from stone), and historically hard entities such as hands and text.}
\end{figure}

\begin{abstract}
We present \model\ (pronounced  ``Chameleon''), a retrieval-augmented, token-based, decoder-only \mm\ language model capable of generating and infilling both text and images. \model\ uses the CM3 \mm\ architecture but additionally shows the extreme benefits of scaling up and tuning on more diverse instruction-style data. It is the first \mm\ model trained with a recipe adapted from text-only language models, including a large-scale retrieval-augmented pretraining stage and a second multi-task supervised fine-tuning (SFT) stage. It is also a general-purpose model that can do both text-to-image and image-to-text generation, allowing us to introduce self-contained contrastive decoding methods that produce high-quality outputs.  
Extensive experiments demonstrate that this recipe is highly effective for \mm\ models. \model\ achieves state-of-the-art performance in text-to-image generation with 5x less training compute than comparable methods (\textbf{zero-shot MS-COCO FID of 4.88}). After SFT, \model\ can also demonstrate unprecedented levels of controllability in tasks ranging from language-guided image editing to image-controlled generation and segmentation.
\end{abstract}

\section{Introduction}

Diffusion models have recently dominated image generation work due to their strong performance and relatively modest computational cost~\citep{IMAGEN, REIMAGEN, LDM}. In contrast, token-based autoregressive models~ \citep{DALLE,PARTI} are known to also produce strong results, with even better global image coherence in particular, but are much more expensive to train and use for inference. In this paper, we show that it is possible to extend training and inference ideas originally developed for text-only models to flip this narrative; autoregressive models can be efficient and performant while also generalizing beyond the strict text-to-image format to be tuneable for a wide range of image and text generation tasks. 

More specifically, we introduce CM3Leon (pronounced ``Chameleon''), a retrieval-augmented, token-based, decoder-only \mm\ language model capable of generating and infilling both text and images. 
\model\ uses the CM3 \mm\ architecture~\citep{CM3}, but additionally shows the extreme benefits of scaling up and training on more diverse data. 
It is the first \mm\ model trained with a recipe adapted from text-only language models, including a large-scale retrieval-augmented pretraining stage and a second multi-task supervised fine-tuning (SFT) stage. 
The pretraining is efficient because it follows the retrieval-augmented CM3 approach~\citep{RA_CM3} but uses a new large-scale Shutterstock dataset that includes only licensed image and text data. The SFT stage follows multi-task instruction tuning for text-only models~\citet{instruct_opt}, which allow arbitrary mixtures of image and text tokens in both the inputs and outputs. 
The generality of \model\ also supports the introduction of an improved, self-contained contrastive decoding method~\cite {contrastive_decoding}, which can provide self-guidance to improve both text and image generation.  

\model\ achieves state-of-the-art performance in text-to-image generation with 5x less training compute than comparable methods (\textbf{zero-shot MS-COCO FID of 4.88}). 
It can also do non-trivial image-to-text generation, even though it was trained on only 3B Shutterstock text tokens. 
After SFT, \model\ demonstrates unprecedented levels of controllability in tasks ranging from language-guided image editing to image-controlled generation and segmentation.
We also show that retrieval augmentation is key for efficient training, and our new contrastive decoding method enables much higher quality generation overall. 
These results strongly suggest that autoregressive models are worth significantly more study for any text and image task. 

\section{Pretraining}

We explore the potential of token-based decoder-only models in the text-to-image domain by building upon the foundation laid by RA-CM3 \cite{RA_CM3}.  We simplify the original settings in RA-CM3 by streamlining the objective, modifying the dataset, and incorporating insights from \mm\ scaling laws presented by \cite{mm_scaling_laws}.

\subsection{Data}
The ethical implications of image data sourcing in the domain of text-to-image generation have been a topic of considerable debate. In this study, we use only licensed images from Shutterstock. As a result, we can avoid concerns related to images ownership and attribution, without sacrificing performance.

\paragraph{Image Tokenization}
We use the image tokenizer from \citet{make_a_scene}, which encodes a $256\times256$ image into $1024$ tokens from a vocabulary of $8192$. For text, we train a custom tokenizer over the \citet{OPT} data with a vocabulary size of $56320$. Additionally, we introduce a novel special token, denoted as \texttt{<break>}, which serves to indicate a transition between modalities. A visualization of one caption-image pair after tokenization and formatting with our special tokens is available in \S~\ref{sec:data_visualizations}(Figure~\ref{fig:single_image_tokenization}).

\paragraph{Retrieval Augmentation}
Our retrieval approach aims to retrieve relevant and diverse \mm\ documents from a memory bank, given an input sequence \citep{RA_CM3}. It includes both a dense retriever and a retrieval strategy.

The dense retriever takes a query $q$ (e.g., the input sequence $x$) and a candidate document $m$ from the memory bank $\mathcal{M}$ and returns a relevance score $r(q, m)$. We adopt the dense retrieval method from \citet{dpr}, which uses a bi-encoder architecture. 
The encoder is CLIP-based. We split the \mm\ document into a text part and an image part, encode them separately using off-the-shelf frozen CLIP text and image encoders, and then average the two as a vector representation of the document~\citep{CLIP}. We use the ViT-B-32 model and normalize the image/text embeddings. 
The final retrieval is done with Maximum Inner Product Search (MIPS) over the memory bank using the dense retriever to obtain a list of candidate documents sorted by relevance score \citep{MIPS}.

To sample informative retrieved documents for the generator during training, we consider three key factors: relevance, modality, and diversity. First, the retrieved documents should be relevant to the input sequence, captured by the dense retriever score based on CLIP. Second, retrieving a \mm\ document consisting of images and text leads to better generator performance than retrieving either image or text. Third, diversity is essential to avoid redundancy in the retrieved documents. Simply taking the top $K$ documents based on relevance score can result in duplicates or highly similar documents, hurting downstream pretraining. We skip a candidate document if it is too similar to the query or if the documents have already been retrieved. In practice, we only use retrieved documents with relevance score $\leq 0.9$. Additionally, we use query dropout, which drops some tokens of the query used in retrieval (20\% of tokens) to encourage diversity and serve as regularization for training.

Throughout our work, we retrieve two documents each, based on image and text, respectively. In training, we randomly select three retrieved samples for every caption-image pair in our dataset, effectively 4x the number of tokens available in the pretraining. A visualization of a single training example can be found in \S~\ref{sec:data_visualizations}(Figure~\ref{fig:multiple_image_tokenization}).

\subsection{Objective Function}
\label{sec:objective}
The CM3 objective accepts \mm\ inputs (e.g., $x_\text{input}=$ "Image of a chameleon: \texttt{[image]}") and transforms them into an infilling instance by masking specific spans and relocating them to the end (e.g., $x_\text{input}=$ "Image of \texttt{<mask>}: \texttt{[image]} \texttt{<infill>} a chameleon"). It uses a standard next token prediction loss, $-\log p(x_\text{input})$. This results in a versatile model capable of infilling and autoregressive generation tasks for both images and text. In the case of caption-to-image generation, CM3 creates a continuation from the prompt "Image of a chameleon:". For image-to-caption generation, CM3 utilizes the prompt "Image of \texttt{<mask>}: \texttt{[image]} \texttt{<infill>}".

\citet{RA_CM3} built upon the original CM3 by including retrieved \mm\ documents in the context for each training example and up weighting the query image-caption pair loss, as illustrated in the last image-caption pair in Figure~\ref{fig:multiple_image_tokenization}. This approach encourages the model to concentrate more on using retrieved samples during the generation process. However, this method adversely affects the zero-shot scenario, where the goal is to generate an image without retrieval, such as predicting a continuation from \texttt{<eos> text <break>}. We remove this weighting in our setting and make a minor modification to the CM3 objective by preventing masking across \texttt{<break>} tokens. This adjustment is justified by the fact that allowing masking across \texttt{<break>} tokens may lead to the model generating image content from an arbitrary midpoint, which is not a desirable outcome. 

\subsection{Model}
The \model{} models follow a decoder-only transformer architecture, similar to \citet{OPT} and \citet{gpt3}. Compared to \citet{OPT}, we remove bias terms, dropout, and learnable parameters for layer norms and use a sequence length of 4096 instead of 2048. For weight initialization, we use a truncated normal distribution with a mean of 0 and a standard deviation of 0.006, truncated to 3 standard deviations. Output layers are initialized as 0, and the learned absolute positional embedding is initialized near zero with a standard deviation of 0.0002. The models were trained with Metaseq\footnote{\url{https://github.com/facebookresearch/metaseq}}, with experiment tracking done with Aim \citet{AIM}.

\subsection{Training}
\begin{figure}[htbp]
  \centering
  \begin{minipage}[b]{0.48\textwidth}
    \centering
    \includegraphics[width=0.9\textwidth]{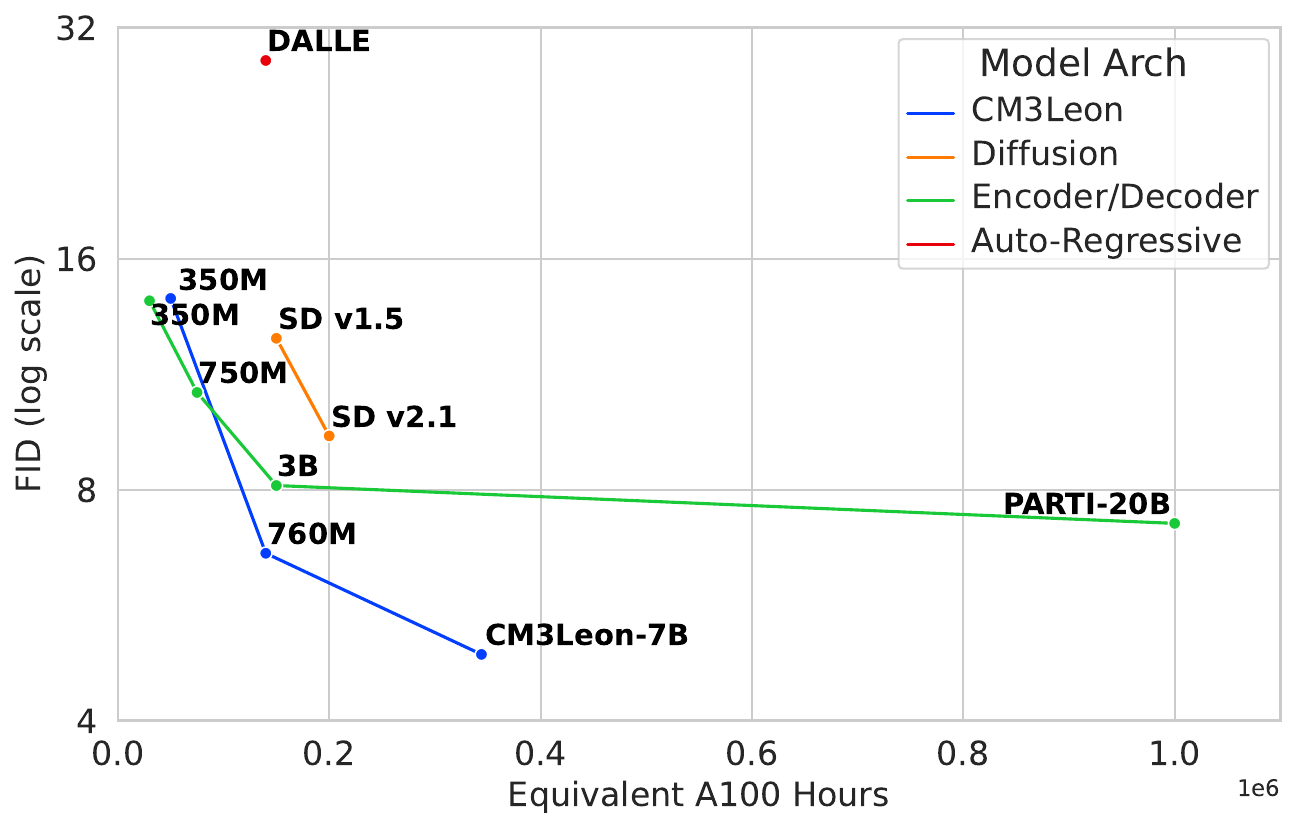}
    \caption{We plot FID score in log scale of various models against the equivalent A100 GPU hours during training. \model{} scales better than DALLE~\citep{DALLE}, stable diffusion~(SD)~\citep{LDM} and PARTI~\citep{PARTI} models. }
  \label{fig:how_we_scale}
  \end{minipage}
  \hfill
  \begin{minipage}[b]{0.50\textwidth}
    \centering
    \includegraphics[width=\textwidth]{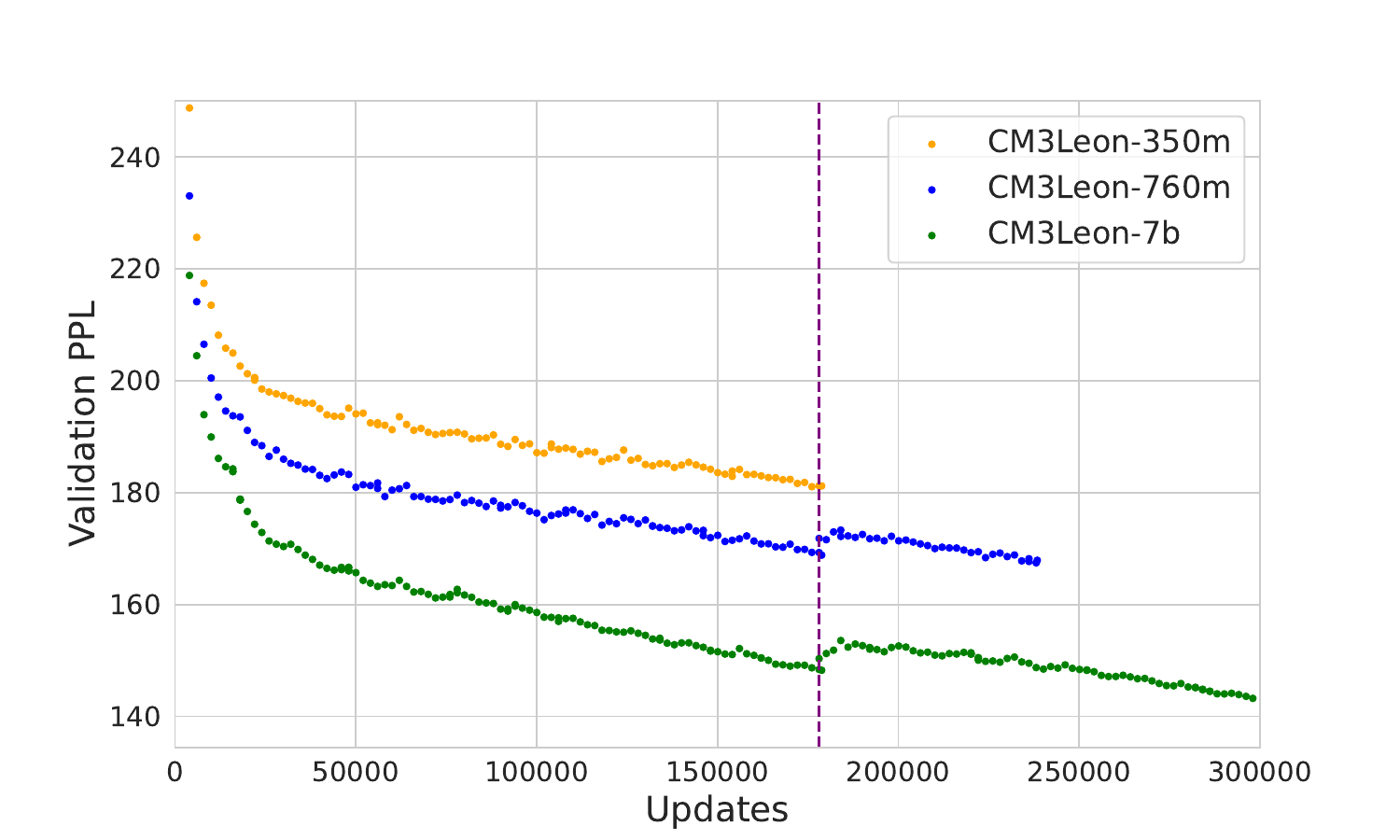}
    \caption{We plot validation perplexity (PPL) against with number of training updates for CM3Leon models in 350m, 760m and 7b size. We resume the training of 760m and 7b models after a full epoch (the purple dashed line), and the small rise in the PPL is due to the sudden increase of the learning rate.}
    \label{fig:training_curve}
  \end{minipage}
\end{figure}

Our models are trained across three distinct sizes, with the corresponding parameters and training setup detailed in Table~\ref{tab:pretraing_para}. The major hyperparameters, such as the learning rate and batch size, are adopted from prior work in \mm\ scaling laws, creating a stable and smooth training progression as illustrated in Figure~\ref{fig:training_curve} \citep{mm_scaling_laws}.
The 350 Million (350M), 760 Million (760M), and 7 Billion (7B) models are trained to 1.4 Trillion (T), 1.9T, and 2.4T tokens, respectively. The losses for all three models decrease steadily throughout training, strongly suggesting they have not saturated. 

\section{Text-To-Image Results}

\subsection{Importance of Decoding Strategies}
\label{sec:decoding}
There has been significant work on developing decoding algorithms for autoregressive text-to-image models, such as DALL-E~\citet{DALLE}, which can have a large effect on the quality of the final outputs. DALL-E employs temperature sampling and a re-ranking stage via CLIP over 512 prompt candidates. Models like PARTI and Make-A-Scene user token-based classifier-free guidance, significantly reducing the number of candidates required for re-ranking to just 16 samples \citep{PARTI, make_a_scene}. Our experiments show that different approaches offer complementary benefits, as decribed in this section. We compare the following options.

\paragraph{Temperatured Sampling} is a probabilistic technique used in autoregressive models, such as \citet{DALLE}. The method involves modifying the softmax temperature during the sampling stage to control the randomness of predictions. We pair this with Classifier Free Guidance in all of our experiments.

\paragraph{TopP Sampling} also known as nucleus sampling, involves sampling from the smallest set of top-ranked tokens with a cumulative probability exceeding a predefined threshold \citep{holtzman2020nucleus}. We pair this with Classifier Free Guidance in all of our experiments.

\paragraph{Classifier Free Guidance (CFG)}
Classifier-free guidance refers to directing an unconditional sample towards a conditional sample \citep{make_a_scene}. We replace the text with the mask token from the CM3 objective to facilitate unconditional sampling. This is one of the core benefits of training with the CM3 objective, allowing us to do classifier-free guidance without the need for finetuning. During the inference stage, two concurrent token streams are generated: a conditional token stream, which is contingent on the input text, and an unconditional token stream, which is conditioned on a mask token. Borrowing the notation from \citet{make_a_scene}:

\begin{align}
&\text{logits}_{\text{cond}} = T(t_y|t_x), \text{logits}_{\text{uncond}} = T(t_y|\texttt{<mask>}), \\
&\text{logits}_{\text{cf}} = \text{logits}_{\text{uncond}} + \alpha_c \cdot (\text{logits}_{\text{cond}} - \text{logits}_{\text{uncond}}) \label{eq:logit_subtraction}
\end{align}

where $T$ denotes the transformer, $t_y$ is the output tokens and $t_x$ is the conditional input text, \texttt{<mask>} represents the absence of input text (and replacement with a mask token), and $\alpha_c$ is a scaling factor. The classifier-free guidance effectively blends the unconditional and conditional logits, influencing the model's output towards a more desired conditional output.

\paragraph{Contrastive Decoding TopK (CD-K)}
A key insight is that the logit subtraction in Equation~\ref{eq:logit_subtraction} resembles the log probability subtraction in contrastive decoding methods in text \citep{contrastive_decoding}. This leads us to propose a variant of the contrastive decoding (CD) algorithm, originally proposed by \citet{contrastive_decoding}, as an alternative to CFG.

Recall that CD defines a score per token:
\begin{align*}
& CD({t_y}_i; {t_y}_{<i})
= \begin{cases}
\log \frac{\pexpert ( {t_y}_i \mid {t_y}_{<i})}{ \pamateur( {t_y}_i \mid {t_y}_{<i})},
& \text{if } {t_y}_i \in \mathcal{V}({t_y}_{<i}) \text{,}\\
-\inf,    	      & \text{otherwise.} 
\end{cases}  \nonumber 
\end{align*}
Here, $\mathcal{V}({t_y}_{<i})$ represents the set of potential subsequent tokens whose probabilities are at least $\alpha$ times the maximum probability value:
\begin{align*}
\mathcal{V}({t_y}_{<i}) = \{{t_y}_i \in \mathcal{V}: \pexpert({t_y}_i \mid {t_y}_{<i}) \geq \alpha \max_w \pexpert(w | {t_y}_{<i}) \}
\end{align*}

Traditionally $\pexpert$ and $\pamateur$ in the CD decoding algorithm represent a strong and weak model where the strong model was trained with more compute (or larger model size) compared to the weak model. Instead we select $\pexpert$ having text conditioning and $\pamateur$ has no text conditioning. Additionally we saw that the $\mathcal{V}({t_y}_{<i})$ constraint was too strict, and would consistently become greedy decoding. Therefore we propose a slight modification of CD we call CD-K that alters $\mathcal{V}({t_y}_{<i})$ to:
\begin{equation}
    \mathcal{V}({t_y}_{<i}) = \{{t_y}_i \in \mathcal{V}: \pexpert({t_y}_i \mid {t_y}_{<i}) \geq \alpha *\kmax{\pexpert(w | {t_y}_{<i})}{k, w} \}
\end{equation}
where instead of taking the largest probability we take the $k$-th largest probability.

\paragraph{Ablation}

\begin{figure}[htbp]
  \centering
  \begin{minipage}[b]{0.49\textwidth}
    \centering
    \includegraphics[width=\textwidth]{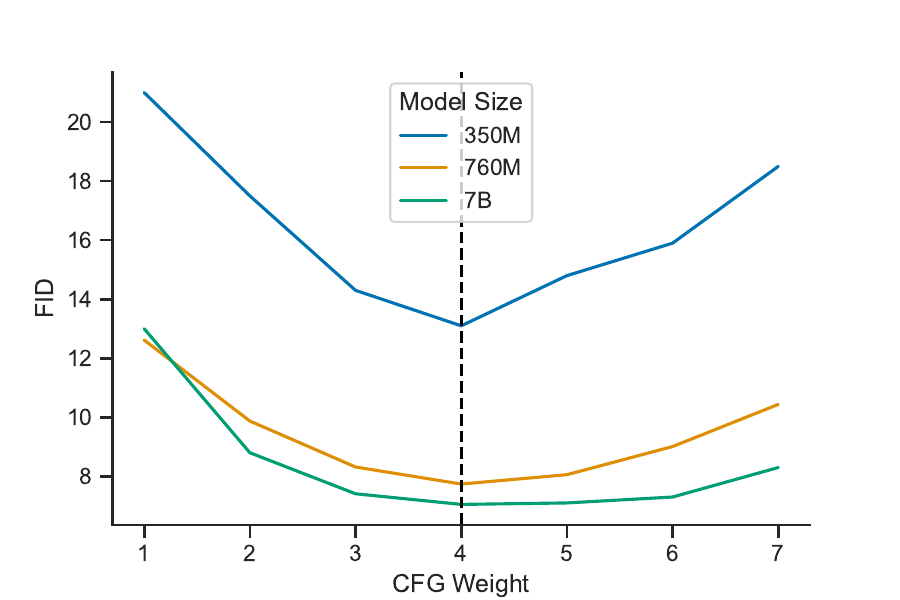}
  \end{minipage}
  \hfill
  \begin{minipage}[b]{0.49\textwidth}
    \centering
    \includegraphics[width=\textwidth]{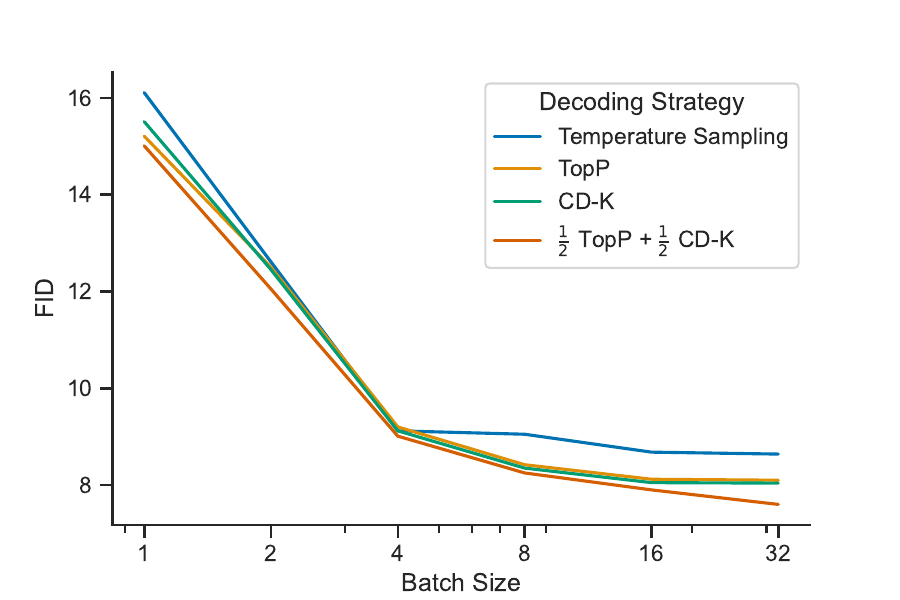}
  \end{minipage}
  \caption{ \textbf{(Left)} Comparison of Classifier-Free Guidance (CFG) weight and FID on 8k held-out MS-COCO data across our series of models. The optimal CFG remains consistent across all model sizes. \textbf{(Right)} Comparison of the number of generated samples per prompt before CLIP-based re-ranking and their respective FID. The data shows that TopP and CD-K are similar across sample counts but exhibit complementary behavior when combined.}\label{fig:cfg_dec_strat}
\end{figure}

In Figure~\ref{fig:cfg_dec_strat} we show that CD-K is competitive with standard CFG based sampling while providing a complementary set of generations to CFG allowing us to continue minimizing FID as we increase number of generations (while both CD-K and CFG independently stagnate).

\subsection{Quantitative Evaluations}

\begin{table}[h!]
\centering\small
\begin{tabular}{@{}lrrrrrrrrr@{}}
\toprule
& \makecell{Retrieval in\\Training}  & Responsible &\makecell{\# of Retrieved\\Documents}  & Dataset Size & Model Size & \makecell{Zero-shot\\FID-30K} \\ \midrule
RA-CM3 & \cmark & \xmark & 2 & 150M & 2.7B & 15.70 \\ 
StableDiffusion & \xmark  & \xmark  & - & 400M & 800M & 12.60 \\
KNN-Diffusion & \cmark & \xmark & 10 & 70M & 400M & 12.50 \\
MUSE & \xmark & \xmark & - & 500M & 3B & 7.88 \\
PARTI & \xmark & \xmark & - & 5B & 20B & 7.23 \\
RE-IMAGEN & \cmark & \xmark  & 2 & 450M & 3.6B & 5.25 \\
\midrule
\model{}-7B   & \cmark & \cmark   & 0 & 340M & 7B & 10.82 \\ 
\model{}-7B   & \cmark & \cmark   & 1 & 340M & 7B & 5.78 \\ 
\midrule
\model{}-350M & \cmark & \cmark& 2  & 340M & 350M & 14.20 \\
\model{}-760M & \cmark & \cmark & 2 & 340M & 760M & 6.61 \\
\model{}-7B & \cmark & \cmark & 2  & 340M & 7B & \textbf{4.88} \\ \bottomrule
\end{tabular}
\caption{Summary of various text-to-image models on the zero-shot MS-COCO task as measured by FID. For all of our models, we generate 8 samples for each input query, and use a CLIP model to select the best generation.}
\label{tab:models-summary}
\end{table}

Table~\ref{tab:models-summary} and Figure~\ref{fig:how_we_scale} provide a comparative overview of CM3Leon and state-of-the-art text-to-image models, evaluated based on the zero-shot MS-COCO (30K) task using the Fréchet Inception Distance (FID) metric \citep{FID_implementation}. CM3Leon-7B model set's a new state-of-the-art FID score of 4.88, while only using a fraction of the training data and compute of other models such as PARTI.

This observation underlines the effectiveness of retrieval-augmented decoder-only models like CM3Leon. In particular, the CM3Leon-7B model, when operated with one or two retrieved examples during inference, records superior FID scores. This result demonstrates the crucial role retrieval plays in expanding the world knowledge provided to the model and its capacity to generate high-quality images. CM3Leon surpasses all other retrieval-augmented models, including KNN-diffusion and RE-IMAGEN.

\section{Supervised Fine-Tuning}
Supervised fine-tuning (SFT) is critical in training large language models (LLMs) like ChatGPT. Despite this, its application in \mm{} settings remains largely unexplored. SFT trains a model to better understand of future instructions or prompts, enhancing its performance in novel and even zero-shot tasks. We have found that instruction tuning notably amplifies \mm{} model performance across various tasks such as image caption generation, visual question answering, text-based editing, and conditional image generation. 

\begin{figure}
    \centering
    \includegraphics[width=\linewidth]{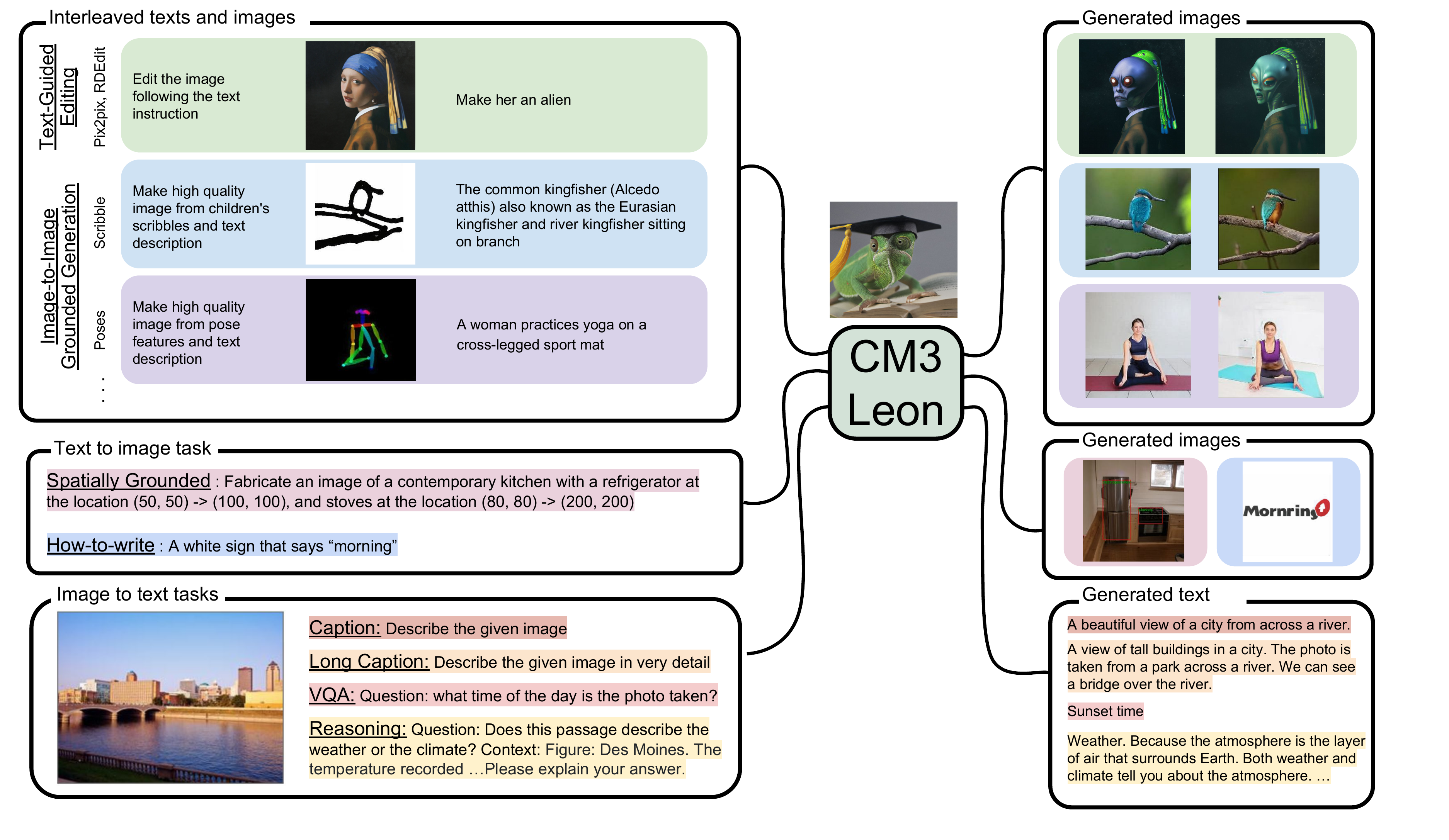}
    \caption{We perform fine-tuning on the \model{} model using a vast assortment of combined image and text tasks. Our retrieval augmented pretraining allows us to fine-tune the model effectively on a mixture of interleaved texts and images, as well as text-to-image and image-to-text tasks. We present some common model inputs for various tasks on the left, with the corresponding model outputs displayed on the right. Throughout the training process, we concatenate the model input and output and train them using the same objective that was utilized during the pretraining stage.}
    \label{fig:sft}
\end{figure}

We fine-tune \model{} on a wide array of mixed image and text tasks. We organized each task as a series of interleaved text and image examples, as shown in Figure~\ref{fig:sft}. 
The fine-tuning process follows the pretraining stage, employing the same CM3 objective by combining the task instruction with the output. Further details about the hyperparameters and scale of the SFT can be found in Section~\ref{sec:sft_hp}.

\subsection{Instructable Image Generation}
\paragraph{Text-Guided Image Editing} allows the modification of an initial image based on text instructions, with changes such as seasonal and weather adjustments, background changes, and material alterations. We used InstructPix2Pix methodology and proprietary face-filtering techniques on their data, yielding around 600,000 examples \citep{instructpix2pix}.
\paragraph{Image-to-Image Grounded Generation} involves producing grounding images with various features and text prompts. Features like edge maps, segmentation maps, key points, and human poses can be derived from user-uploaded images or sketches. We used ControlNet processing code on Shutterstock datasets to curate 7 million examples with features like canny edge, hed boundary, user sketching, human pose, and more \citep{controlnet}.
\paragraph{Spatially Grounded Image Generation} allows the user to integrate spatial information into text prompts for image generation, with each object represented by discrete tokens. We used object detection datasets like MS-COCO, Openimage, and Object365 to compile 3 million training examples\citep{mscoco,openimage,object365}.
\paragraph{How-to-write} task enables users to request the model to create signs or logos based on text prompts. We used an OCR detector to find suitable examples from Shutterstock datasets, resulting in 200,000 examples.

\begin{figure}[h]
    \centering
    \includegraphics[width=\linewidth]{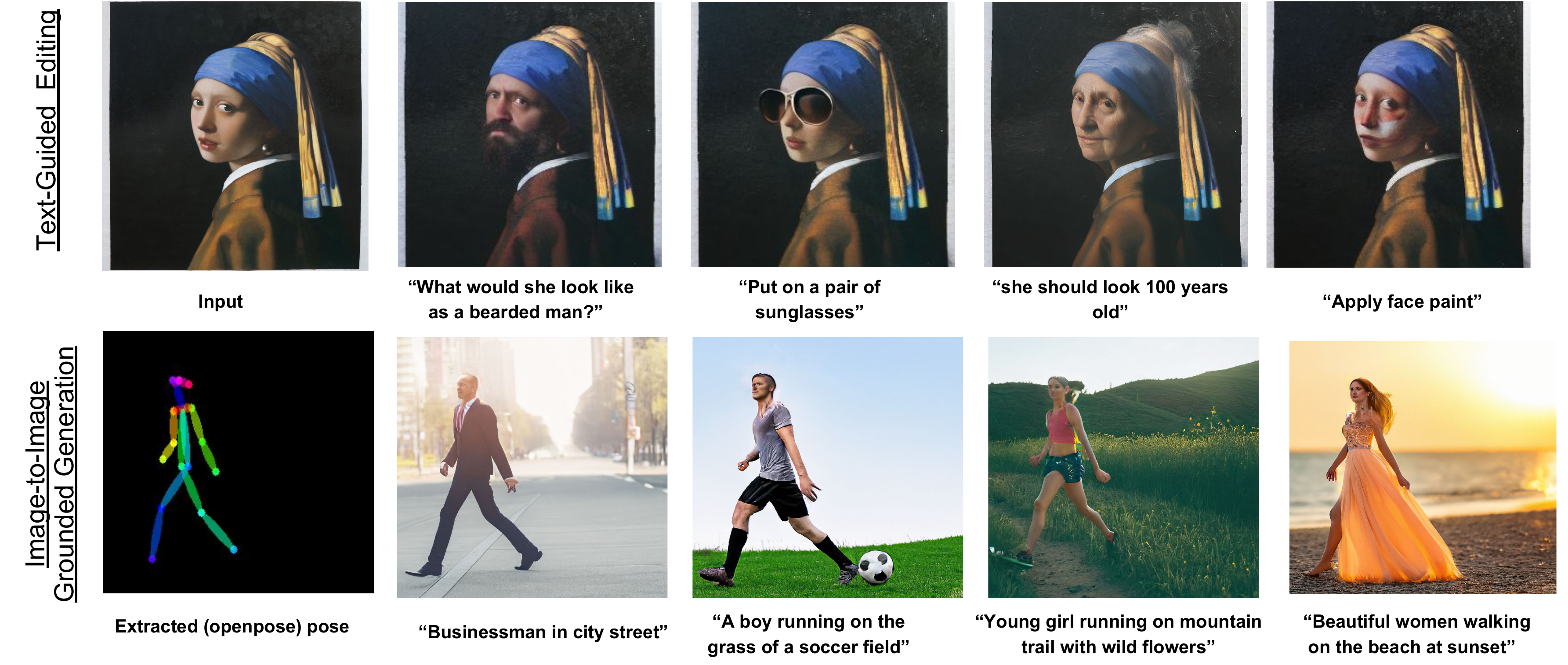}
     \caption{Qualitative examples of finetuned CM3Leon-7B model.
    \label{fig:sft_image}
    }
\end{figure}

\paragraph{Results:} We showcase qualitative examples of images produced by a fine-tuned CM3Leon-7B model, as depicted in Figure~\ref{fig:sft_image}. All instances in text-guided editing and image-image-grounded generation utilize a task prefix. For instance, we precede every text-guided editing example with the phrase, "Edit the image following the text instruction," and every scribble generation example with "Create a high-quality image from children's scribble and text description," amongst others. The top row of Figure~\ref{fig:sft_image} presents text-guided image examples. We employ separate image CFG (1.5) and text CFG (7.5) values during decoding. This approach is crucial for producing edited images that mirror the original image and closely align with the text editing instruction. The second row in Figure~\ref{fig:sft_image} show Structure-Guided Image Editing examples. For decoding, we utilized a single CFG value of 3. Given identical input open pose features, our model can generate markedly distinct images that follow different text prompts while maintaining the same pose as in the input image. More examples in ~\ref{fig:MORE_sft_image}

\subsection{Conditional Text Generation}
We also include several vision-language tasks to teach \model{} to respond in text to various kinds of textual prompts conditioned on an image, such as visual question answering, long-form captioning, etc. We use the following 8 vision-language tasks: MS-COCO~\citep{chen2015microsoft}, Flickr30k~\citep{young2014image}, Image Paragraph~\citep{krause2017hierarchical}, Localized Narratives~\citep{pont2020connecting}, VQA2~\cite{goyal2017making}, VizWiz~\citep{gurari2018vizwiz}, OKVQA~\citep{marino2019ok}, and ScienceQA~\citep{lu2022learn}. We use multiple prompt templates for each task to make the model robust to prompt variations (more details on the templates in Table~\ref{tab:sft-datasets} of the Appendix).

\begin{figure}[h]
    \centering
    \includegraphics[width=0.75\linewidth]{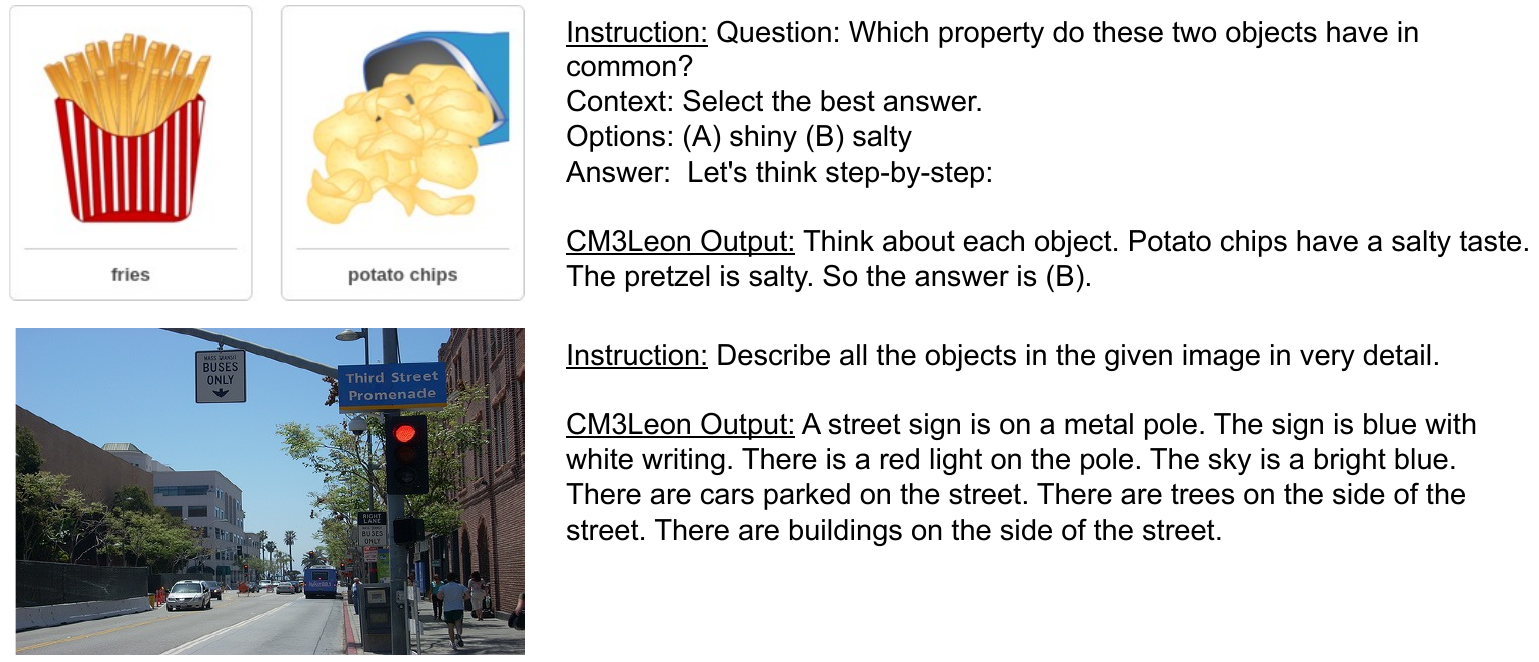}
     \caption{Qualitative examples showing our SFT-\model{}-7B model's generations for various long form generation tasks. 
    \label{fig:long_text_generation}
    }
\end{figure}

\paragraph{Results:}
Table~\ref{tab:vision-language-results} presents the performance comparison of our SFT-\model{} model w.r.t. previous state-of-the-art (SoTA) such as Flamingo~\citep{alayrac2022flamingo} and OpenFlamingo\footnote{\url{https://laion.ai/blog/open-flamingo/}}. We show that our SFT-\model{} model achieves strong zero-shot performance on several vision-language tasks even though they saw significantly fewer text data ($\approx$ 3B tokens) compared to Flamingo (100B tokens) and OpenFlamingo (40B tokens). Notably, SFT-\model{} even beats Flamingo on the VizWiz task. Figure~\ref{fig:caption_vqa_examples} presents our SFT-\model{}-7B model generations, given an image context and an instruction. The model is quite flexible with the instruction and can generate captions or answer a variety of questions. Further, the ability of to follow instructions is more evident in Figure~\ref{fig:long_text_generation} where the model can generate very long captions or reason over an image based on the given instruction. 

\begin{table}[h!]
\centering
\resizebox{\textwidth}{!}{
\begin{tabular}{@{}lcccccccc}
\toprule
Model & \makecell{MS-COCO \\ \small CIDEr \\ \small (test)} & \makecell{VQA2 \\ \small Acc. \\ 
 \small (test-dev)} & \makecell{VizWiz \\ \small Acc. \\ \small (test-dev)} & \makecell{OKVQA \\ \small Acc. \\ \small (val)} & \makecell{Image Paragraph \\ \small CIDEr \\ \small (test)} & \makecell{VisDial \\ \small NDCG \\ \small (val) } \\
\midrule
OpenFlamingo-9B$^\dagger$ (0-shot) & 65.5 & 43.5 & - & - & - & - \\
Flamingo-9B (0-shot) & 79.4 & 51.8 & 28.8 & 44.7 & - & 48.4 \\
\midrule
SFT-\model{}-7B (0-shot) & 61.6 & 47.6 & 37.6 & 23.8 & 10.5 & 22.6 \\ 
\bottomrule
\end{tabular}
}
\caption{Comparison of our supervised fine-tuning (SFT) \model{} with state-of-the-art models in zero-shot and few-shot settings. $^\dagger$ Reported numbers are all based on validation set.}
\label{tab:vision-language-results}
\end{table}

\section{Related Work}

\paragraph{Diffusion Models} Significant progress in the domain of text-to-image generation has been achieved through the use of diffusion models \citep{LDM, GLIDE, DALLE2}. The underlying mechanism involves sequentially adding noise to an image and then learning to reverse the noise based on provided text inputs or features \citep{unified_perspective_diffusion}. 
Diffusion models generally incorporate pretrained text or language representations such as the text encoder of the CLIP \citep{CLIP} image-text model or text encoders like T5 \citep{T5}. The recursive application of multi-resolution diffusion model (by employing multiple steps of super-resolution) has further enhanced their capability to generate high-quality images from text prompts, leading to state-of-the-art zero-shot non-retrieval based MS-COCO FID scores 

\paragraph{Retrieval Augmented Diffusion Models}
Conventional text-to-image models often struggle to capture the vast array of concepts and entities in the image domain. Methods like enabling retrieval during inference time can help address the complexity of these tail entities by delegating it to a retrieval step. Building on the work of \citet{IMAGEN}, \citet{REIMAGEN} incorporates retrieval to enhance zero-shot MS-COCO FID scores, demonstrating further improvement in this area.

\paragraph{Autoregressive Token Models}
Significant advancements have been made in the field by utilizing LLMs over tokenized image representations \citep{taming, DALLE}. A widely-used approach in the field \citep{vqvae,vqvae2,vqvae_gan} involves an initial stage of converting images into discrete latent variables through tokenization, which transforms a text-to-image generation problem into a sequence-to-sequence problem, thereby enabling subsequent application of LLM techniques \citep{DALLE, makeascene}.

\paragraph{Non-Autoregressive Token Models}
Although autoregressive models have benefited from extensive research in NLP, autoregressive decoding can be quite computationally expensive. Non-autoregressive models, such as \citet{maskpredict}, have been proposed in NLP and extended to text-to-image models, exemplified by \citet{MUSE} which achieves state-of-the-art image generation performance and higher efficiency than diffusion or autoregressive models by employing masked modeling in discrete token space (non-autoregressively with iterative decoding).

\paragraph{Retrieval Augmented Autoregressive Token Models}
Token-based models face challenges akin to those encountered by non-retrieval augmented diffusion models. To address these issues, \citet{RA_CM3} suggested prefixing decoder-only text-to-image models, such as \citet{DALLE, CM3}, with statically retrieved instances during training, resulting in significant efficiency gains during the training process. 

Our paper primarily concentrated on scaling this strategy.

\section{Conclusion}

We presented CM3Leon, a retrieval-augmented, token-based, decoder-only \mm\ language model that efficiently and flexibly generates and infills text and images. Our approach extends the scope of autoregressive models, demonstrating their potential to compete with and exceed diffusion models in terms of cost-effectiveness and performance.
By integrating a retrieval-augmented pretraining stage with a diverse, large-scale Shutterstock dataset and a second multi-task supervised fine-tuning stage, CM3Leon demonstrates the benefits of a comprehensive training approach. Further enhanced by an innovative, self-contained contrastive decoding method, our model offers improved text and image generation quality.
Our results support the value of autoregressive models for a broad range of text and image tasks, encouraging further exploration for this approach. 
\bibliography{arxiv}
\bibliographystyle{arxiv}

\appendix
\section{Showcase Prompts}\label{sec:showcase_prompts}
\begin{enumerate}
    \item Chameleon and octopus, side by side, high quality render, drawing, professional.
    \item A plush toy koala bear relaxing on a lounge chair and working on a laptop. The chair is beside a rose flower pot. There is a window on the wall beside the flower pot with a view of snowy mountains.
    \item A photo of an astronaut riding a horse in the forest. There is a river in front of them with water lilies.
    \item A teddy bear wearing a motorcycle helmet and cape is riding a motorcycle in Rio de Janeiro with Dois Irmãos in the background. dslr photo.
    \item A black german shepherd wearing a red beret
    \item An Armenian church on the surface of Mars, with Astronaut walking into the church, in Focus. Photo. Fantasy. Dramatic.
    \item Armenian khachkars surrounded by pomegranates in a bright green forest.
    \item A cat wearing sunglasses
    \item A small cactus wearing a straw hat and neon sunglasses in the Sahara desert.
    \item A close up photo of a human hand, hand model. High quality
    \item A raccoon main character in an Anime preparing for an epic battle with a samurai sword. Battle stance. Fantasy, Illustration
    \item A stop sign in a Fantasy style with the text "1991"
\end{enumerate}
\section{Pre-Training}
\subsection{Data Visualizations}\label{sec:data_visualizations}
\begin{figure}[h]
    \centering
    \includegraphics[width=\linewidth]{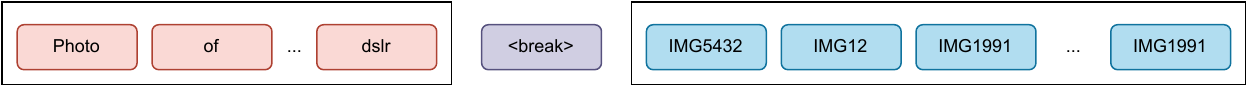}
    \caption{Visualization of the tokenization of one caption-image pair.}
    \label{fig:single_image_tokenization}
\end{figure}
\begin{figure}[h]
    \centering
    \includegraphics[width=\linewidth]{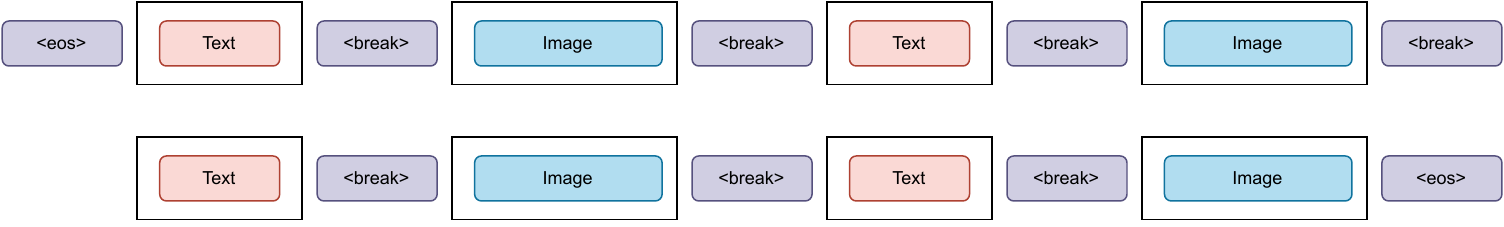}
    \caption{Visualization of the tokenization of a full training sample consisting of retrieved sampled and query caption-image pair.}
    \label{fig:multiple_image_tokenization}
\end{figure}
\subsection{Model Hyper-Parameters}
\begin{table}[h]
\centering\small
\begin{tabular}{l|cccccccc}
\toprule
Model size & \#~L & d$_{\text{model}}$ & Seq Length & Batch & LR & Warmup Steps & \#~GPUs & \#~Tokens \\
\midrule
350M   & 24  & 1024  & 4096      & 8M & 6e-04     & 1500  & 256  & 1.4T      \\
760M  & 24  & 1536  & 4096      & 8M & 5e-04     & 1500  & 256 & 1.9T    \\
7B    & 32  & 4096 & 4096      & 8M & 1.2e-04     & 1500   & 512 & 2.4T      \\
\bottomrule
\end{tabular}
\caption{{\bf Model architecture details.}  We report the
number of layers (\#~L), embedding size (d$_{\text{model}}$), sequence length, batch size, peak learning rate (LR), learning rate warmup steps, number of GPUs used, and number of tokens consumed by each model.}
\label{tab:pretraing_para}
\end{table}
\section{Inference Latency and Throughput}
\begin{figure}[htbp]
  \centering
  \begin{minipage}[c]{0.49\textwidth}
    \centering\small
    \begin{tabular}{lcc}
    \toprule
    Model & Resolution & Time \\
    \midrule
    Imagen & $256 \times 256$ & 9.1s  \\
    Imagen & $1024 \times 1024$ & 13.1s  \\
    LDM (50 steps) & $512 \times 512$ & 3.7s \\
    LDM (250 steps) & $512 \times 512$ & 18.5s \\
    Parti (3B) & $256 \times 256$ & 6.4s \\
    MUSE (3B) & $256 \times 256$ & 0.5s \\
    CM3Leon (7B, BF16) & $256 \times 256$ & 11.8s \\
    CM3Leon (7B, INT8) & $256 \times 256$ & 9.1s \\
    \bottomrule
    \end{tabular}
    \caption{\small Inference latency for several models.}
    \label{tab:inference-tine}
  \end{minipage}
  \hfill
  \begin{minipage}[c]{0.45\textwidth}
    \centering
    \includegraphics[width=\textwidth]{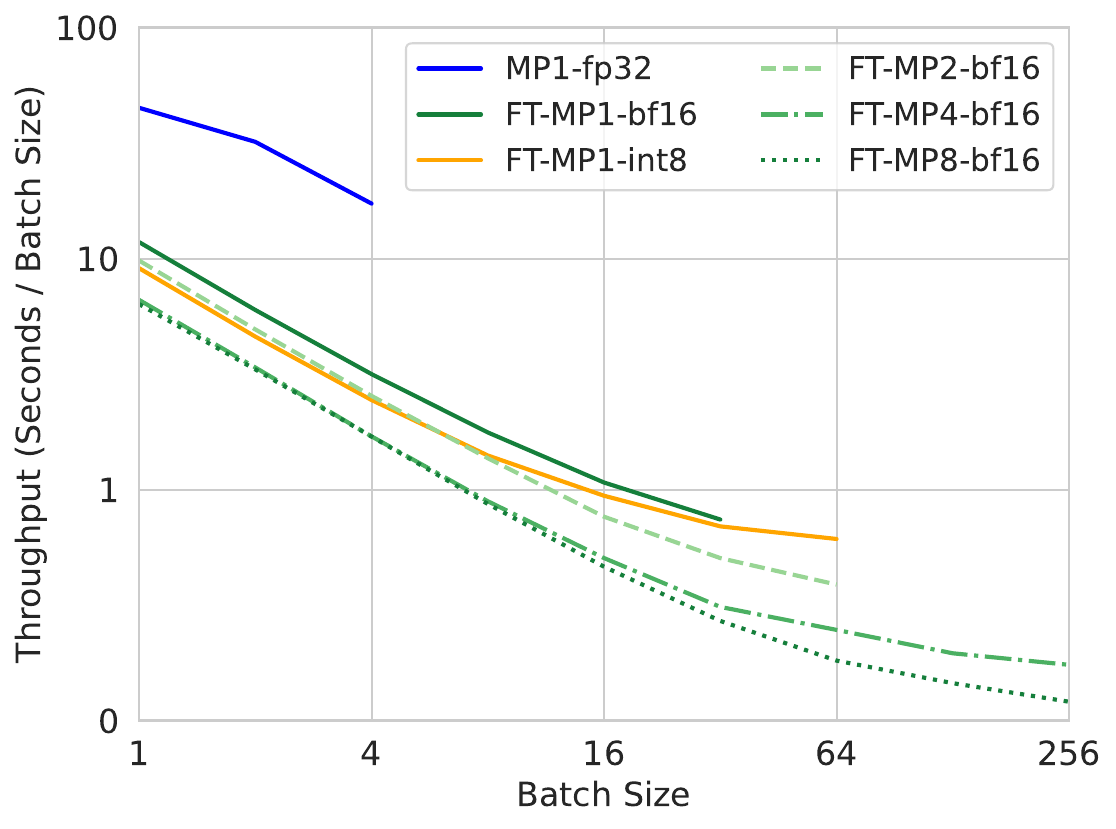}
  \caption{Inference throughput of \model{}-7B for generating images, without retrieval, across different model parallelism (MP), FasterTransformer (FT) implementation, data type (DType) and batch sizes}\label{fig:inference_perf}
  \end{minipage}
\end{figure}
\section{Image Generation}
\begin{figure}[H]
    \centering
    \begin{tabular}{cc}
        \begin{subfigure}{0.25\textwidth}
    \includegraphics[width=\textwidth]{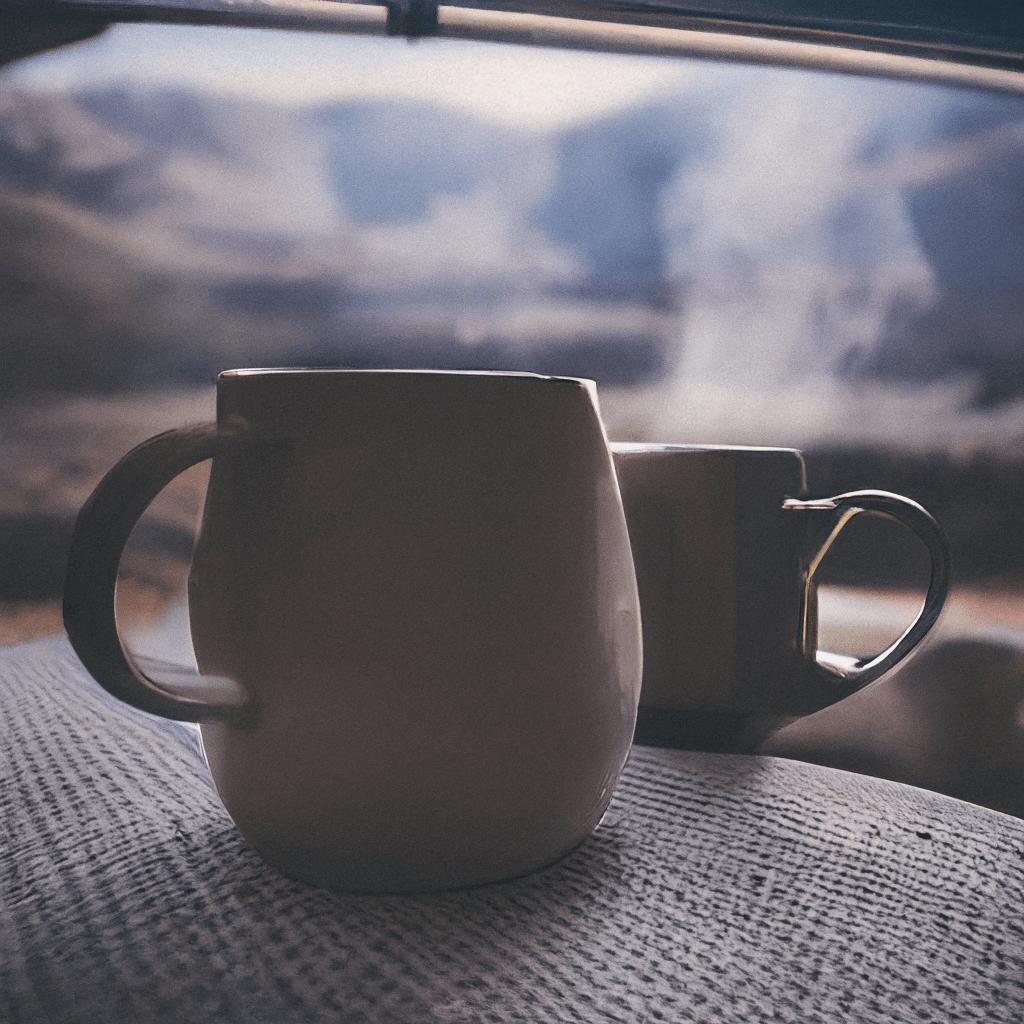}
        \end{subfigure}
        \begin{subfigure}{0.25\textwidth}
    \includegraphics[width=\textwidth]{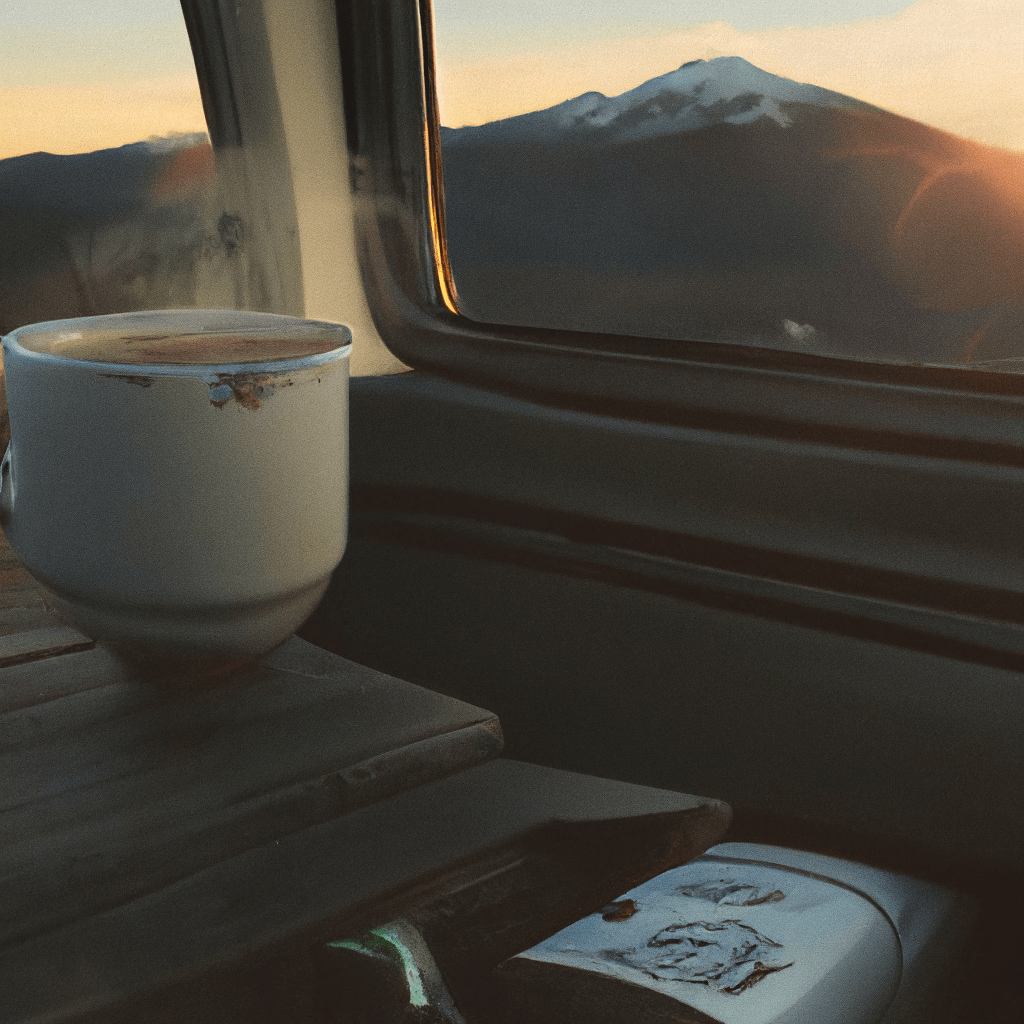}
        \end{subfigure}
        \begin{subfigure}{0.25\textwidth}
    \includegraphics[width=\textwidth]{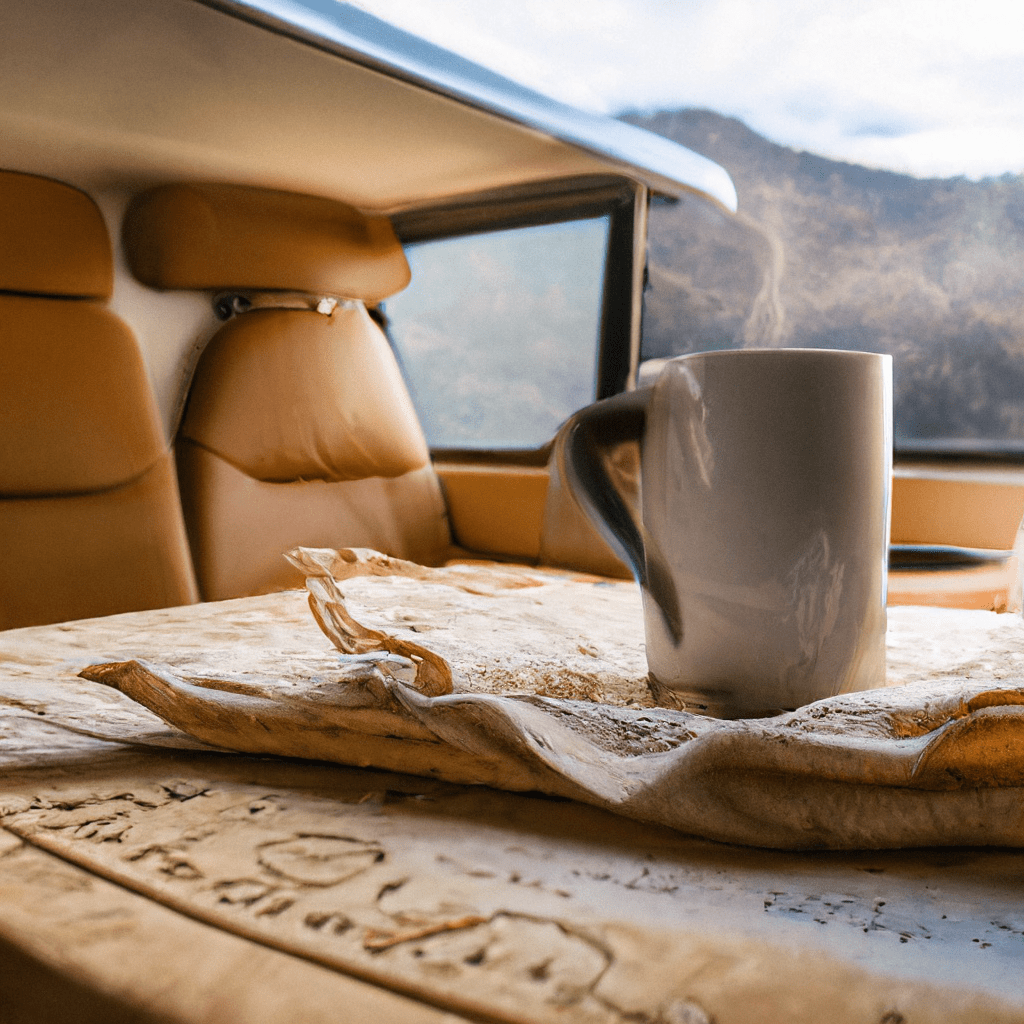}
        \end{subfigure}
        \begin{subfigure}{0.25\textwidth}
    \includegraphics[width=\textwidth]{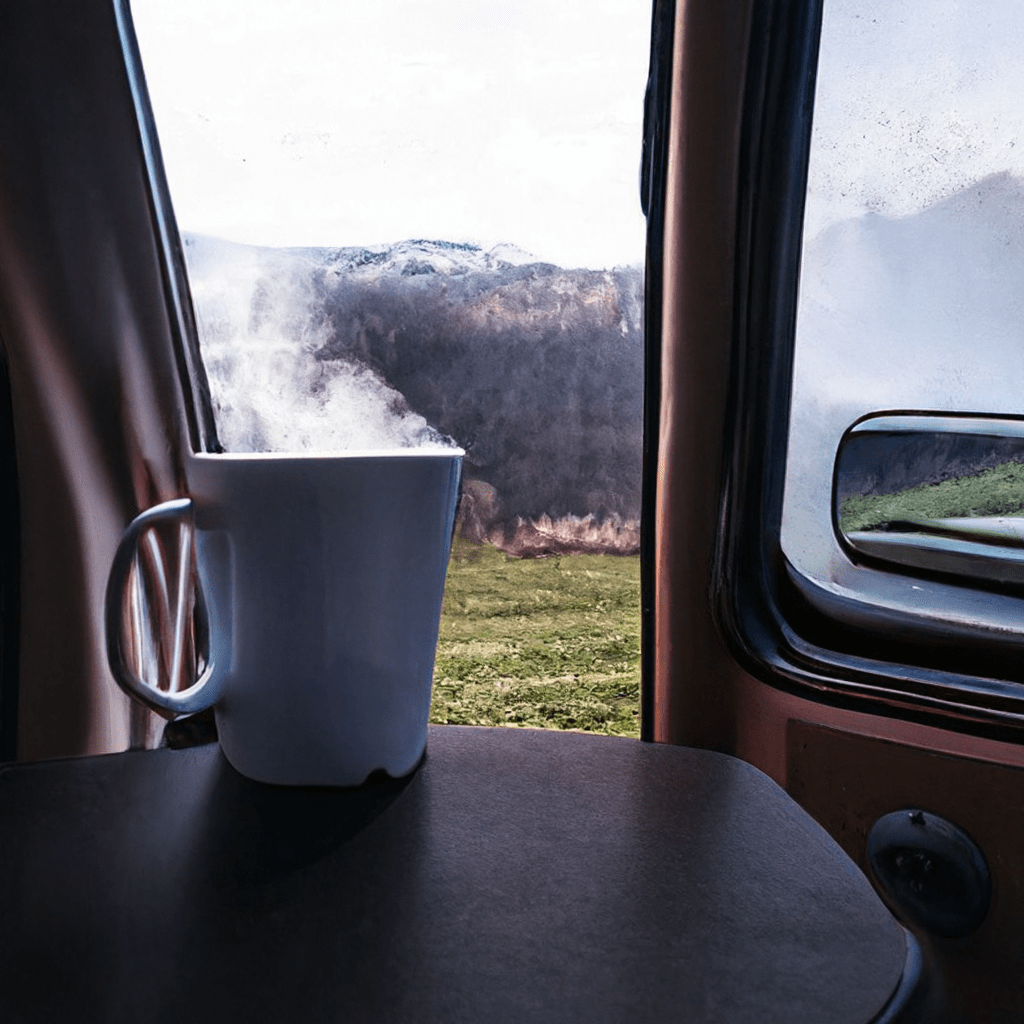}
        \end{subfigure}\\
        \begin{subfigure}{0.25\textwidth}
    \includegraphics[width=\textwidth]{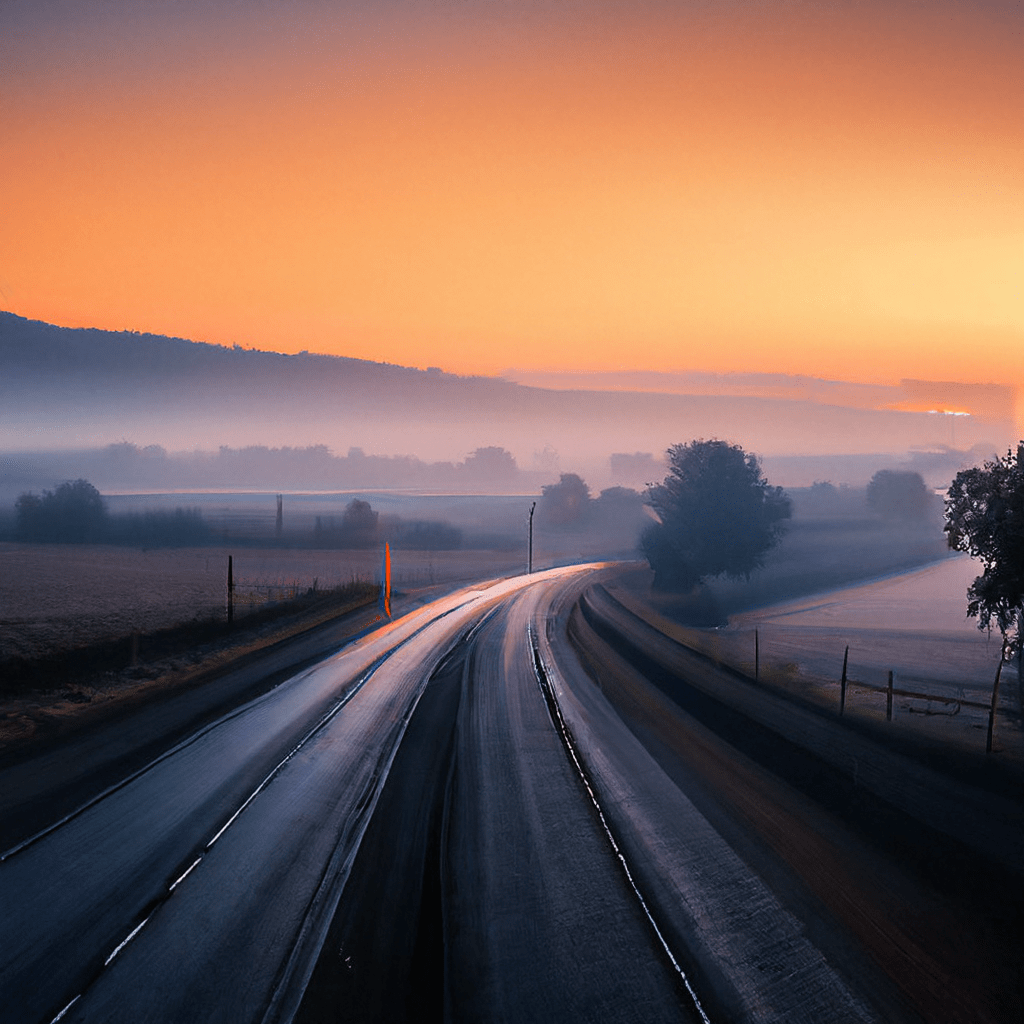}
        \end{subfigure}
        \begin{subfigure}{0.25\textwidth}
    \includegraphics[width=\textwidth]{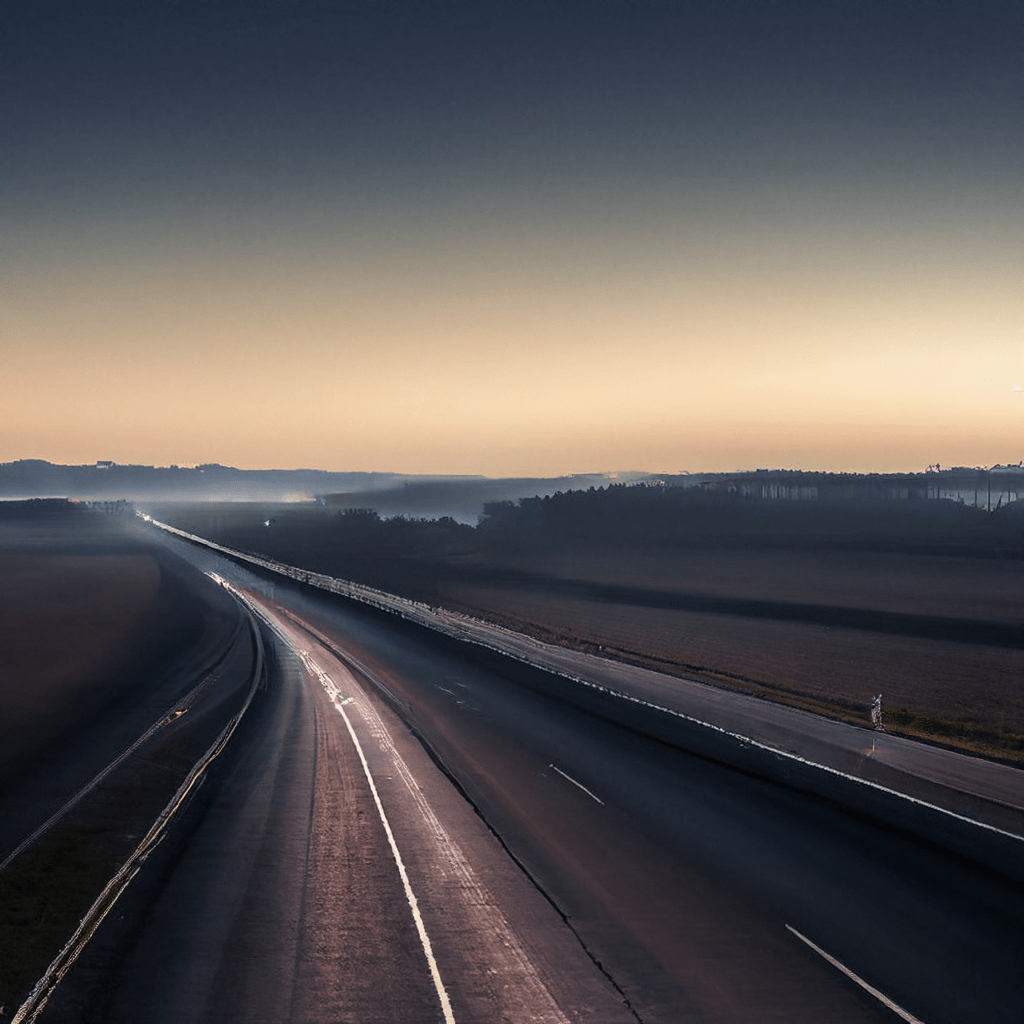}
        \end{subfigure}
        \begin{subfigure}{0.25\textwidth}
    \includegraphics[width=\textwidth]{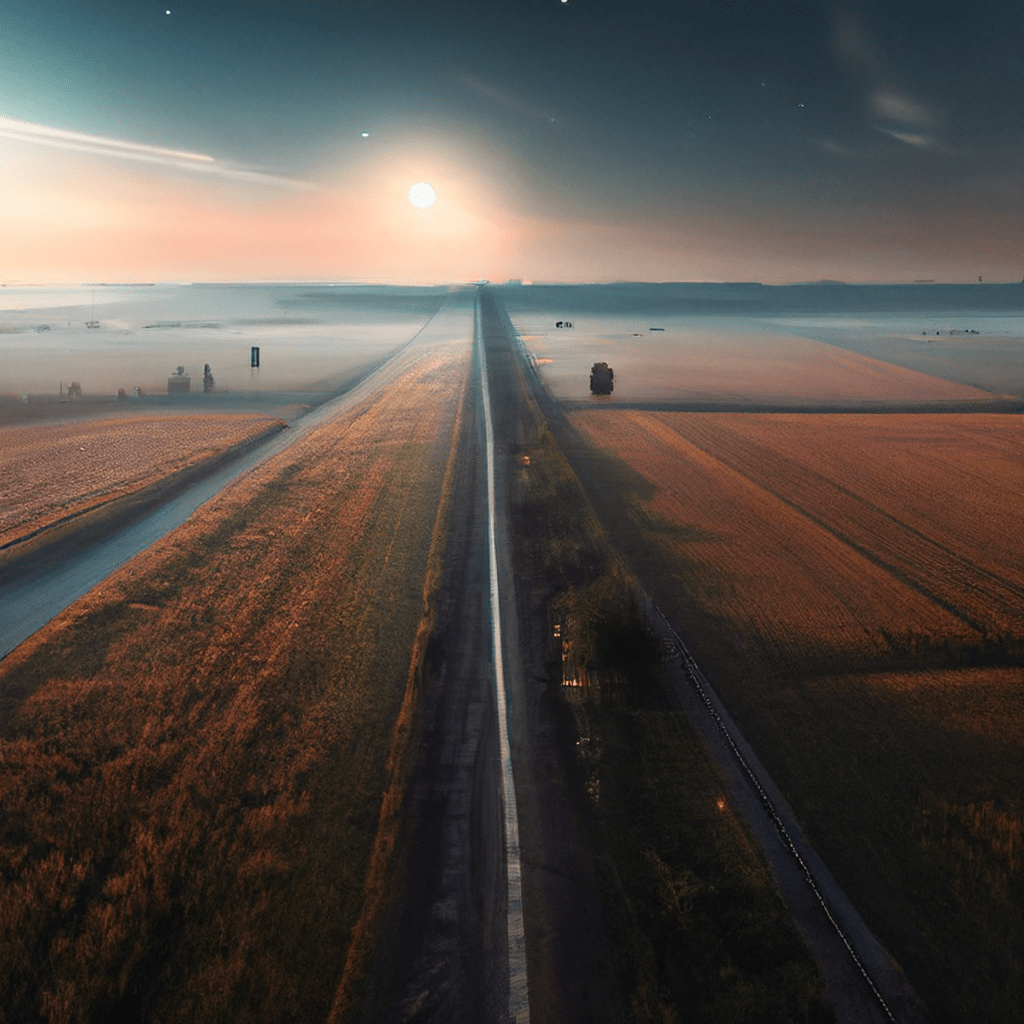}
        \end{subfigure}
        \begin{subfigure}{0.25\textwidth}
    \includegraphics[width=\textwidth]{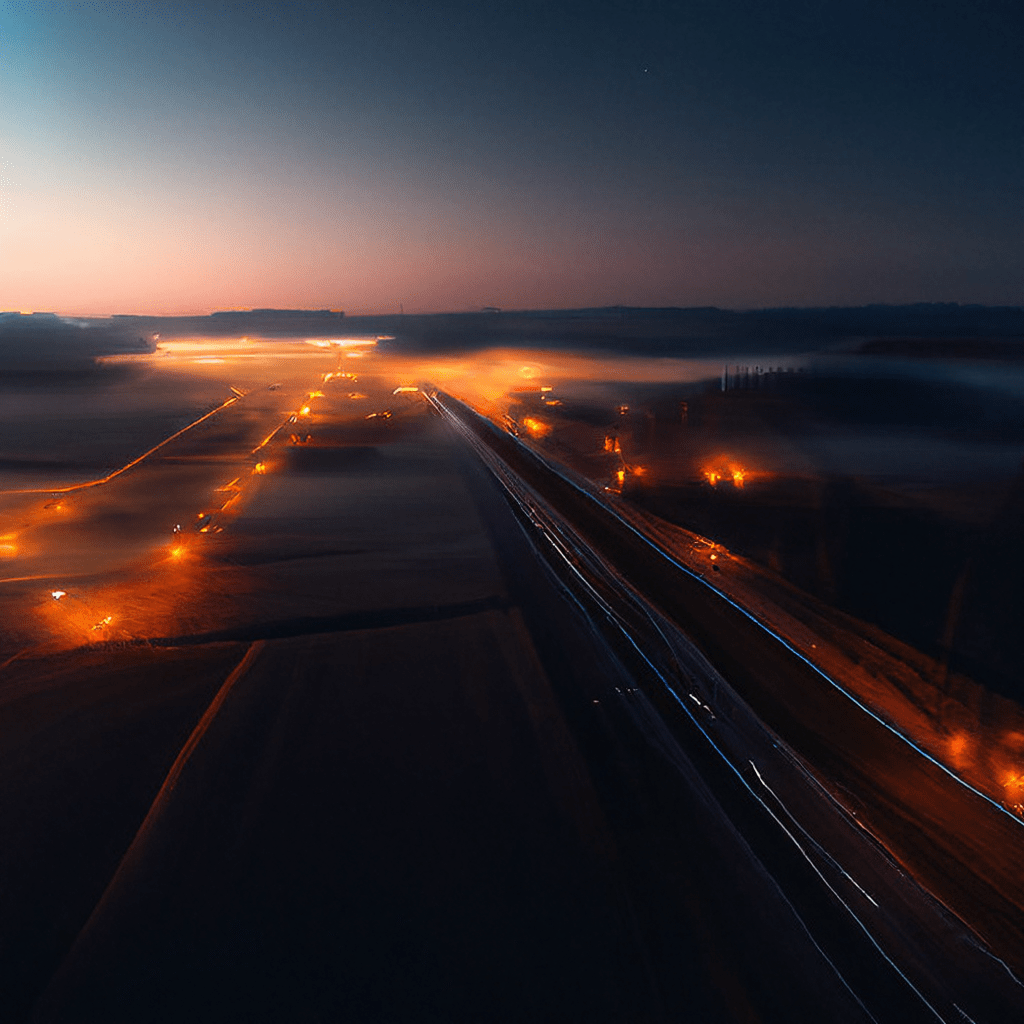}
        \end{subfigure}\\
        \begin{subfigure}{0.25\textwidth}
    \includegraphics[width=\textwidth]{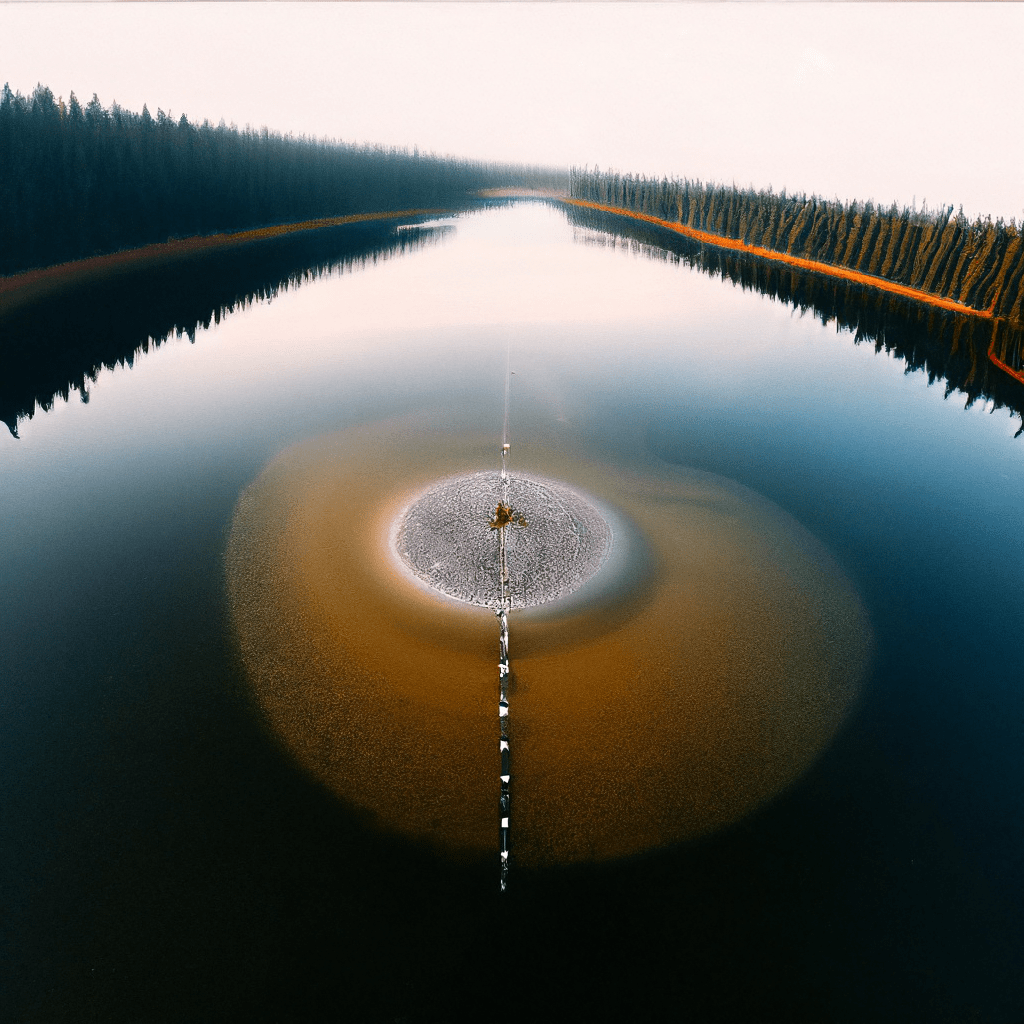}
        \end{subfigure}
        \begin{subfigure}{0.25\textwidth}
    \includegraphics[width=\textwidth]{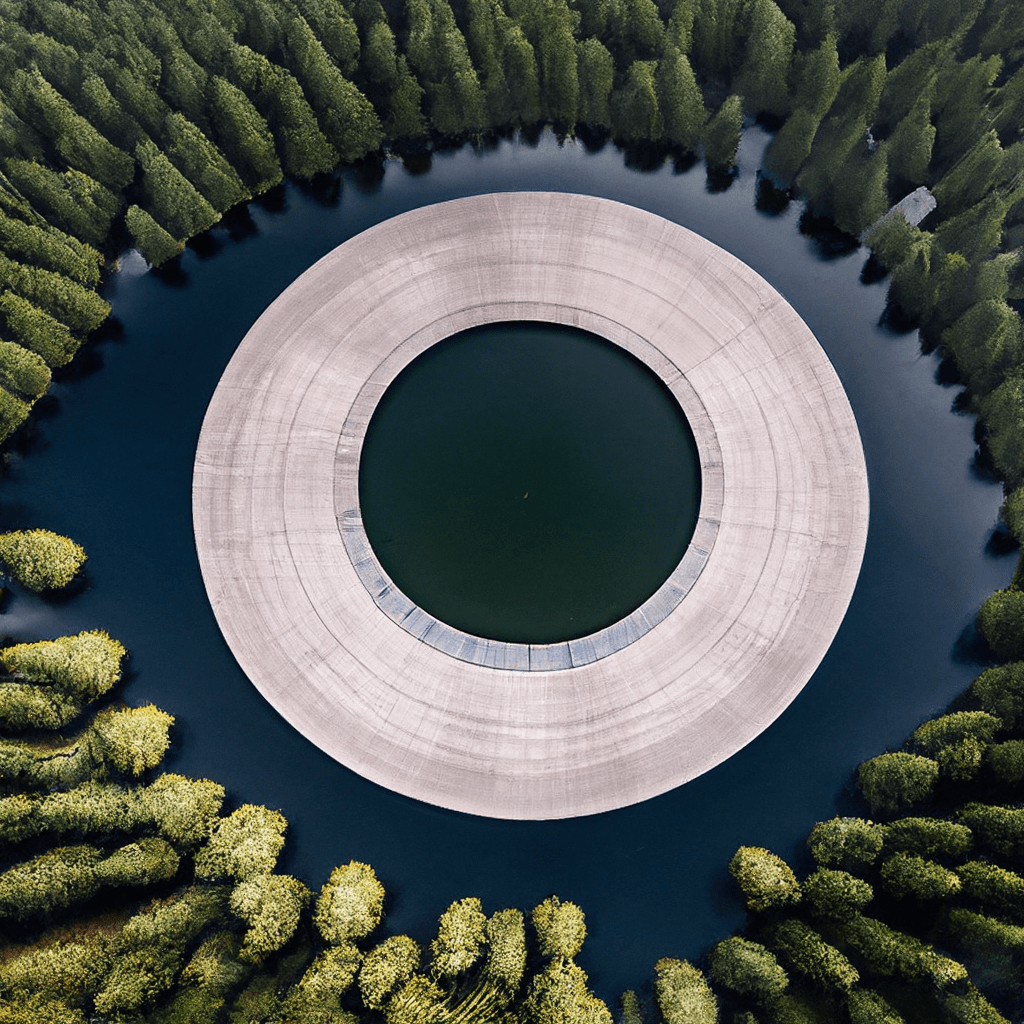}
        \end{subfigure}
        \begin{subfigure}{0.25\textwidth}
    \includegraphics[width=\textwidth]{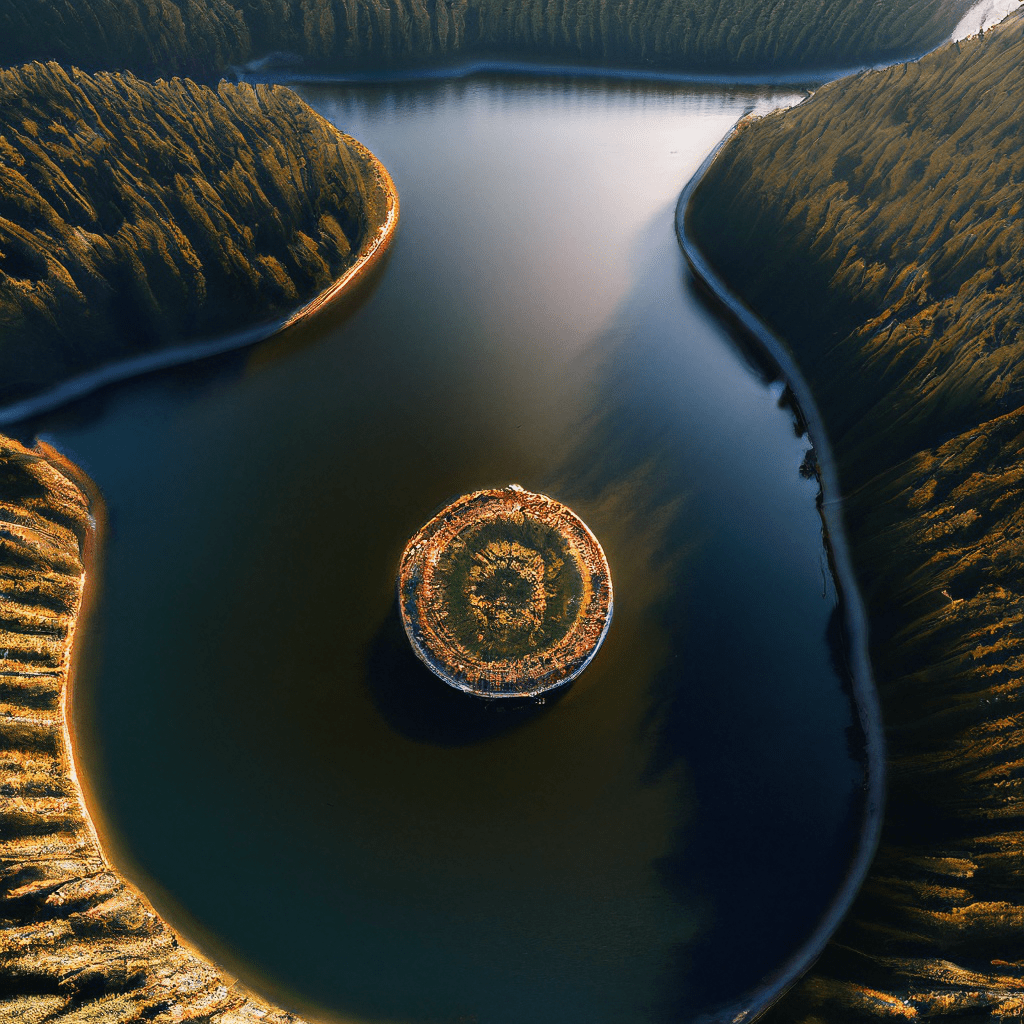}
        \end{subfigure}
        \begin{subfigure}{0.25\textwidth}
    \includegraphics[width=\textwidth]{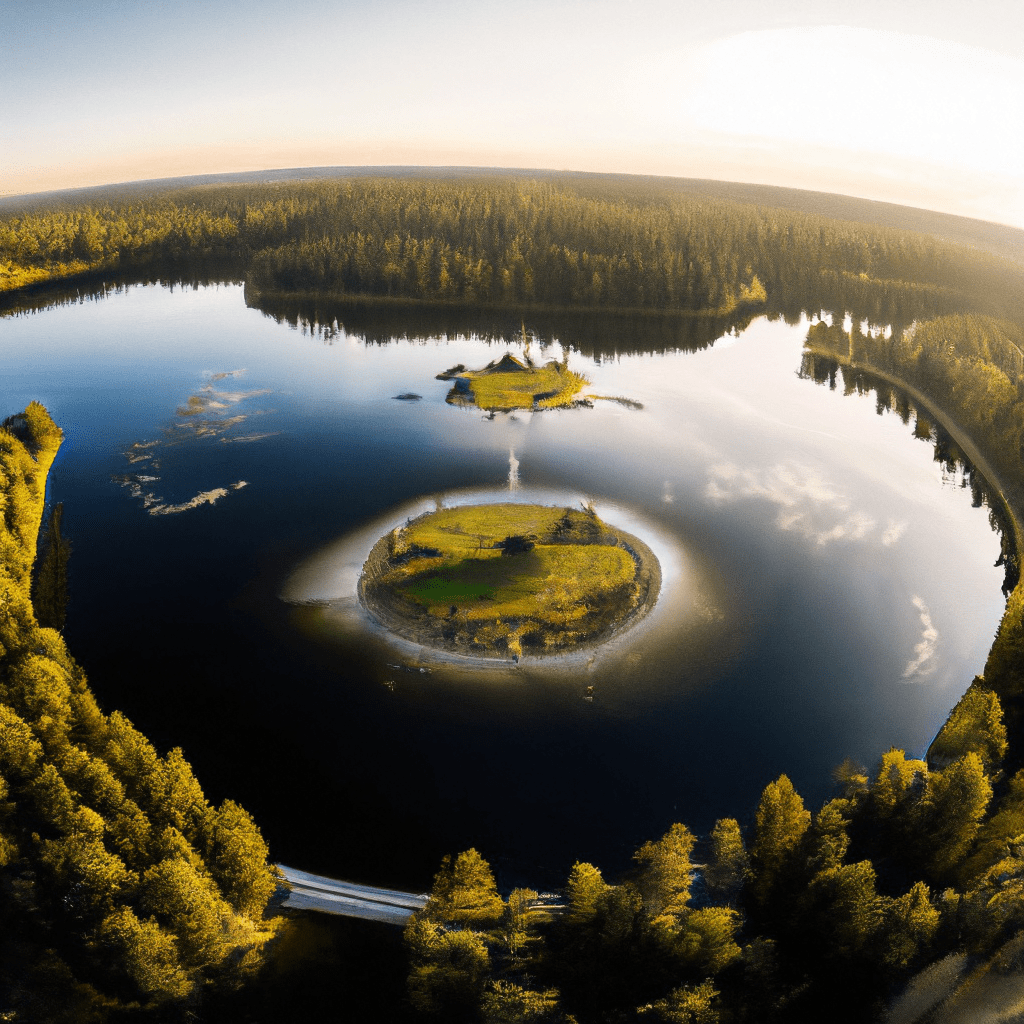}
        \end{subfigure}
    \end{tabular}
    \caption{Top to bottom prompts: \texttt{A steaming cup of coffee with mountains in the background. Resting during road trip.}, \texttt{beautiful, majestic road during sunset. Aesthetic.}, \textit{small circular island in the middle of a lake. Forests surrounding the lake. High Contrast.}}
\end{figure}

\begin{figure}[H]
    \centering
    \begin{tabular}{cc}
        \begin{subfigure}{0.25\textwidth}
    \includegraphics[width=\textwidth]{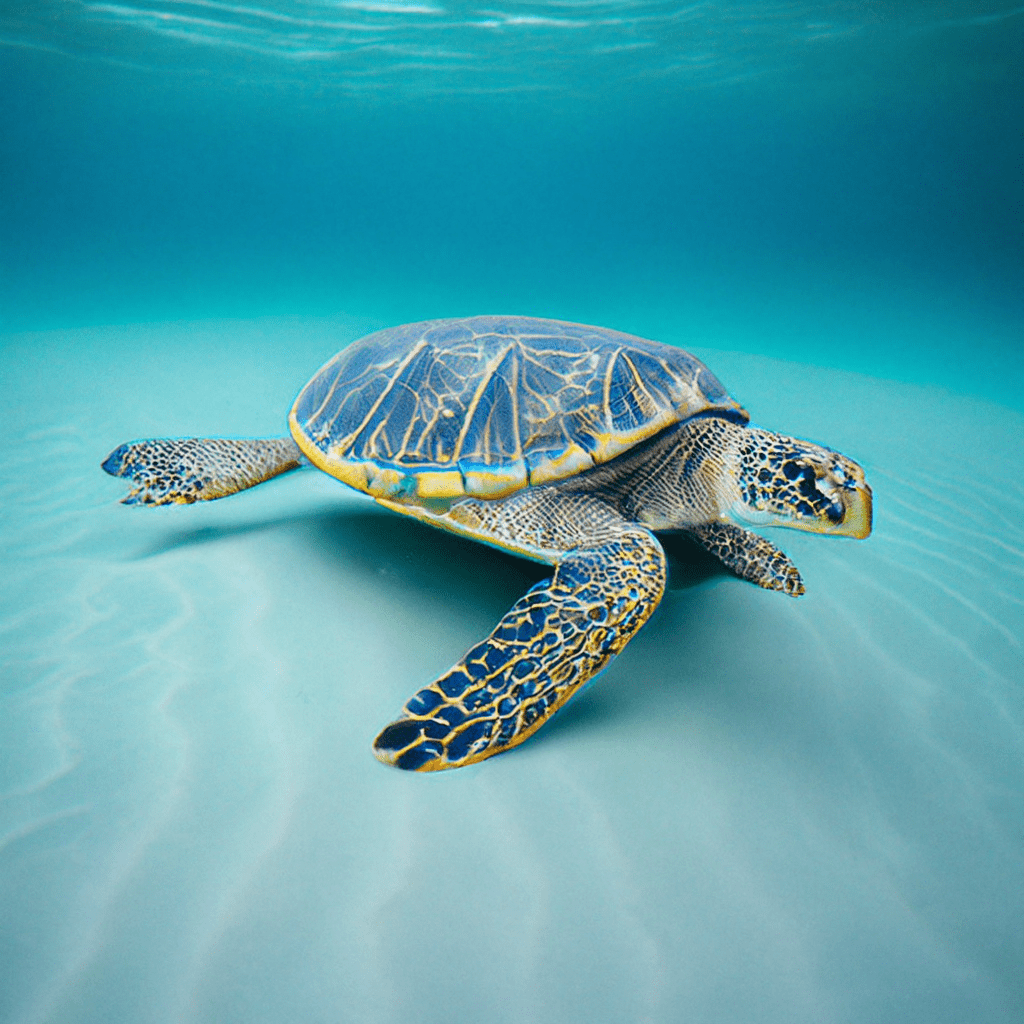}
        \end{subfigure}
        \begin{subfigure}{0.25\textwidth}
    \includegraphics[width=\textwidth]{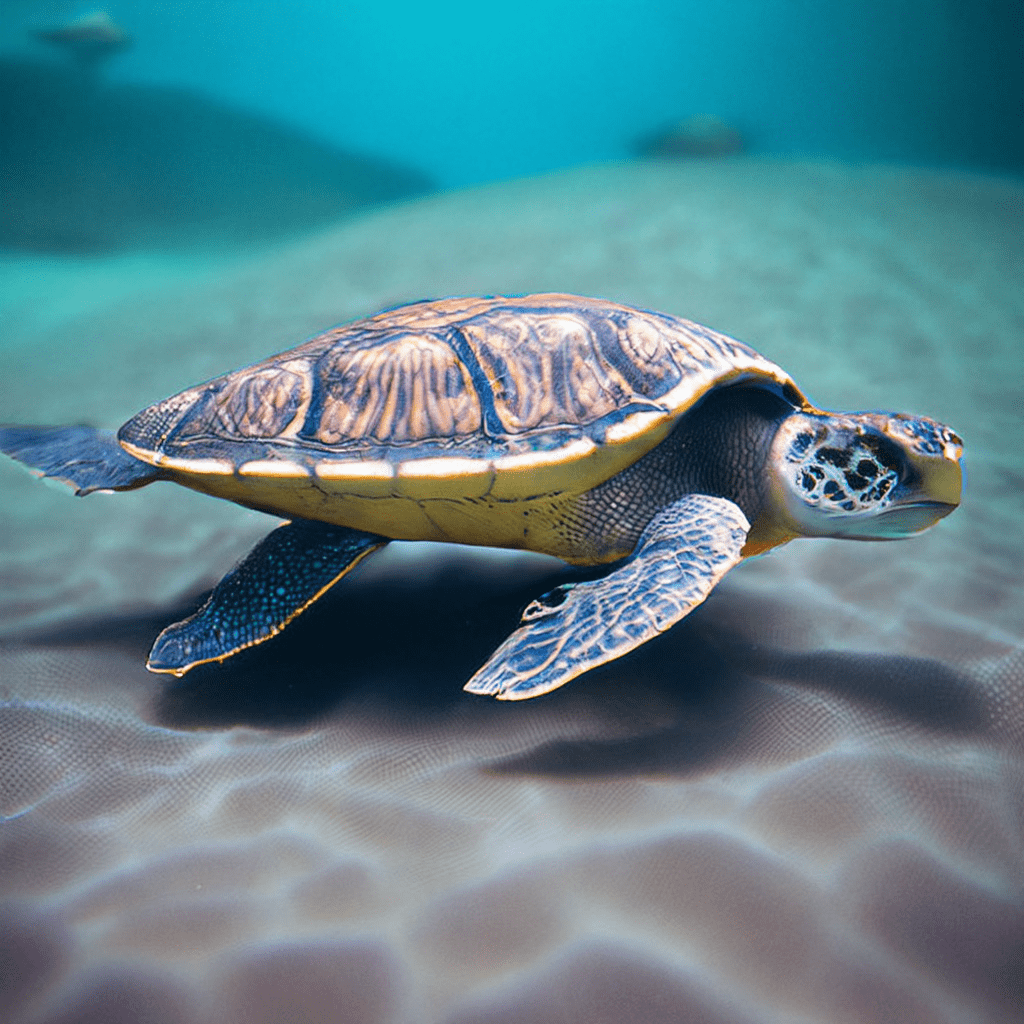}
        \end{subfigure}
        \begin{subfigure}{0.25\textwidth}
    \includegraphics[width=\textwidth]{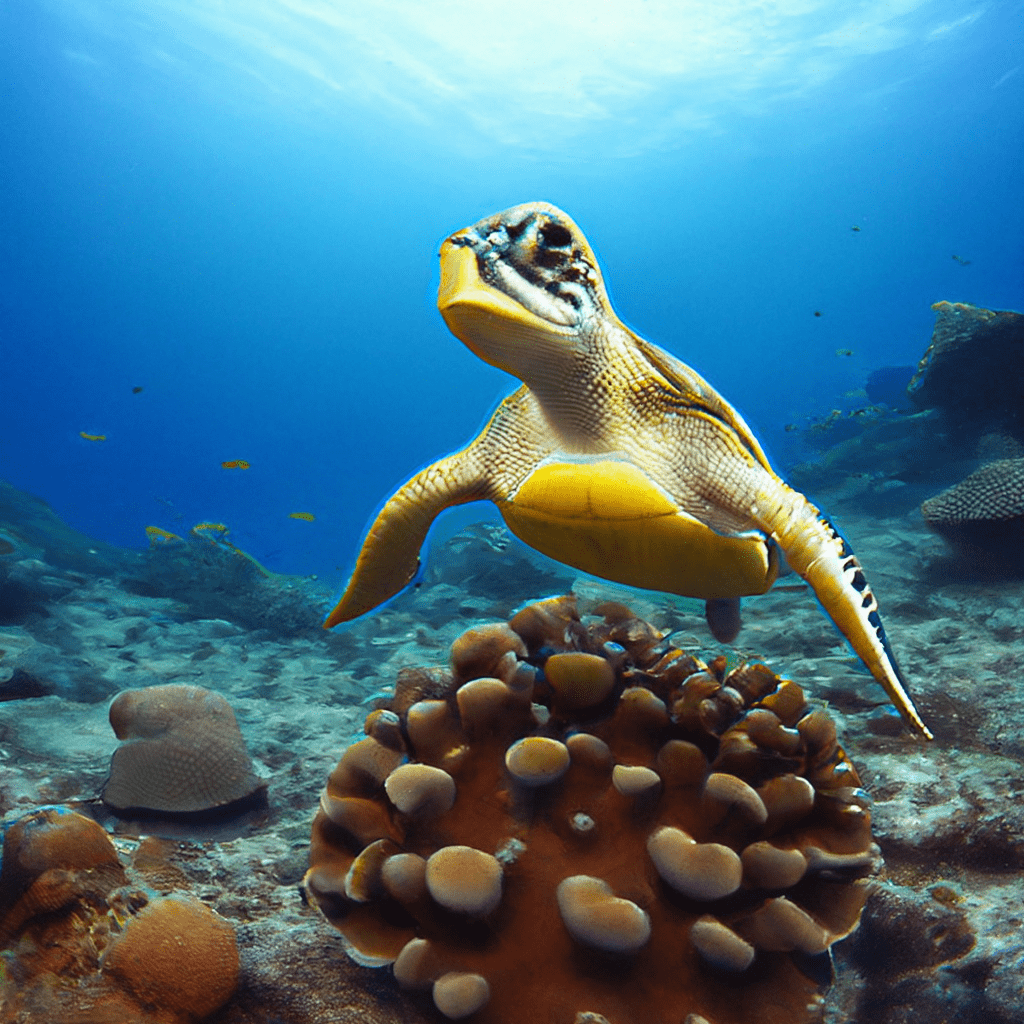}
        \end{subfigure}
        \begin{subfigure}{0.25\textwidth}
    \includegraphics[width=\textwidth]{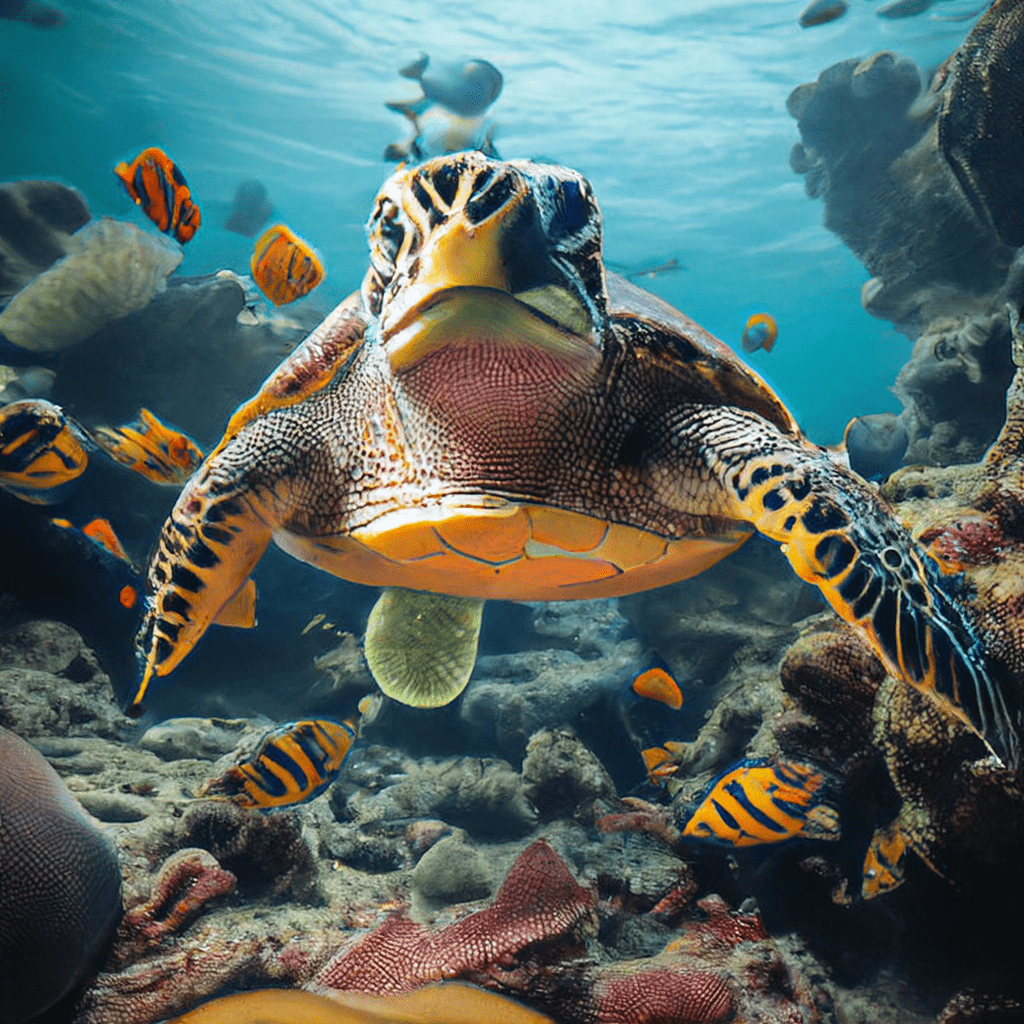}
        \end{subfigure}\\
        \begin{subfigure}{0.25\textwidth}
    \includegraphics[width=\textwidth]{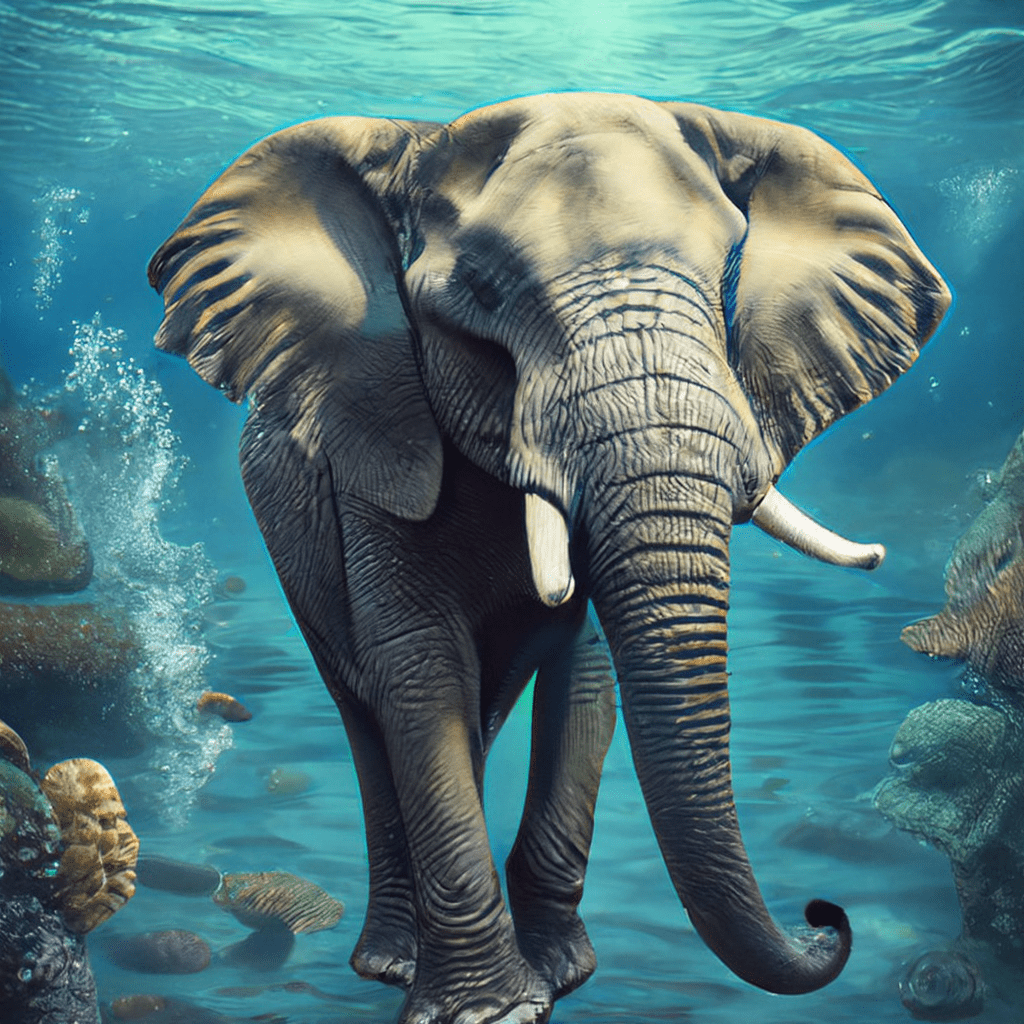}
        \end{subfigure}
        \begin{subfigure}{0.25\textwidth}
    \includegraphics[width=\textwidth]{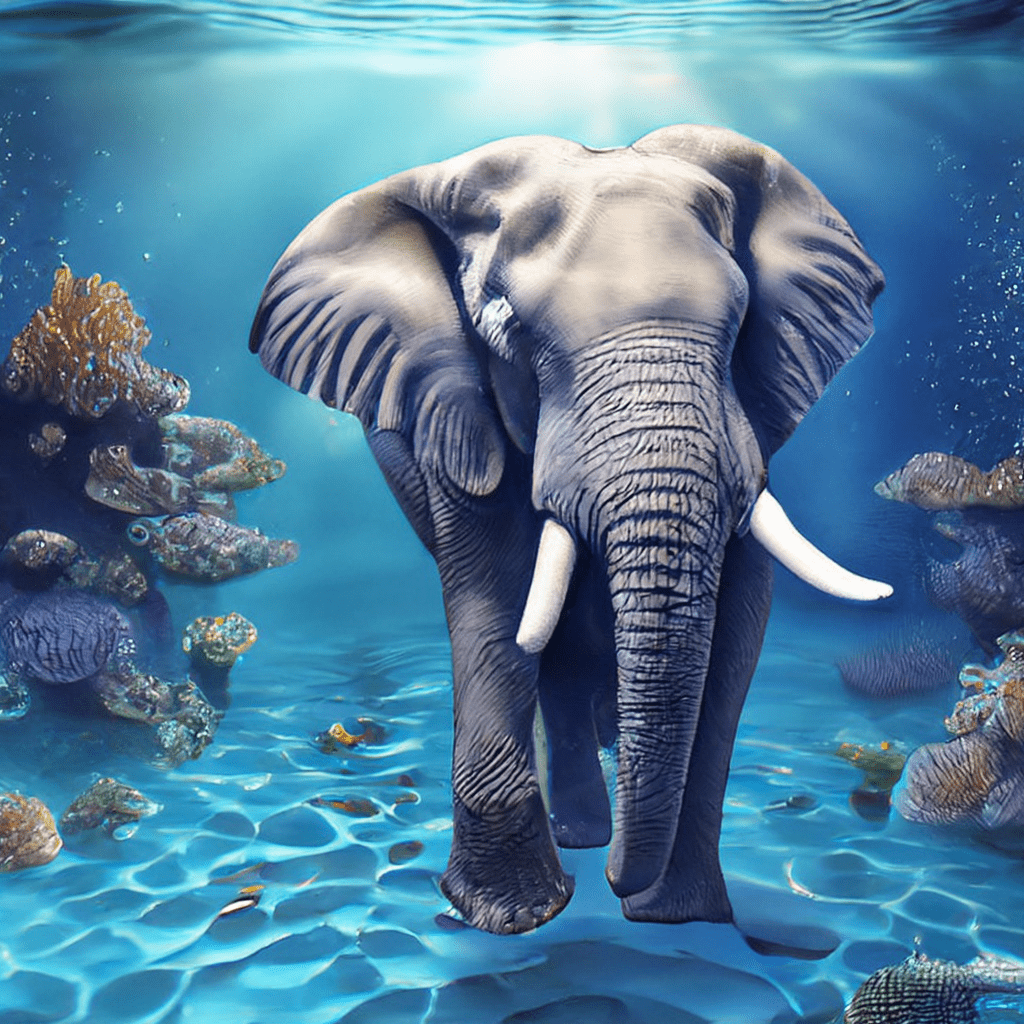}
        \end{subfigure}
        \begin{subfigure}{0.25\textwidth}
    \includegraphics[width=\textwidth]{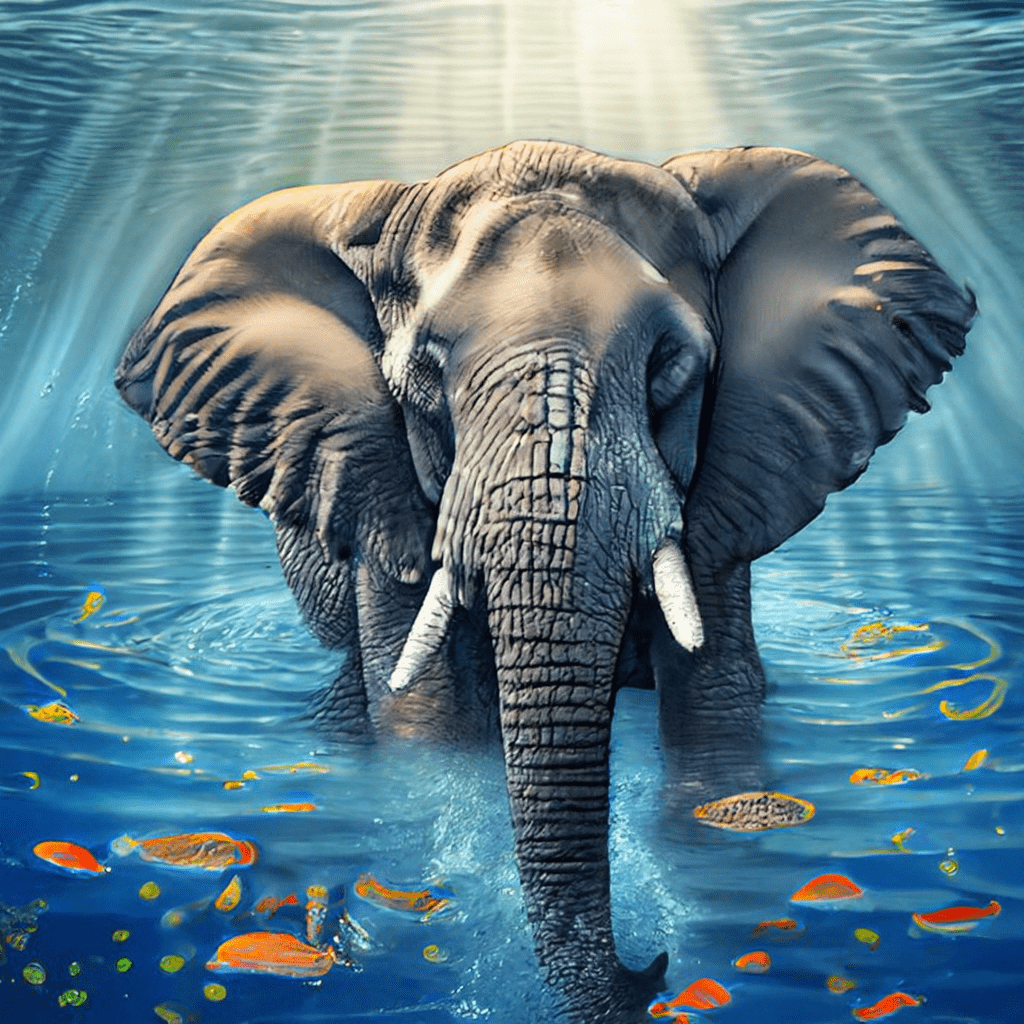}
        \end{subfigure}
        \begin{subfigure}{0.25\textwidth}
    \includegraphics[width=\textwidth]{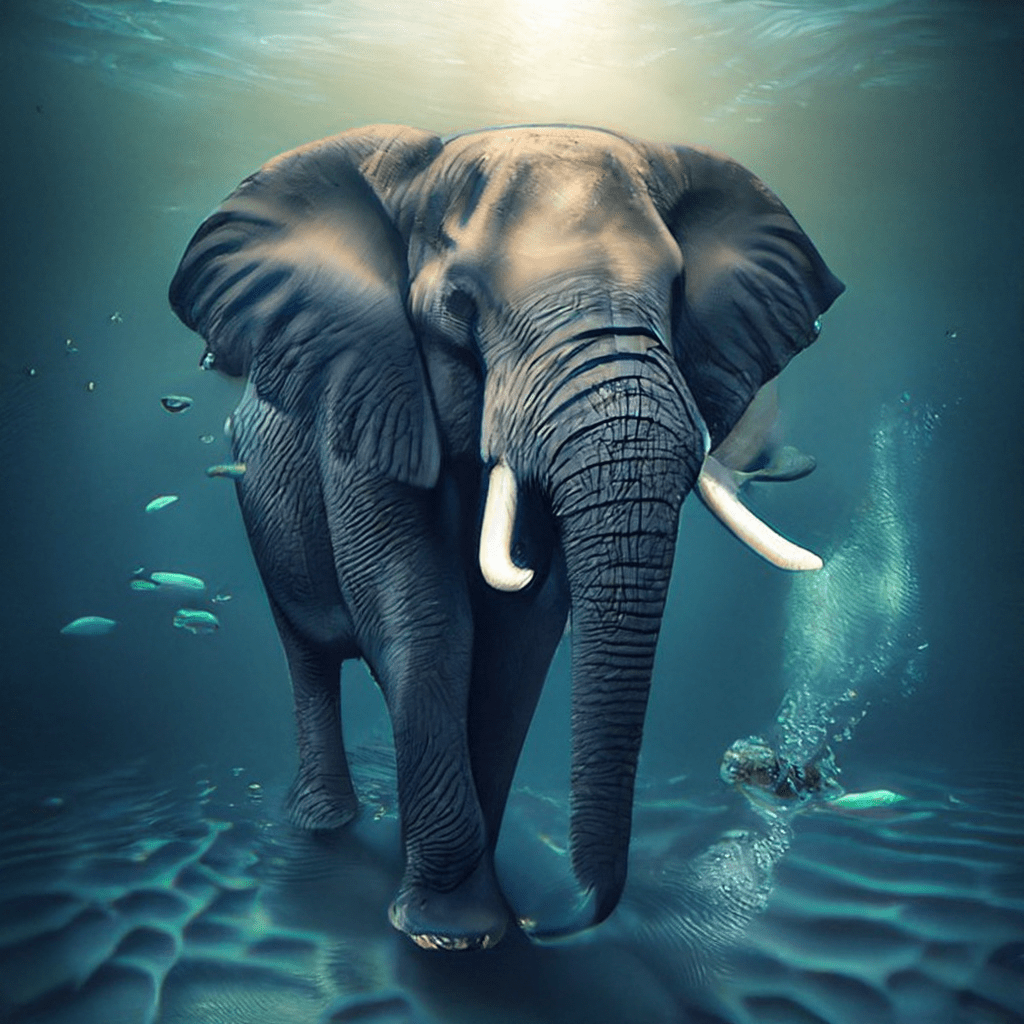}
        \end{subfigure}\\
        \begin{subfigure}{0.25\textwidth}
    \includegraphics[width=\textwidth]{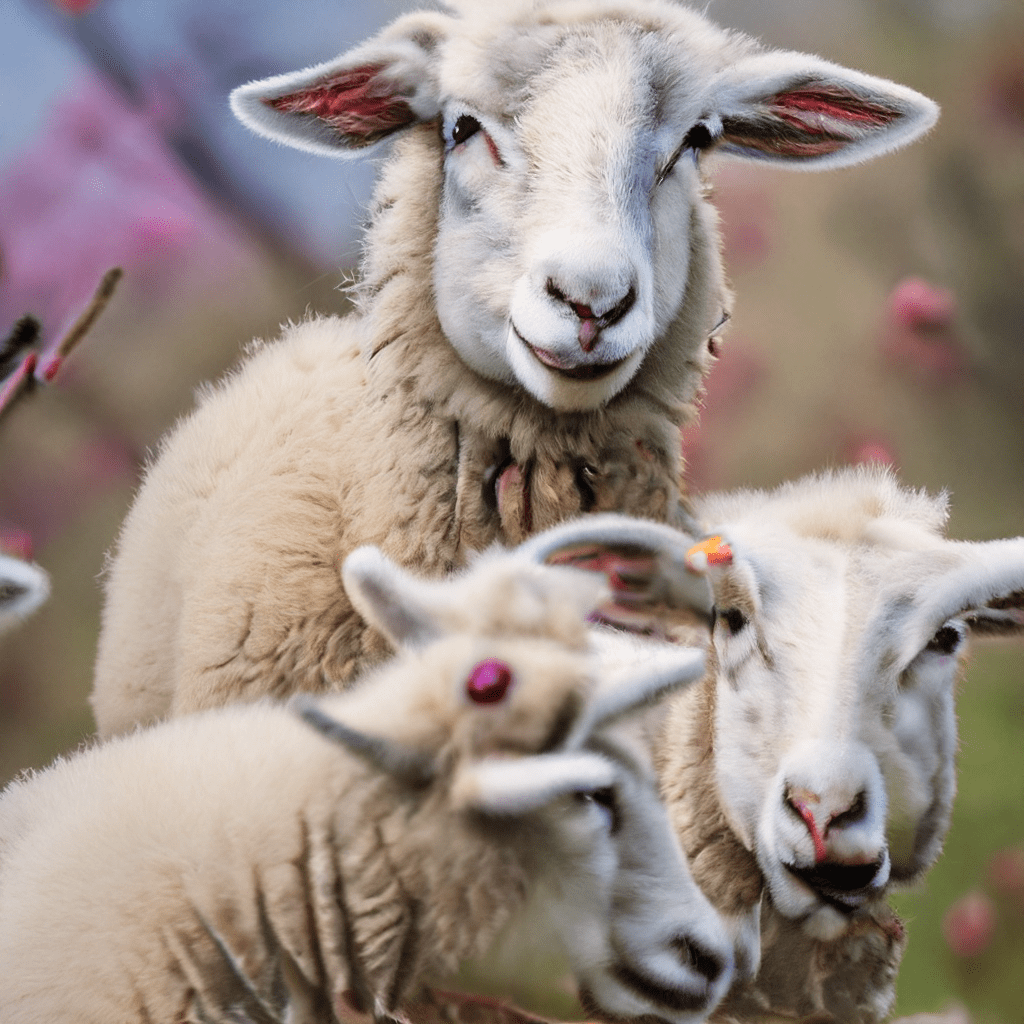}
        \end{subfigure}
        \begin{subfigure}{0.25\textwidth}
    \includegraphics[width=\textwidth]{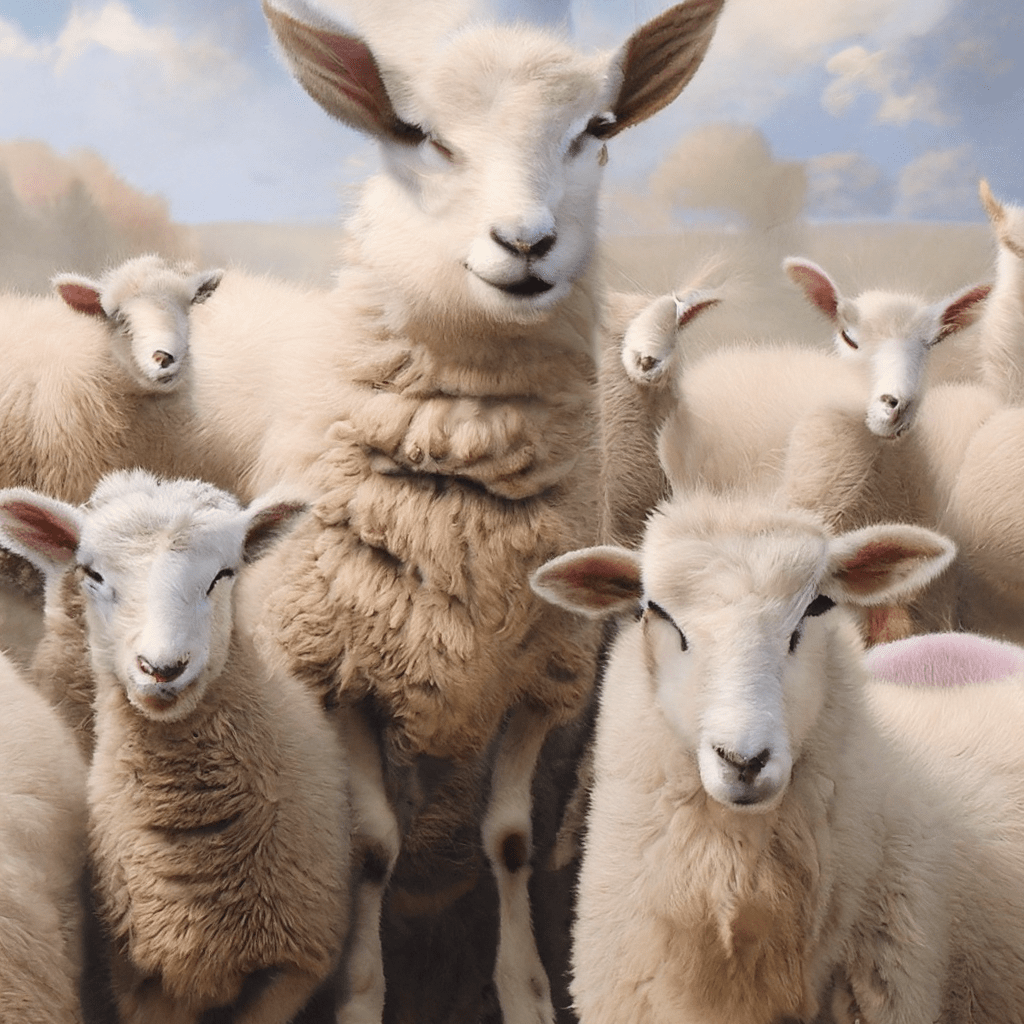}
        \end{subfigure}
        \begin{subfigure}{0.25\textwidth}
    \includegraphics[width=\textwidth]{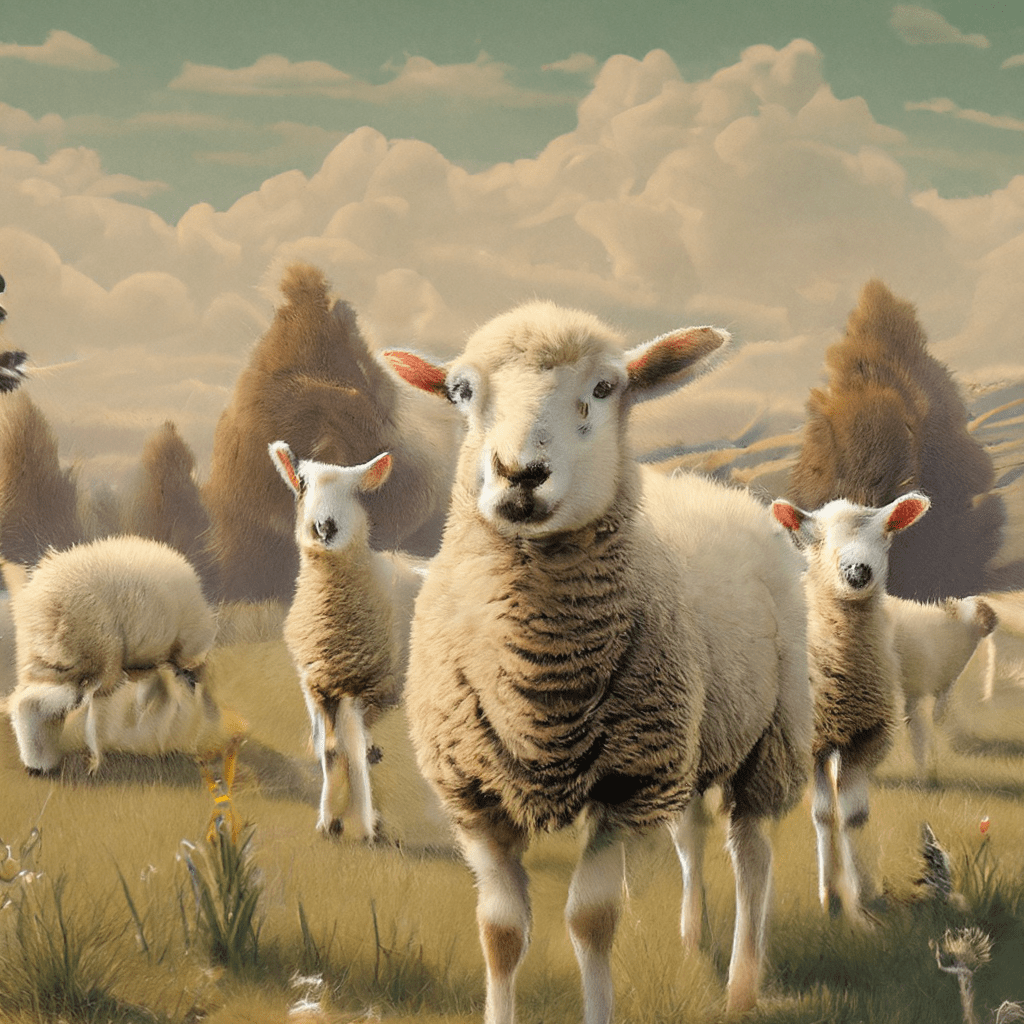}
        \end{subfigure}
        \begin{subfigure}{0.25\textwidth}
    \includegraphics[width=\textwidth]{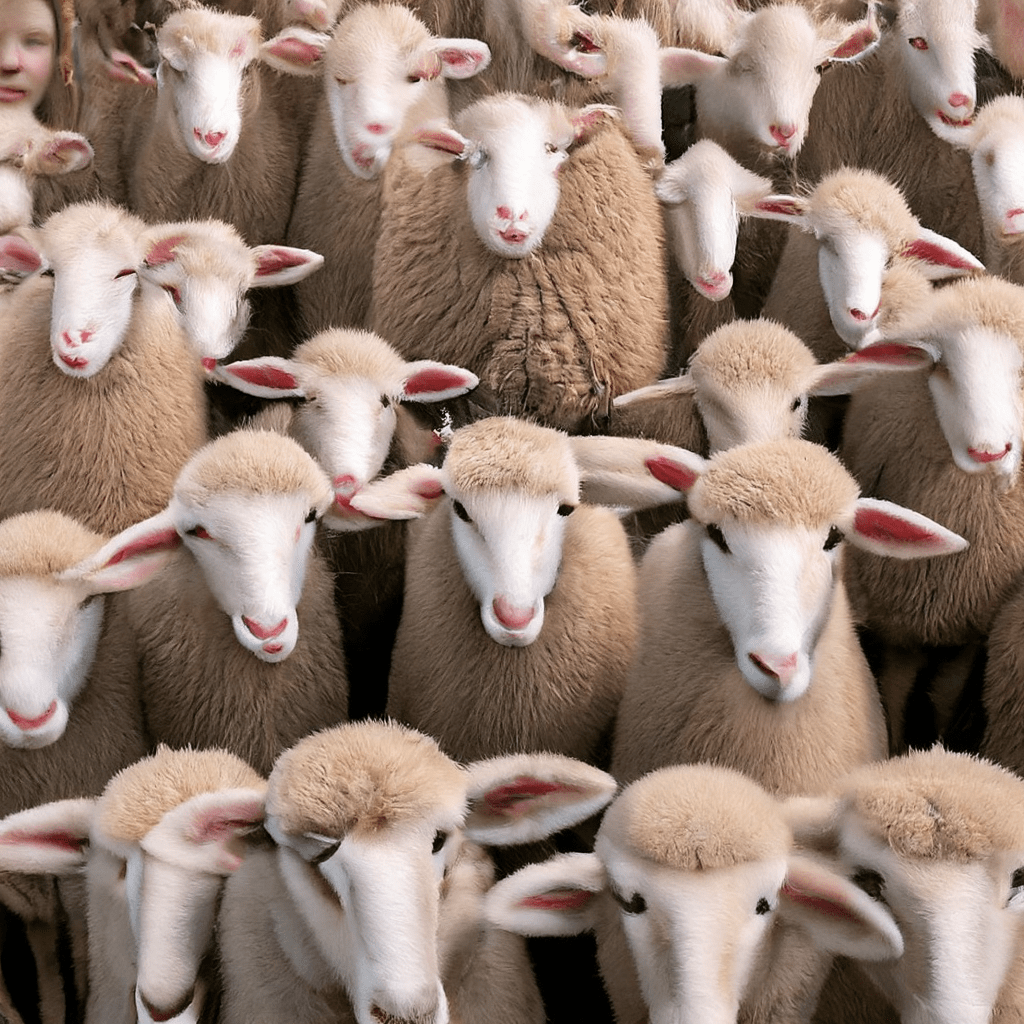}
        \end{subfigure}
    \end{tabular}
    \caption{\texttt{turtle swimming underwater. aesthetic. Fantasy.}, elephant swimming underwater. aesthetic. Fantasy.}, \texttt{flock of sheep. aesthetic. Fantasy.}
\end{figure}

\begin{figure}[H]
    \centering
    \begin{tabular}{cc}
        \begin{subfigure}{0.25\textwidth}
    \includegraphics[width=\textwidth]{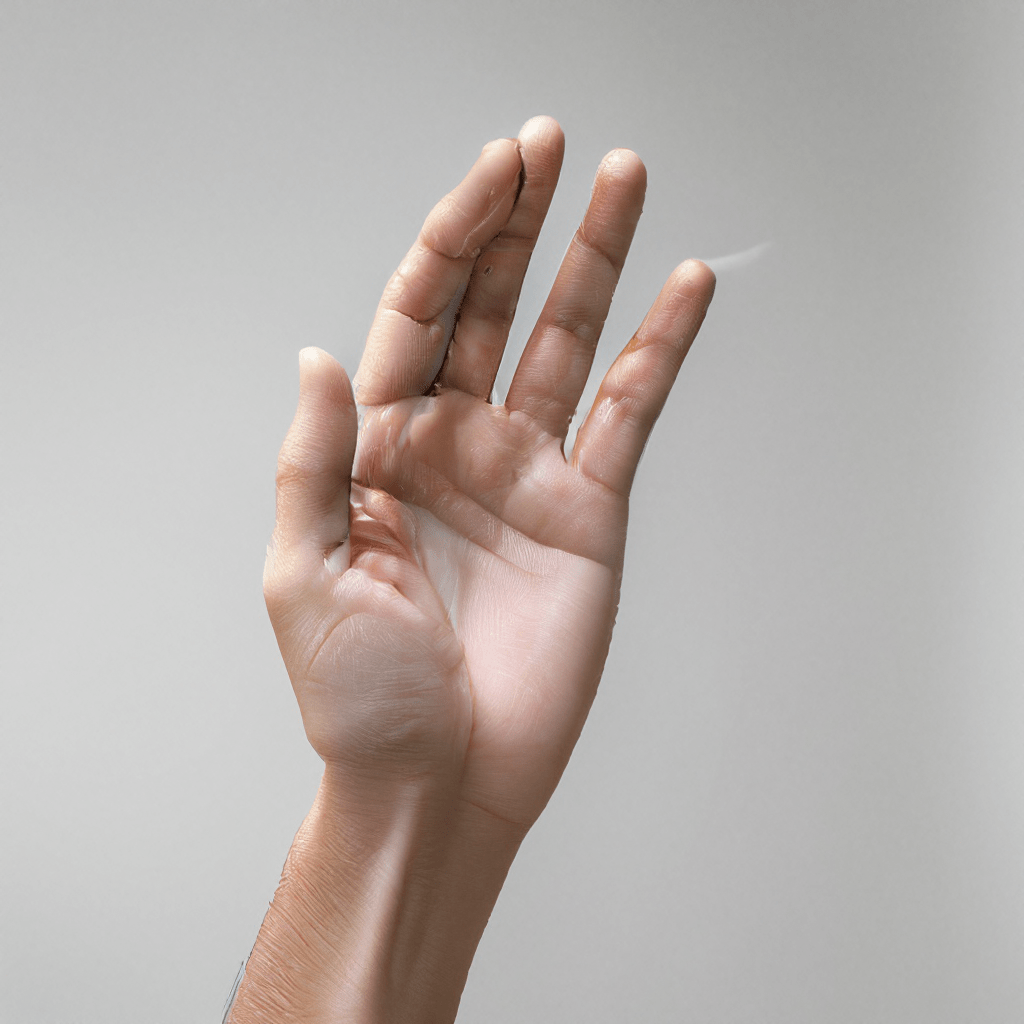}
        \end{subfigure}
        \begin{subfigure}{0.25\textwidth}
    \includegraphics[width=\textwidth]{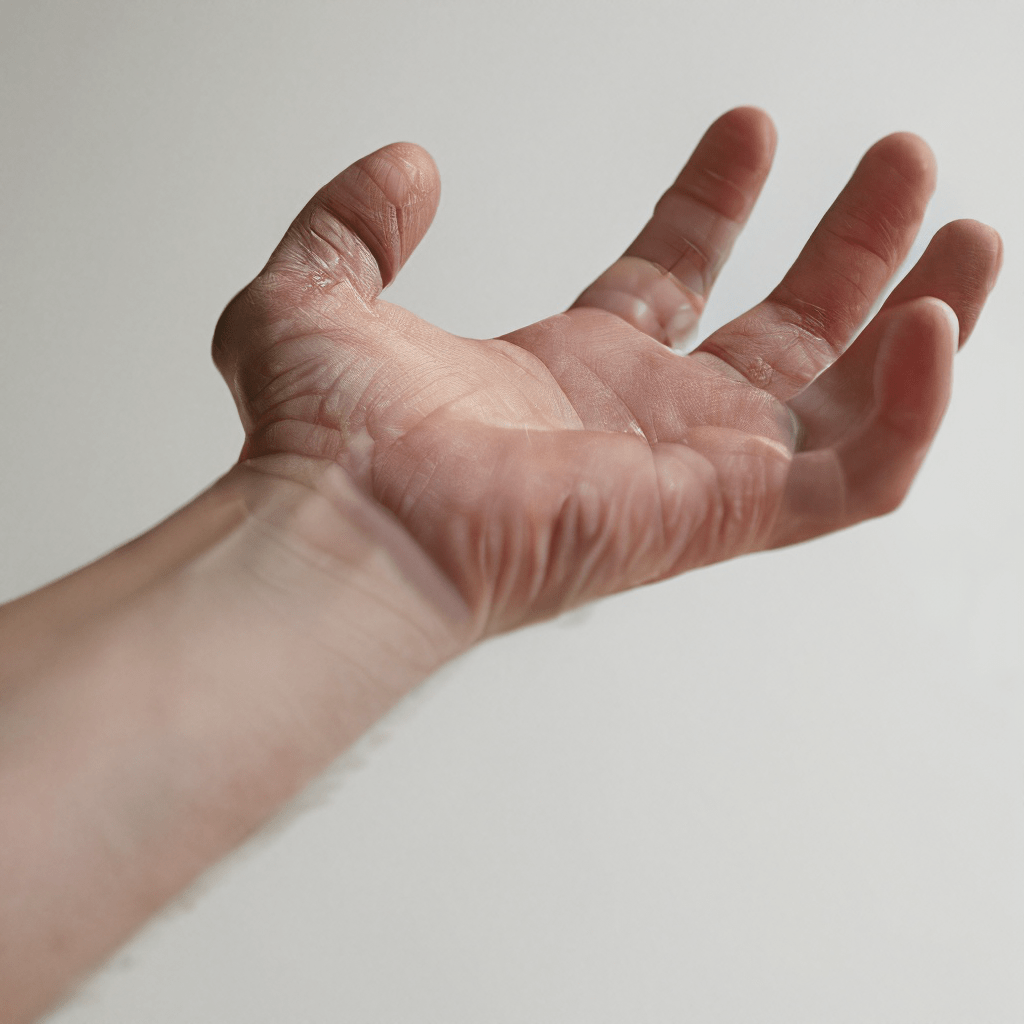}
        \end{subfigure}
        \begin{subfigure}{0.25\textwidth}
    \includegraphics[width=\textwidth]{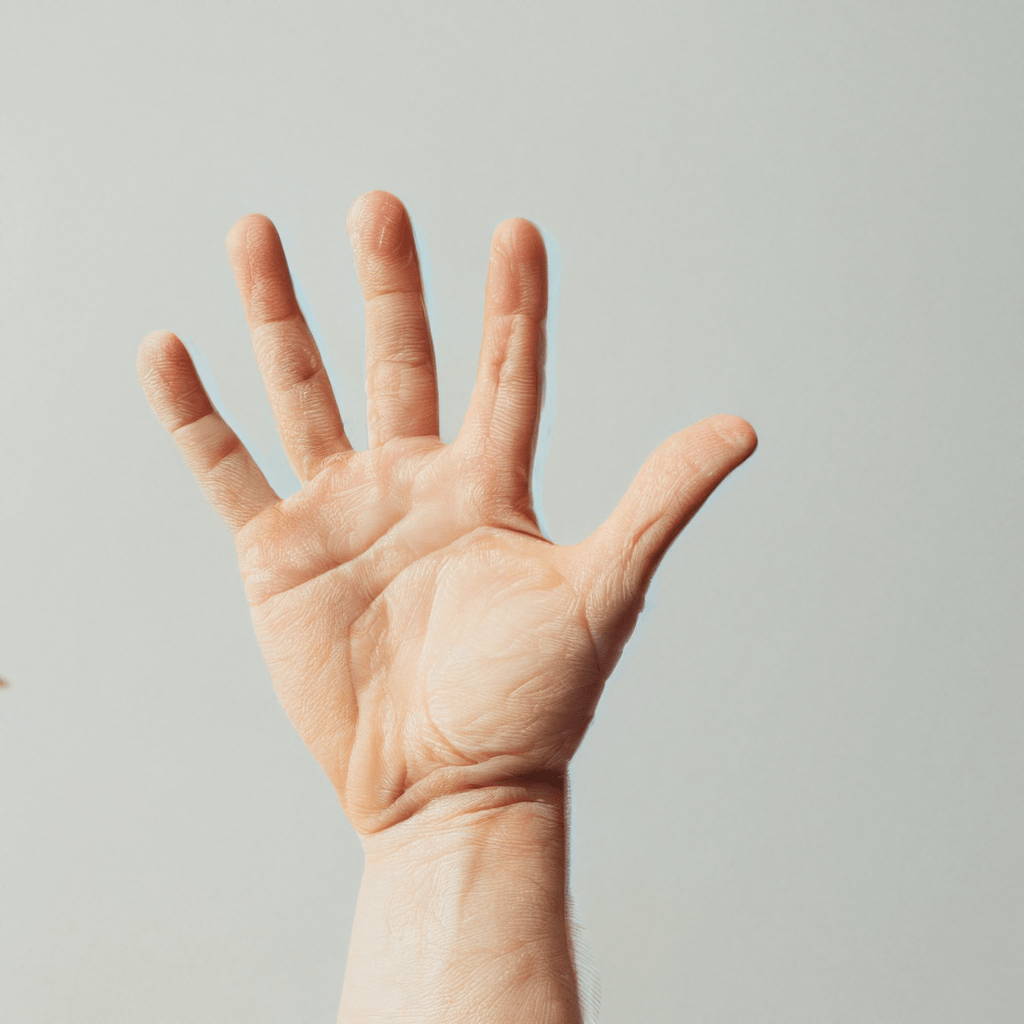}
        \end{subfigure}
        \begin{subfigure}{0.25\textwidth}
    \includegraphics[width=\textwidth]{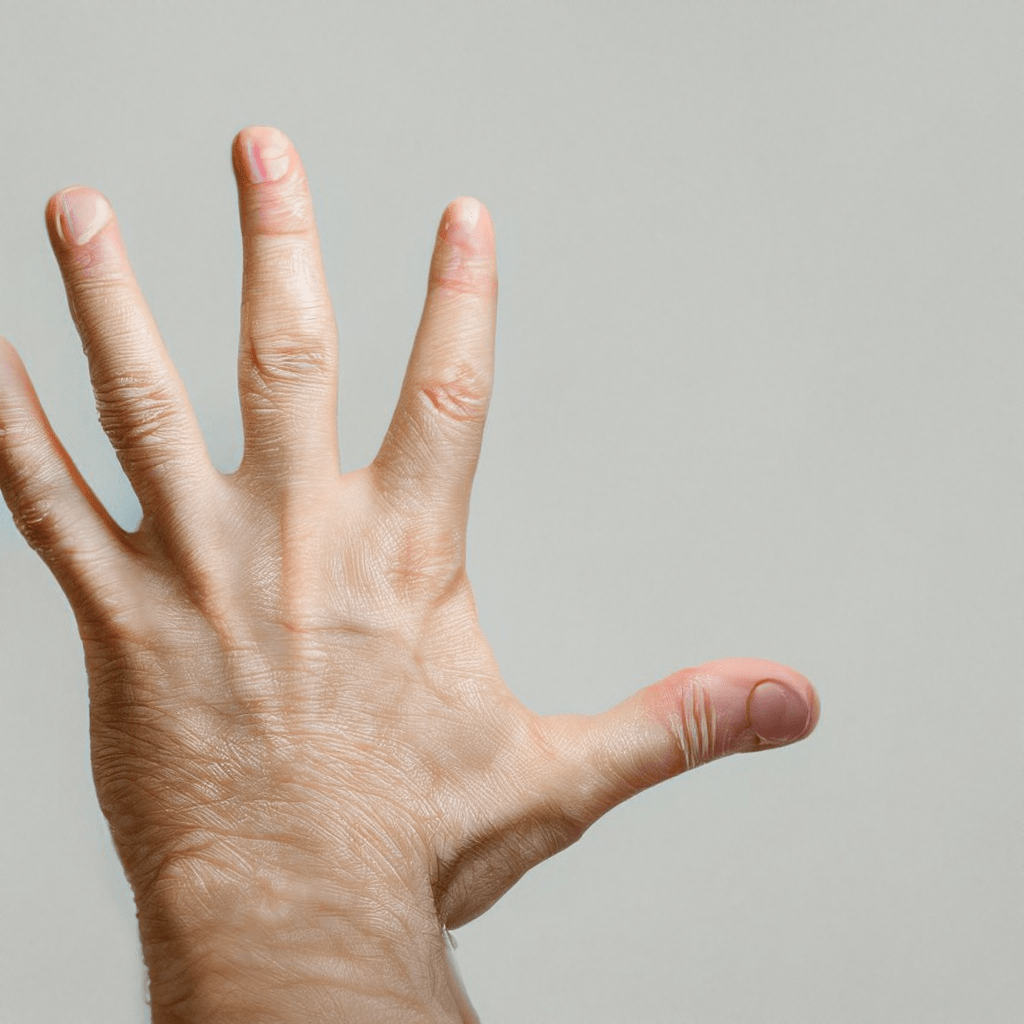}
        \end{subfigure}\\
        \begin{subfigure}{0.25\textwidth}
    \includegraphics[width=\textwidth]{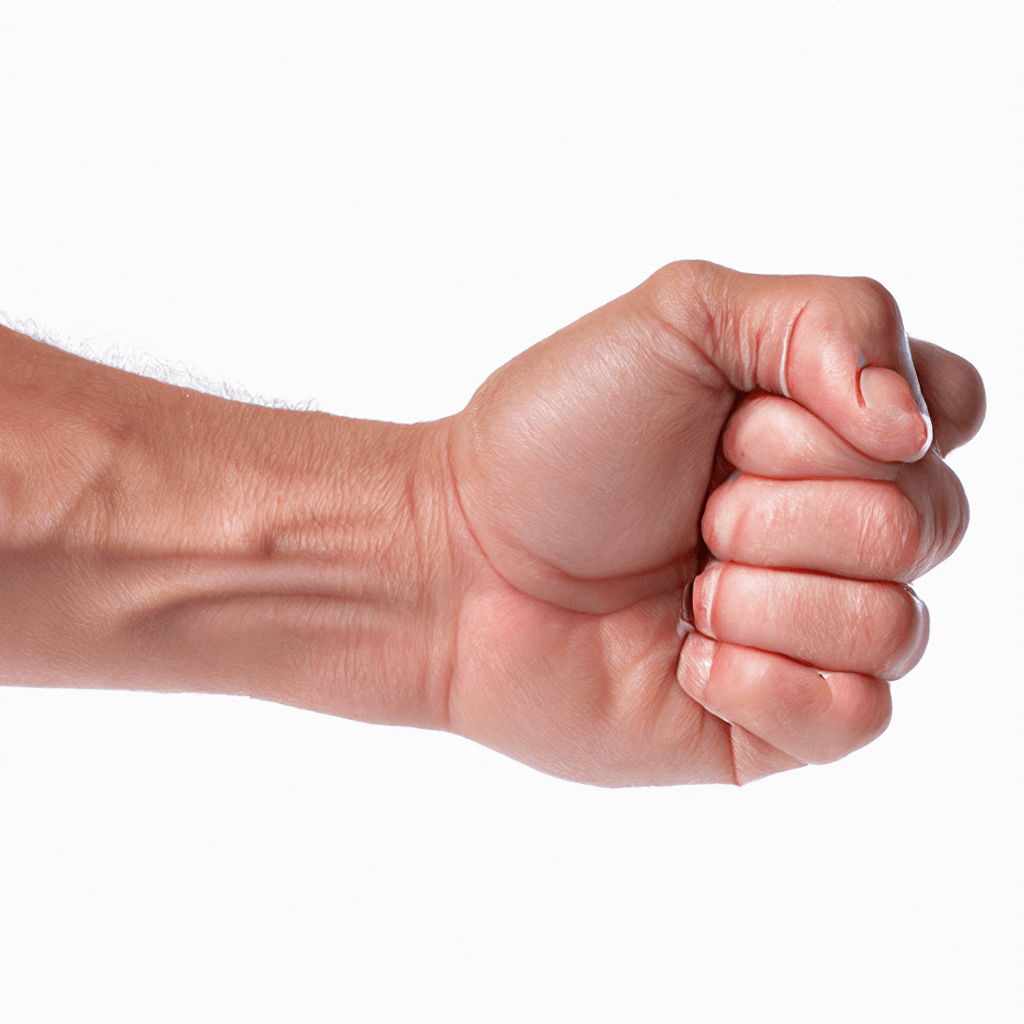}
        \end{subfigure}
        \begin{subfigure}{0.25\textwidth}
    \includegraphics[width=\textwidth]{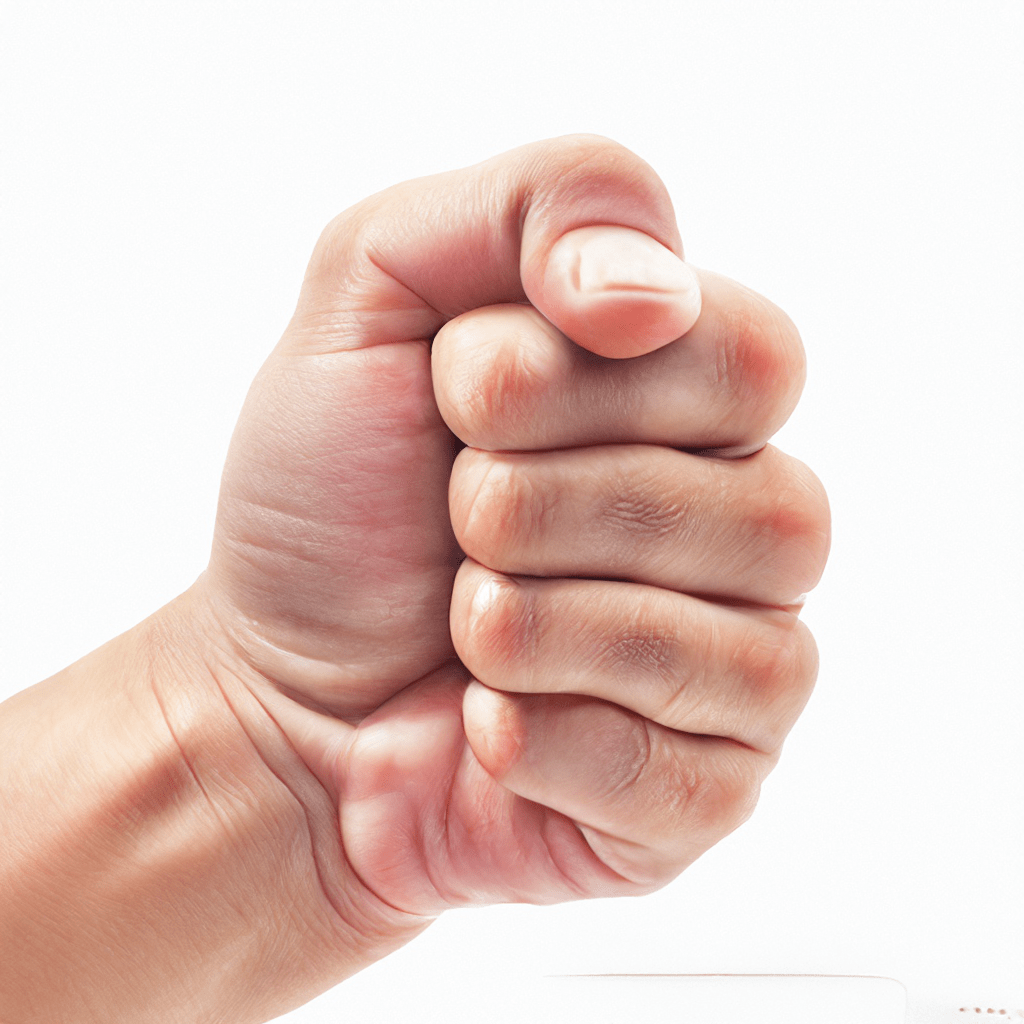}
        \end{subfigure}
        \begin{subfigure}{0.25\textwidth}
    \includegraphics[width=\textwidth]{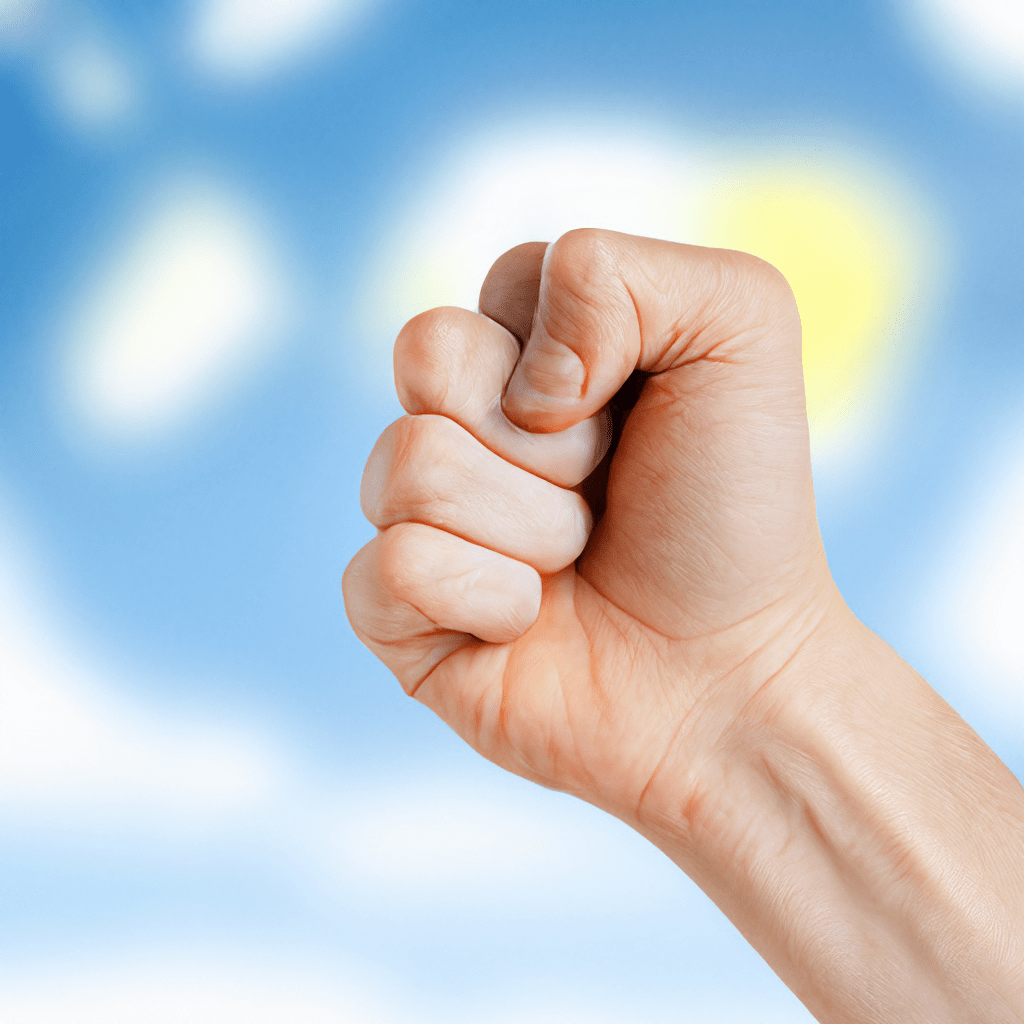}
        \end{subfigure}
        \begin{subfigure}{0.25\textwidth}
    \includegraphics[width=\textwidth]{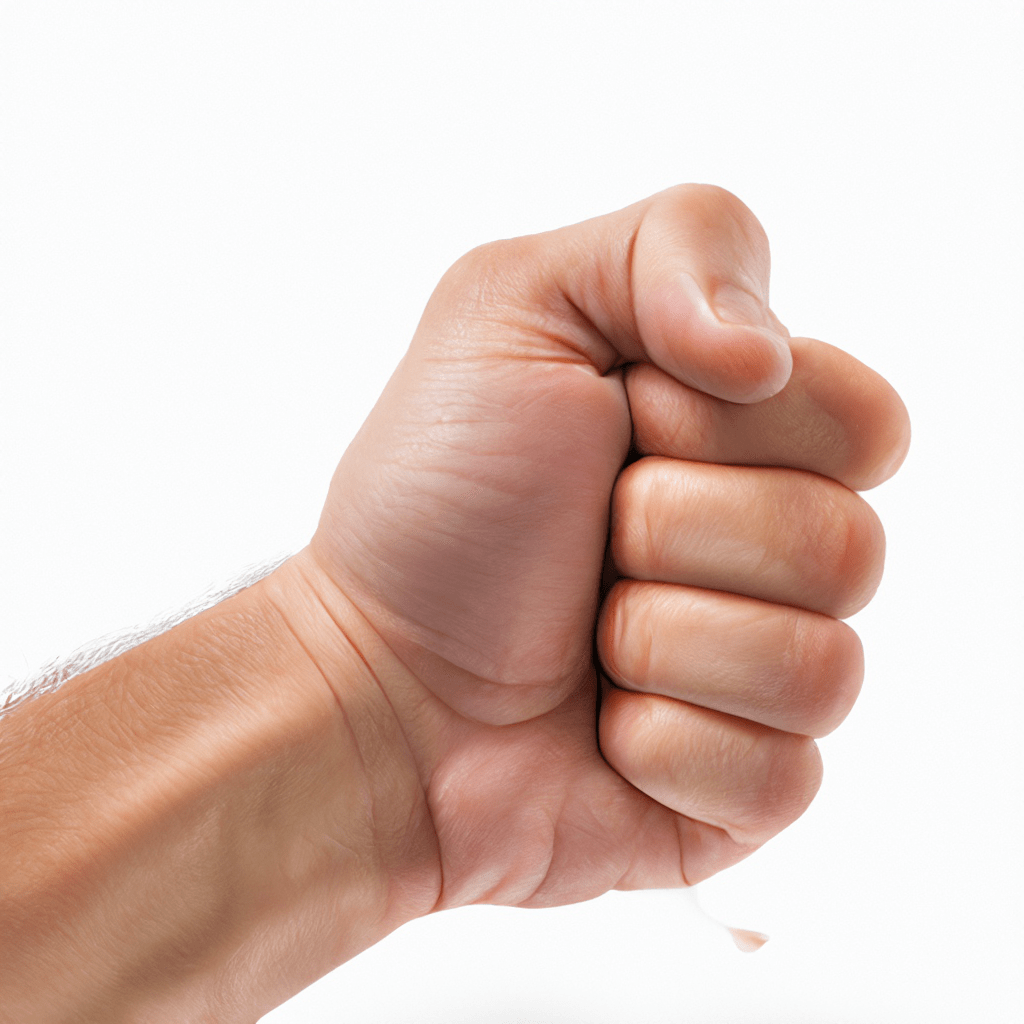}
        \end{subfigure}
    \end{tabular}
    \caption{\texttt{open hand, hand model. 4k. white background}, \texttt{fist, hand model. 4k. white background}}
\end{figure}
\section{Supervised Fine Tuning}
\subsection{Hyper-Parameters}\label{sec:sft_hp}
To maintain a balanced dataset during training, we implemented an up/down sampling strategy with a threshold of 3/0.3.  This process was executed on the 760M and 7B models using 64 and 128 80GB A100s, respectively. We assembled our training examples into sequences of length 4096. Preliminary experiments were conducted to identify optimal learning rates from a range of ${ 1\mathrm{e}{-5}, 3\mathrm{e}{-5}, 5\mathrm{e}{-5}, 1\mathrm{e}{-4}}$ and per-GPU batch sizes from ${4, 8, 16}$ using our validation split. The selected hyperparameters are cataloged in Table \ref{tab:sft_para}. Throughout the fine-tuning phase, our models processed approximately 30 billion tokens.
\begin{table}[h]
\centering\small
\begin{tabular}{l|cccccc}
\toprule
Model & \#~GPUS & Seq Length & Batch Size & LR & Warm-up Steps & \#~Tokens \\
\midrule
CM3Leon-760m  & 64 & 4096        & 2M & 5e-05     & 150    & 30B      \\
CM3Leon-7b   & 128  & 4096      & 2M & 5e-05     & 150    & 30B      \\
\bottomrule
\end{tabular}
\caption{Fine-tuning parameters for \model{} models }
\label{tab:sft_para}
\end{table}
\subsection{Training Data Breakdown}
\begin{table}[ht]
\centering
\resizebox{\textwidth}{!}{
\begin{tabular}{p{0.2\textwidth}|p{0.65\textwidth}|p{0.12\textwidth}}
\toprule
Dataset & Template &  \# Examples \\
\midrule
\multicolumn{3}{c}{Image Focused Datasets} \\
\midrule
InstructPix2Pix & Edit first image following the instruction \texttt{<break>} \{image1\} \texttt{<break>} {edit instruction} \texttt{<break>} \{image 2\} & 127k \\[0.5cm]

OCR & draw ``\{ocr\_content\}'' \texttt{<break>} \{image\} & 300k \\[0.5cm]

Object Detection & Generate high quality image of \{caption\} with segmentations \{obj1\} at \{loc1\}, \{obj2\} at \{loc2\} ... \texttt{<break>} \{image\} & 3M \\[0.5cm]

Edge-to-Image & Make high quality image from canny edge features \texttt{<break>} \{edge image\} \texttt{<break>} \{caption\} \texttt{<break>} \{image\} & 1M \\[0.5cm]

Seg-to-Image & Make high quality image from a segmentation map \texttt{<break>} \{seg image\} \texttt{<break>} \{caption\} \texttt{<break>} \{image\} & 1M \\[0.5cm]

Hed-to-Image & Make high quality image from hed features \texttt{<break>} \{seg image\} \texttt{<break>} \{caption\} \texttt{<break>} \{image\} & 1M \\[0.5cm]

Pose-to-Image & Make high quality image from openpose features \texttt{<break>} \{seg image\} \texttt{<break>} \{caption\} \texttt{<break>} \{image\} & 142k \\[0.5cm]

Depth-to-Image & Make high quality image from depth features \texttt{<break>} \{depth image\} \texttt{<break>} \{caption\} \texttt{<break>} \{image\} & 1M \\[0.5cm]

Norm-to-Image & Make high quality image from 3D norm features \texttt{<break>} \{depth image\} \texttt{<break>} \{caption\} \texttt{<break>} \{image\} & 1M \\[0.5cm]

Scribbe-to-Image & Make high quality image from children's scribbles \texttt{<break>} \{scribble image\} \texttt{<break>} \{caption\} \texttt{<break>} \{image\} & 500k \\

\midrule
\multicolumn{3}{c}{Text Focused Datasets} \\
\midrule

\makecell[l]{COCO Captioning \\ \citep{chen2015microsoft}} & \makecell[l]{\{caption\} \texttt{<break>} \{image\} \\ Describe the given picture. \{caption\} \texttt{<break>} \{image\} \\ } & 591k \\[0.5cm]

\makecell[l]{Flickr30k \\ \citep{young2014image}} & \makecell[l]{\{caption\} \texttt{<break>} \{image\} \\ Describe the given picture. \{caption\} \texttt{<break>} \{image\} \\ } & 144k \\[0.5cm]

\makecell[l]{Image Paragraph \\ \citep{krause2017hierarchical}} & \makecell[l]{Describe the given picture in very detail. \{caption\} \texttt{<break>} \\ \{image\} \\ Describe all the objects in the given image in very detail. \\ \{caption\} \texttt{<break>} \{image\} \\ Generate a long caption for the given image. \\ \{caption\} \texttt{<break>} \{image\} } & 14k \\[0.5cm]

\makecell[l]{Localized Narratives \\ \citep{pont2020connecting}} & \makecell[l]{Describe the given picture in very detail. \{caption\} \texttt{<break>} \\ \{image\} \\ Generate a long narration of what is happening in the \\ given image. \{caption\} \texttt{<break>} \{image\} \\ Generate a long caption for the given image. \\ \{caption\} \texttt{<break>} \{image\} } & 164k \\[0.5cm]

\makecell[l]{VQA2 \\ \citep{goyal2017making}} & \makecell[l]{Question: \{question\} Answer: \{answer\}. \texttt{<break>} {image} \\ Question: \{question\} [newline] \{answer\} \texttt{<break>} {image} \\ Question: \{question\} The answer is \{answer\}. \texttt{<break>} {image} } & 1.3M \\[0.5cm]

\makecell[l]{VizWiz \\ \citep{gurari2018vizwiz}} & \makecell[l]{Question: \{question\} Answer: \{answer\}. \texttt{<break>} {image} \\ Question: \{question\} [newline] \{answer\} \texttt{<break>} {image} \\ Question: \{question\} The answer is \{answer\}. \texttt{<break>} {image} } & 92k \\[0.5cm]

\makecell[l]{OKVQA \\ \citep{marino2019ok}} & \makecell[l]{Question: \{question\} Answer: \{answer\}. \texttt{<break>} {image} \\ Question: \{question\} [newline] \{answer\} \texttt{<break>} {image} \\ Question: \{question\} The answer is \{answer\}. \texttt{<break>} {image} } & 26k \\[0.5cm]

\makecell[l]{ScienceQA \\ \citep{lu2022learn}} & \makecell[l]{ Question: \{question\} [newline] Context: \{context\} [newline] \\ Options: \{choices\_text\} [newline] Answer: \{answer\}. \\ \texttt{<break>} \{image\}  \\  Question: \{question\} [newline] Context: \{context\} [newline] \\ Options: \{choices\_text\} [newline] Answer: Let's think \\ step-by-step: \{ explanation\} So the answer is \{answer\}. \\ \texttt{<break>} \{image\}} & 6k \\[0.5cm]

\bottomrule
\end{tabular}
} 
\caption{Details of the datasets and their prompt templates used in our supervised fine-tuning of \model{} models.}
\label{tab:sft-datasets}
\end{table}
\subsection{More Qualitative Samples}
\begin{figure}[h]
    \centering
    \includegraphics[width=\linewidth]{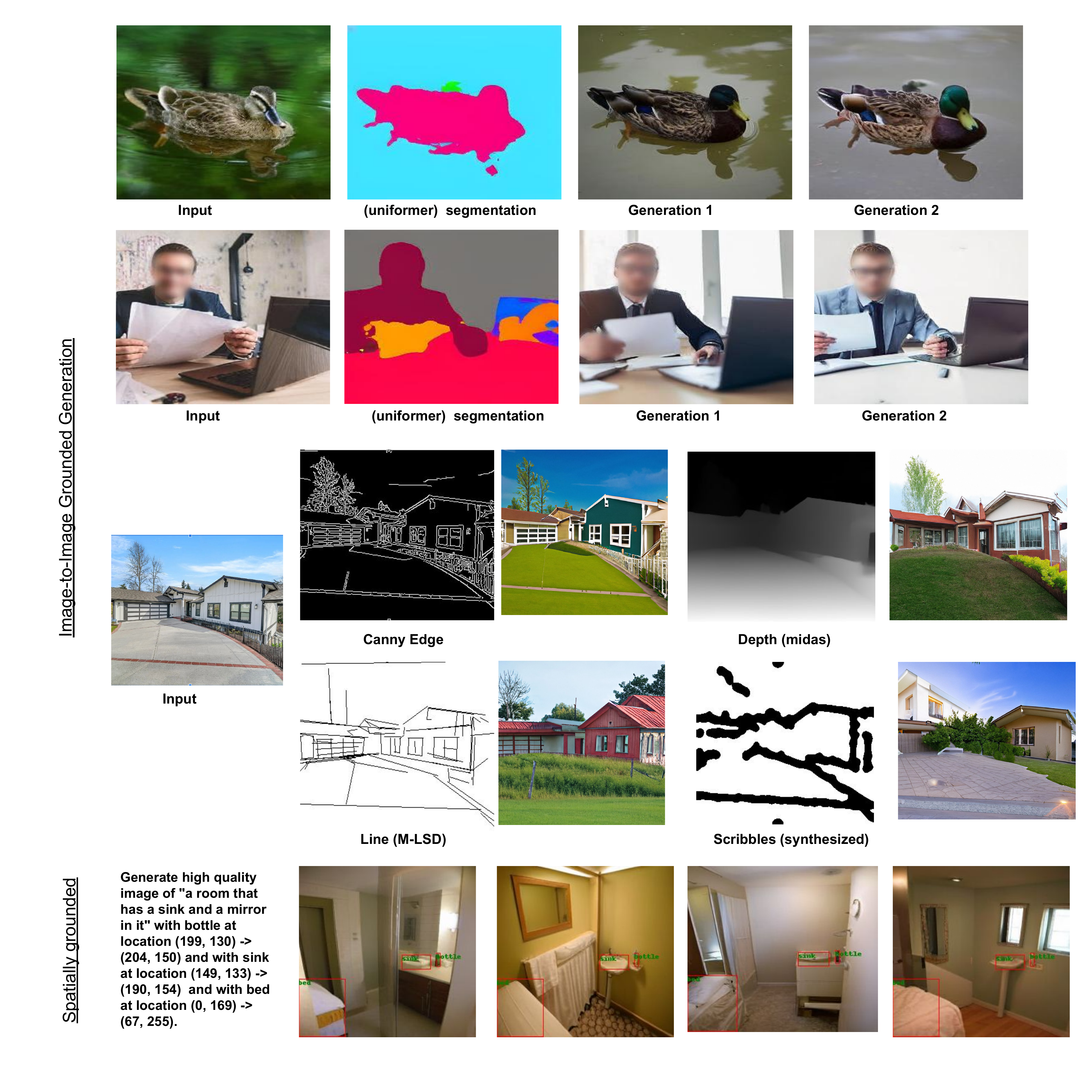}
    \vspace*{-5mm}
     \caption{Qualitative examples of finetuned CM3Leon-7b model. Human faces are blurred to remove PII information.
    \label{fig:MORE_sft_image}
    }
\end{figure}
\begin{figure}[h]
    \centering
    \includegraphics[width=\linewidth]{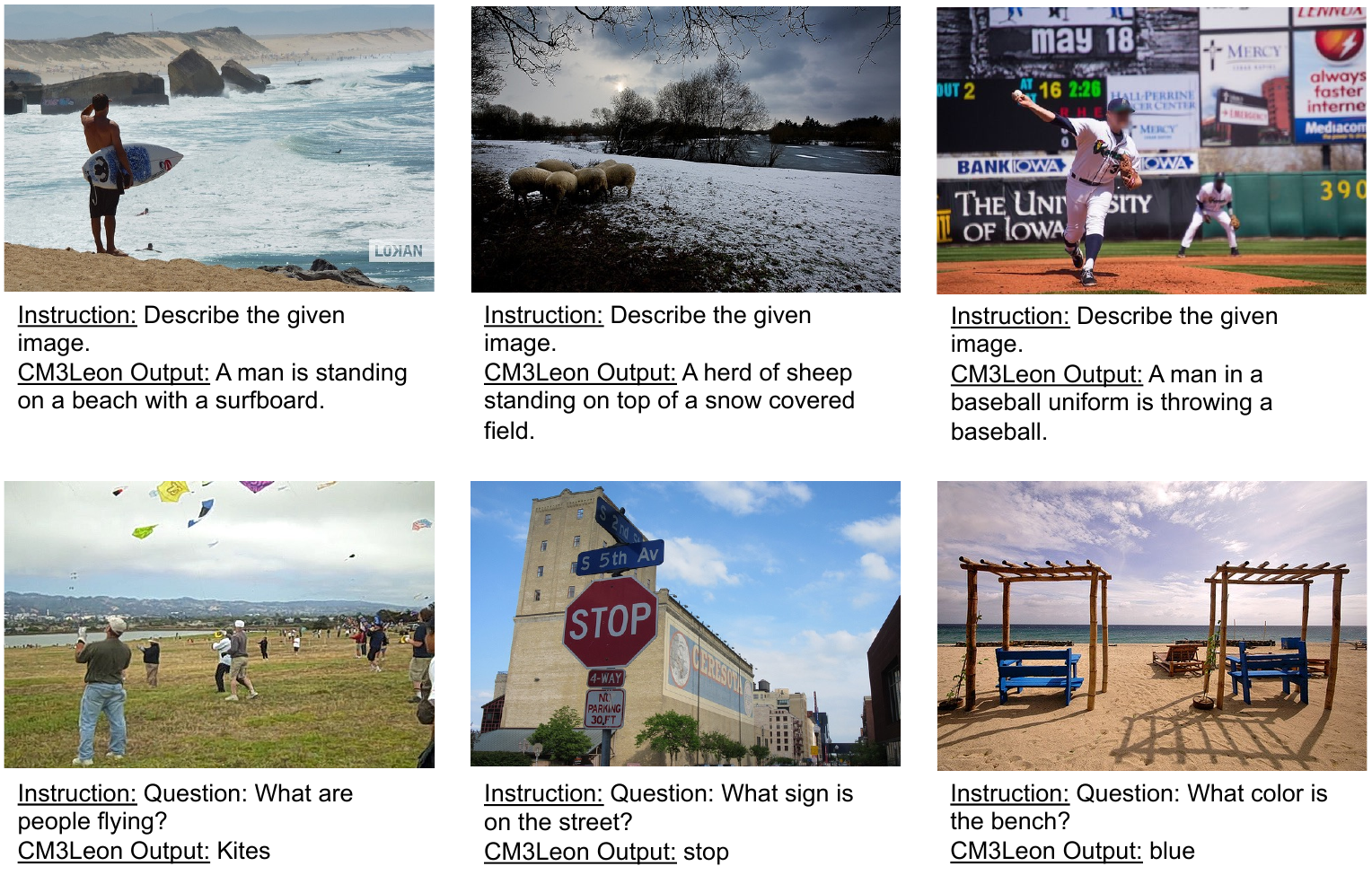}
    \vspace*{-5mm}
     \caption{Qualitative examples showing our SFT-\model{}-7B model's generations for image captioning and visual question answering tasks.  Human faces are blurred to remove PII information.
    \label{fig:caption_vqa_examples}
    }
\end{figure}

\end{document}